%% file: iclr2027_conference.tex
\documentclass{article} 
\usepackage{iclr2027_conference,times}

\input{math_commands.tex}

\usepackage{hyperref}
\usepackage{url}

\usepackage[T1]{fontenc}
\usepackage[utf8]{inputenc}
\usepackage{microtype}
\usepackage{inconsolata}
\usepackage{xurl}

\usepackage{amsmath}
\usepackage{array}
\usepackage{booktabs}
\usepackage{multirow}
\usepackage{graphicx}
\usepackage{xcolor}
\usepackage{algorithm}
\usepackage{algpseudocode}
\algrenewcommand\algorithmicrequire{\textbf{Input:}}
\algrenewcommand\algorithmicensure{\textbf{Output:}}
\usepackage{wrapfig}
\usepackage{capt-of}

\definecolor{ACacheDarkGreen}{RGB}{18,105,63}
\newcommand{\TableHeadMultirow}[1]{\multirow[c]{2}{*}{\raisebox{-0.35ex}{#1}}}
\newcommand{\TableModelMultirow}[1]{\multirow[c]{6}{*}{\raisebox{-0.35ex}{#1}}}
\newcommand{\smartparagraph}[1]{%
  \par\vspace{0pt plus 0.5pt}
  \noindent\textbf{#1}
}

\newcounter{smartlistcounter}
\newenvironment{smartlist}{%
    \begin{list}{(\roman{smartlistcounter})}{%
        \usecounter{smartlistcounter}%
        \setlength{\topsep}{-\parskip}\addtolength{\topsep}{0pt plus 0.5pt}
        \setlength{\partopsep}{0pt plus 0.5pt}%
        \setlength{\itemsep}{0pt plus 0.5pt}%
        \setlength{\parsep}{0pt plus 0.5pt}%
        \setlength{\leftmargin}{2em}
        \setlength{\labelsep}{0.5em}
        \setlength{\labelwidth}{1.5em}
        \setlength{\listparindent}{0pt}%
    }%
}{%
    \end{list}%
}

\title{Affix Cache for Diffusion Large Language Models}

\author{%
\textbf{Kaihua Liang$^{1}$ \quad An Zhong$^{1}$ \quad Xin Tan$^{2}$ \quad Zafar Ayyub Qazi$^{1,3}$} \\
\textbf{Hong Xu$^{2}$ \quad Jian Weng$^{1}$ \quad Marco Canini$^{1}$} \\
$^{1}$KAUST \quad $^{2}$CUHK \quad $^{3}$LUMS \\
\texttt{\{kaihua.liang,marco\}@kaust.edu.sa}
}

\newif\ifarxiv
\arxivtrue
\ifarxiv\iclrfinalcopy\fi

\begin{document}

\maketitle
\ifarxiv\lhead{Preprint.}\fi
\ifarxiv{\renewcommand{\thefootnote}{}\footnotetext{Code: \raisebox{-0.12em}{\includegraphics[height=0.95em]{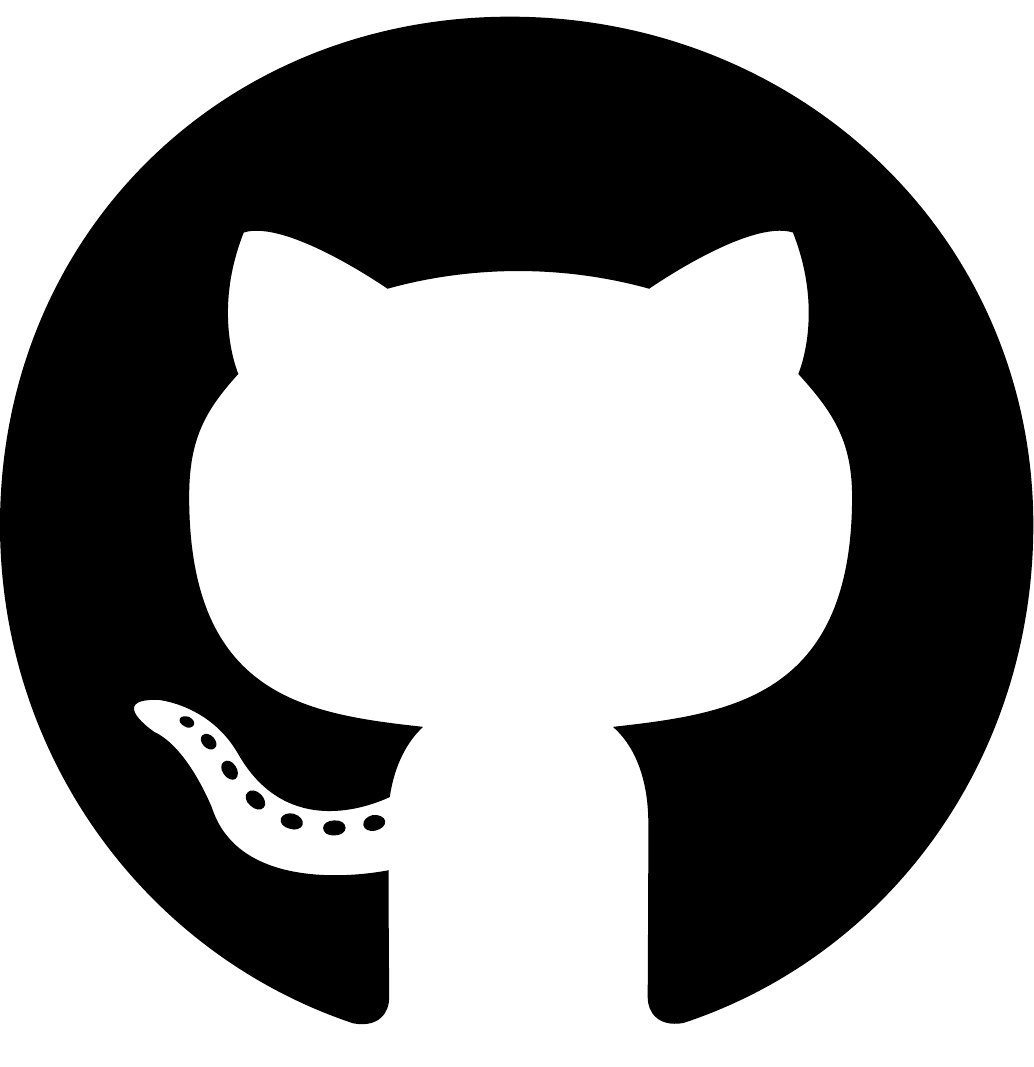}}\hspace{0.25em}\url{https://github.com/sands-lab/ACache}}}\fi

\input{sections/0_Abstract}
\input{sections/1_Introduction}
\input{sections/2_Background}
\input{sections/3_ACache_Design}
\input{sections/4_System_Design}
\input{sections/5_Evaluation}
\input{sections/6_Conclusion}
\input{sections/7_Statements}

\bibliography{ACache}
\bibliographystyle{iclr2027_conference}

\clearpage
\appendix
\input{sections/8A_Appendix_Discussion}
\input{sections/8B_Appendix_SystemDesign}
\input{sections/8C_Appendix_EvaluationDetails}
\input{sections/8D_Appendix_AdditionalEvaluation}

\end{document}

%% file: math_commands.tex
\usepackage{amsmath,amsfonts,bm}

\def\eqref#1{equation~\ref{#1}}

\def\1{\bm{1}}

\DeclareMathAlphabet{\mathsfit}{\encodingdefault}{\sfdefault}{m}{sl}
\SetMathAlphabet{\mathsfit}{bold}{\encodingdefault}{\sfdefault}{bx}{n}



%% file: sections/0_Abstract.tex
\begin{abstract}
Diffusion Large Language Models (DLLMs) enable non-autoregressive decoding, but efficient inference support remains immature: unlike autoregressive models, whose requests reuse a shared prefix key-value (KV) cache, DLLMs use bidirectional attention, so a shared context's KV states depend on the tokens still being decoded, leaving directly reused caches stale and full recomputation necessary. We present \textbf{ACache}, a cross-request cache reuse mechanism for shared spans, or \textit{affixes}, at any position: prefix, infix, or suffix. ACache measures the influence of affix tokens on the masked generation region to identify a small request-specific subset as \textit{Anchor Tokens}, and recomputes only their KV states while reusing the remaining affix cache. Built on state-of-the-art intra-request caching mechanisms, ACache recovers most of the accuracy lost to direct affix-cache reuse on average when recomputing around 20\% of affix tokens, and at that budget preserves more accuracy than selection criteria adapted from prior cross-request cache-reuse systems. We co-design ACache with a modern inference engine, whose attention reads each request's recomputed Anchor KV states alongside one affix cache shared across concurrent requests.
Against the same system with only intra-request caching, ACache cuts recompute latency by up to 56.7\%, translating to as much as $1.71\times$ end-to-end throughput, while reducing peak KV cache memory by up to 45.8\%.
\end{abstract}

%% file: sections/1_Introduction.tex
\section{Introduction}

Large Language Models (LLMs) have traditionally been autoregressive (AR); Diffusion Large Language Models (DLLMs)~\citep{nie2025llada, ye2025dream} instead decode by iterative denoising, predicting masked positions in arbitrary order over a context that attention spans bidirectionally, which enables low-latency parallel decoding~\citep{israel2025APD, ben-hamu2025eb, wu2026fast-dllm} and gains in data efficiency and reasoning~\citep{ni2025data-learner, he2026reasoning-with-latent}.

\begin{figure*}[t]
\centering
\includegraphics[width=0.904\textwidth]{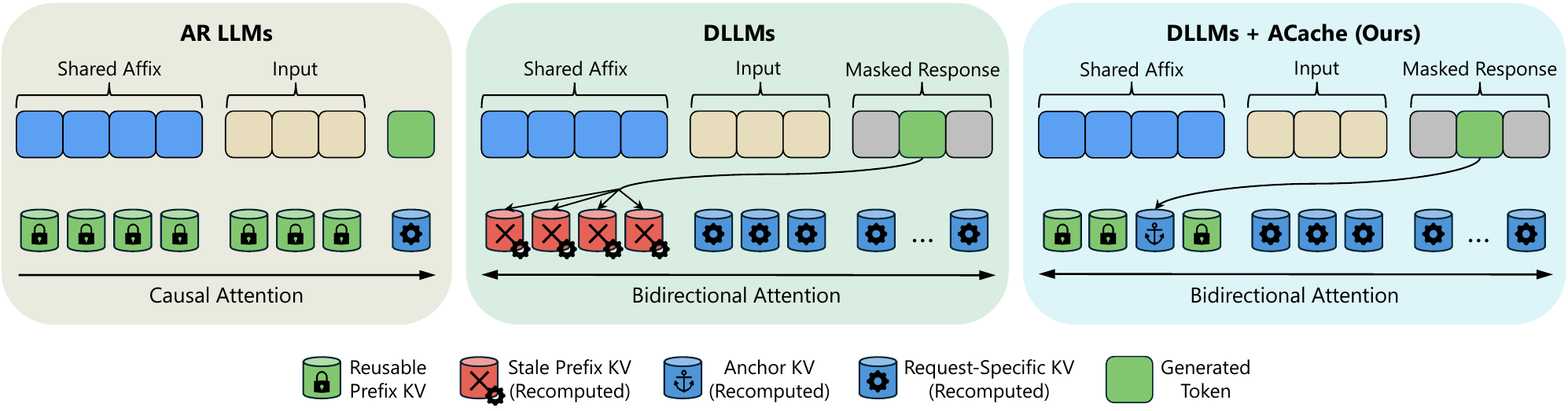}
\caption{\textbf{Cache reuse in AR LLMs and DLLMs.} \textit{Left}: AR LLMs reuse a shared prefix cache directly. \textit{Middle}: in DLLMs, bidirectional attention makes that cache depend on the tokens still being decoded, leaving it stale. \textit{Right}: ACache recomputes only the Anchor Tokens' KV states.}
\label{fig:ar-vs-dllm}
\end{figure*}

For AR LLMs (Figure~\ref{fig:ar-vs-dllm}, left), prefix caching is a well-established mechanism for efficiency: requests reuse a key-value (KV) cache for shared prefixes such as system prompts and prompt templates~\citep{gim2024prompt_cache, zheng2024sglang}. In contrast, DLLMs use bidirectional attention, so the KV states of a shared prefix depend on the tokens still being decoded, and a directly reused prefix cache goes stale (Figure~\ref{fig:ar-vs-dllm}, middle). As a result, practical DLLM inference frameworks~\citep{ma2025dkv-cache, liu2026dllm-cache, wu2026fast-dllm, jiang2026d2cache, nguyen-tri2026attention, xiong2026prefilling-dllm} confine reuse to a single request (intra-request reuse), recomputing KV states at intervals to maintain context consistency; reusing caches directly across requests substantially degrades quality (Section~\ref{sec:evaluation}).

To address this limitation, we propose ACache (Affix Cache), a cross-request cache reuse mechanism for DLLMs. We study shared contiguous text spans, which we call \textit{affixes}, as a span may be a prefix, infix, or suffix of the context. Unlike text cached by AR LLMs, which always precedes generation, a suffix follows the generation region, as in infilling-style prompts~\citep{lou2024sedd, xu2026pill}. We observe that the influence of an affix's tokens on the masked generation region concentrates on a small subset, and that this subset stays stable as decoding proceeds. ACache exploits this finding (Figure~\ref{fig:ar-vs-dllm}, right): it reuses the shared affix cache while recomputing only that subset, which we term \textit{Anchor Tokens}. ACache selects them once per request by mask-to-affix attention importance.

ACache composes with intra-request caching mechanisms rather than replacing them, and brings to DLLMs the cross-request cache reuse on which many AR inference-stack optimizations build~\citep{zheng2024sglang, srivatsa2025preble, pan2025fasttree, qin2025mooncake, yao2025cache-blend, yuan2026dualmap, ouyang2026unicache, wang2026prophetkv}. We build ACache on Nano-vLLM, a vLLM-style engine with paged memory management~\citep{kwon2023vllm, geeeekexplorer2025NanoVLLM}, and co-design the engine with the algorithm: since within the affix only the Anchor KV differs across requests, the engine stores one affix cache, and a per-request read-slot map lets attention read it alongside the request's own Anchor KV. We also extend the engine to reuse that cache at different offsets and to interleave recomputation with decoding at step granularity~\citep{yu2022orca, vllm-omni2026}.

To summarize, our key contributions are as follows:
\begin{smartlist}
    \item We present ACache, a cross-request cache reuse mechanism for shared spans in DLLMs beyond prefixes, recomputing only the Anchor Tokens it selects by mask-to-affix importance.
    \item We implement ACache on top of two state-of-the-art intra-request DLLM caching frameworks, dKV-Cache~\citep{ma2025dkv-cache} and Fast-dLLM~\citep{wu2026fast-dllm}, where recomputing around 20\% of affix tokens recovers most of the accuracy lost to direct reuse on average.
    \item We co-design an engine that shares one affix cache: up to 56.7\% lower recompute latency, $1.71\times$ end-to-end throughput, and 45.8\% lower peak KV cache memory over intra-request caching.
\end{smartlist}

%% file: sections/2_Background.tex
\section{Background \& Motivation}

\smartparagraph{Diffusion large language models.}
DLLMs now scale to billions of parameters~\citep{nie2025smdm, inception2025mercury, song2025seed-diffusion}. LLaDA~\citep{nie2025llada} trains a diffusion model from scratch, while Dream~\citep{ye2025dream}, DiffuLLaMA~\citep{gong2025diffullama}, and TESS 2~\citep{tae2025tess2} start from pretrained AR LLMs; both routes remain competitive on general, mathematical, and coding tasks, and keep improving~\citep{zhu2025llada-moe, zhu2026llada-1.5, nie2026illada}. A separate line of models instead adopts block diffusion~\citep{han2023ssd-lm, arriola2025block-diffusion}, modeling tokens bidirectionally within a block and causally across blocks~\citep{bie2025llada2, wu2026fast-dllm-v2, fu2026efficient-dlm, cheng2026sdar, diffusiongemma2026}. That structure keeps AR-style prefix caching workable, but confines both bidirectional context and parallelism to a block. ACache targets standard DLLMs, whose bidirectional attention spans the full sequence.

\smartparagraph{Caching techniques in DLLMs.}
DLLMs update a masked generation region over many decoding steps~\citep{nie2025llada, ye2025dream}, paying for full-sequence attention per step, motivating cross-step KV reuse. Unlike AR LLMs, DLLMs are unable to maintain causal decoding's append-only KV cache~\citep{pope2023efficiently}: with bidirectional attention, KV states change during generation, prompt tokens included, so entries go stale rather than accumulate~\citep{nguyen-tri2026attention}. Recent work therefore explores approximate caching: one line reuses KV states or intermediate features across steps on a predetermined schedule, varying the recomputation amount~\citep{ma2025dinfer, liu2026dllm-cache, hu2026flash-dlm}. dKV-Cache~\citep{ma2025dkv-cache} carries over already-decoded tokens' KV states with a one-step delay, while Fast-dLLM~\citep{wu2026fast-dllm} retains KV states within a decoding block~(Table~\ref{tab:recompute-comparison}). A second line replaces the predetermined schedule with a runtime signal, selecting recomputation or eviction timing and layer or token granularity~\citep{huang2026maskkv, song2026sparse-dllm, cheong2026entropy-cache, nguyen-tri2026attention, jiang2026d2cache, wu2026dynamic-dllm, frumkin2026dare, yang2026wavefilter, xiong2026prefilling-dllm, lee2026polestar, he2026archer}.{\parfillskip=0pt\par}%

\setlength{\intextsep}{6pt}
\setlength{\columnsep}{20pt}
\begin{wrapfigure}[7]{r}{0.38\textwidth}
\centering
\setlength{\abovecaptionskip}{0pt}
\includegraphics[width=\linewidth]{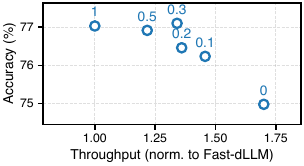}
\caption{\textbf{Quality--throughput tradeoff.} ACache's Fast-dLLM-based system on LLaDA/GSM8K, 2-shot prefix, batch 16. Labels: Recompute ratios.}
\label{fig:quality-throughput-tradeoff}
\end{wrapfigure}
\noindent\raisebox{4pt}{\begin{minipage}[t]{0.57\textwidth}
\setlength{\abovecaptionskip}{0pt}
\captionof{table}{\textbf{Cache recomputation.} Full-sequence attention frequency and cross-request prefix cache reuse.}
\label{tab:recompute-comparison}
\small
\centering
\begin{tabular}{>{\arraybackslash}m{0.7in}>{\arraybackslash}m{1.15in}>{\centering\arraybackslash}m{0.75in}}
\toprule
Method & Full-sequence attention & Cross-req. cache reused? \\
\midrule
AR LLM & \textbf{Once}, at prefill & Yes \\
dKV-Cache & \textbf{Every few steps} & No \\
Fast-dLLM & \textbf{Every block} boundary & No \\
\bottomrule
\end{tabular}
\vspace{9.39pt}
\end{minipage}}
\setlength{\columnsep}{10pt}

Both lines remain intra-request. Concurrently, BiCache~\citep{go2026bicache} reuses shared prefix KV states across requests in shallow layers where they remain stable, whereas ACache selects tokens rather than layers.

\textbf{Why DLLMs need affix cache reuse.} With intra-request caching, DLLMs still recompute the shared prefix repeatedly, with each full-sequence pass comparable to a prefill~\citep{kedia2026sangam}, whereas an AR LLM pays for prefill once~(Table~\ref{tab:recompute-comparison}). Across requests, AR LLMs reuse that prefix cache, while DLLMs rebuild it for each request and retain separate copies concurrently. Figure~\ref{fig:quality-throughput-tradeoff} previews ACache's quality--throughput tradeoff. Direct reuse offers the highest throughput in this comparison, at the cost of lower accuracy. Selecting 20\%--30\% of the shared tokens as request-specific Anchors, ACache exceeds $1.3\times$ full recomputation's throughput at comparable accuracy. Beyond this prefix example, shared text can occur elsewhere in the context: CacheBlend~\citep{yao2025cache-blend}, EPIC~\citep{hu2025epic}, ProphetKV~\citep{wang2026prophetkv}, and SparseX~\citep{zhang2026sparsex} already reuse mid-prompt text in AR inference, though their cached text still precedes generation. In a DLLM, the shared span may also follow the generation region, as in infilling-style prompts~\citep{lou2024sedd, xu2026pill}, where bidirectionality couples it to generation. To the best of our knowledge, ACache is the first cross-request cache reuse mechanism for DLLMs that is not prefix-restricted: it reuses a shared span's cache anywhere, recomputing only the part a request needs.

%% file: sections/3_ACache_Design.tex
\section{ACache Design}
\label{sec:design}

We formalize the DLLM inference loop~\citep{sahoo2024mdlm}. Let $T$ denote the total number of decoding steps, $z^{(t)}$ the sequence at step $t$, and $\mathcal{M}_t$ the masked positions, with $t$ descending from $T$ to $0$, so $\mathcal{M}_T$ is the whole generation region, while the tokens outside it carry the request's context. They may precede the generation region, follow it, or surround it. At each step, the model predicts all masked positions in parallel:
\begin{equation}
\hat{y}^{(t)}_i \sim p_\theta(\cdot \mid z^{(t)}), \quad i \in \mathcal{M}_t
\end{equation}
Only a subset of predictions is committed. Let $\mathcal{C}_t \subseteq \mathcal{M}_t$ denote positions committed at step $t$. The next state is updated as:
\begin{equation}
z^{(t-1)}_i =
\left\{
\begin{array}{ll}
\hat{y}^{(t)}_i, & i \in \mathcal{C}_t \\
\texttt{[MASK]}, & i \in \mathcal{M}_t \setminus \mathcal{C}_t \\
z^{(t)}_i, & i \notin \mathcal{M}_t
\end{array}
\right.
\end{equation}
Decoding schedules define $\mathcal{C}_t$ differently: the signal may be a confidence threshold~\citep{nie2025llada, wu2026fast-dllm, luo2026dsb}, an entropy bound~\citep{ben-hamu2025eb}, or inter-token dependencies~\citep{luo2026dawn}. All, however, repeat the same ``predict-then-commit'' cycle.

Now consider a shared affix that occupies a contiguous span of non-masked positions $\mathcal{F} = \{l, l+1, \ldots, r\}$ in $z^{(t)}$, with length $L = |\mathcal{F}|$. Under intra-request caching mechanisms~\citep{ma2025dkv-cache, liu2026dllm-cache, wu2026fast-dllm, jiang2026d2cache, nguyen-tri2026attention, xiong2026prefilling-dllm}, the affix in $\mathcal{F}$ is still recomputed from scratch for each request, at the start or at each of several recomputation points. ACache instead recomputes only a subset of the $L$ affix tokens.

\begin{figure*}[t]
\centering
\includegraphics[width=0.912\textwidth]{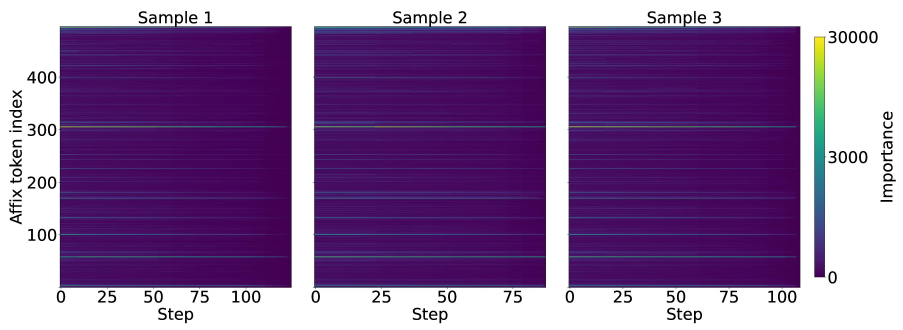}
\caption{\textbf{Anchor Token importance.} Each panel shows a randomly selected GSM8K sample (Section~\ref{sec:evaluation} setup, 2-shot prompts). Color encodes mask-to-affix attention importance over affix tokens (vertical) and decoding steps (horizontal), highlighting a small subset at stable positions across steps.}

\label{fig:anchor-observation}
\end{figure*}

\subsection{Anchor Selection}
We refer to this subset of affix tokens as Anchor Tokens, following prior work that identifies a small critical subset of tokens in the KV cache~\citep{zhang2023eviction, xiao2024streaming-llm}. ACache selects them before the decoding loop starts and keeps that set fixed throughout decoding for each request. To make this selection request-specific while preserving affix reuse, we score affix tokens by mask-to-affix attention, a signal similar to the one MaskKV uses for cache eviction~\citep{huang2026maskkv}. Specifically, we obtain this signal from a single forward pass over the non-affix tokens---the request-owned prompt and the masked generation region---with the affix contributing only its precomputed keys and values.

Formally, let $Q_{\mathrm{mask}}^{(n,h)}$ denote the queries of masked positions at layer $n \in \{1, \dots, N\}$ and head $h \in \{1, \dots, H\}$, let $K_{\mathrm{affix}}^{(n,h)}$ denote the precomputed keys of affix tokens, and let $d_h$ be the per-head hidden dimension. For an affix token $j \in \mathcal{F}$ and a masked position $i \in \mathcal{M}_T$, we denote by $a_{i,j}^{(n,h)}$ the attention weight that masked query $i$ places on affix key $j$, normalized over the affix keys alone:
\begin{equation}
a_{i,j}^{(n,h)} =
\Bigl[
\mathrm{Softmax}\Bigl(
\frac{Q_{\mathrm{mask}}^{(n,h)}
(K_{\mathrm{affix}}^{(n,h)})^\top}{\sqrt{d_h}}
\Bigr)
\Bigr]_{i,j}
\end{equation}
Summing these weights over all layers, heads, and initial masked positions gives the importance score of affix token $j$:
\begin{equation}
s_j = \sum_{n=1}^{N} \sum_{h=1}^{H} \sum_{i \in \mathcal{M}_T} a_{i,j}^{(n,h)}
\end{equation}
ACache then takes the positions of the top-$K$ scores as the Anchor set:
\begin{equation}
\mathcal{A} = \mathrm{TopK}\big(\{s_j\}_{j \in \mathcal{F}}, K\big), \qquad
\mathcal{S} = \mathcal{F} \setminus \mathcal{A}
\end{equation}
Here $\mathcal{S}$ denotes the remaining non-anchor affix tokens. We set $K = \lceil \rho L \rceil$ for a fixed \textit{Anchor ratio} $\rho$, which is the only hyperparameter of ACache; Section~\ref{sec:accuracy-shared-affixes} sweeps this ratio.

Because importance concentrates on a small subset, recomputing only that subset preserves context consistency; because a selection made at request start captures nearly as much importance as one made later, one selection per request suffices, as in prior work~\citep{liu2023scissorhands, li2024snapkv, jiang2026d2cache, huang2026maskkv}. Figure~\ref{fig:anchor-observation} shows both; Appendix~\ref{sec:appendix-anchor-stability} measures them over the full test sets. That selection costs one extra forward pass, cheaper than a full one, against the $T$ decoding steps. Within it, the mask-to-affix aggregation costs $O(N H m L)$, where $m = |\mathcal{M}_T|$; top-$K$ extraction over the $L$ scores then adds $O(L \log K)$. Appendix~\ref{sec:appendix-anchor-selection-overhead} confirms these costs are small.

\subsection{Selective Recomputation}

The shared affix KV cache is precomputed once from a forward pass over the affix span alone, and reused across every request whose context contains that affix; each request selects its Anchor set $\mathcal{A}$ at request start. At all recomputation points, including request start, ACache recomputes the KV states for Anchor Tokens and current request-owned tokens, while tokens in $\mathcal{S}$ continue to use the shared cache. Any decoding schedule can proceed unchanged on top of this cache layout. In this way, the recomputation cost on a shared affix drops from $O(L)$ to $O(K)$, and since $K$ is typically much smaller than $L$, ACache preserves most of the efficiency benefit of affix reuse. At the same time, the Anchor Tokens remain synchronized with the evolving request-specific context.

%% file: sections/4_System_Design.tex
\begin{figure*}[t]
\centering
\includegraphics[width=0.962\textwidth]{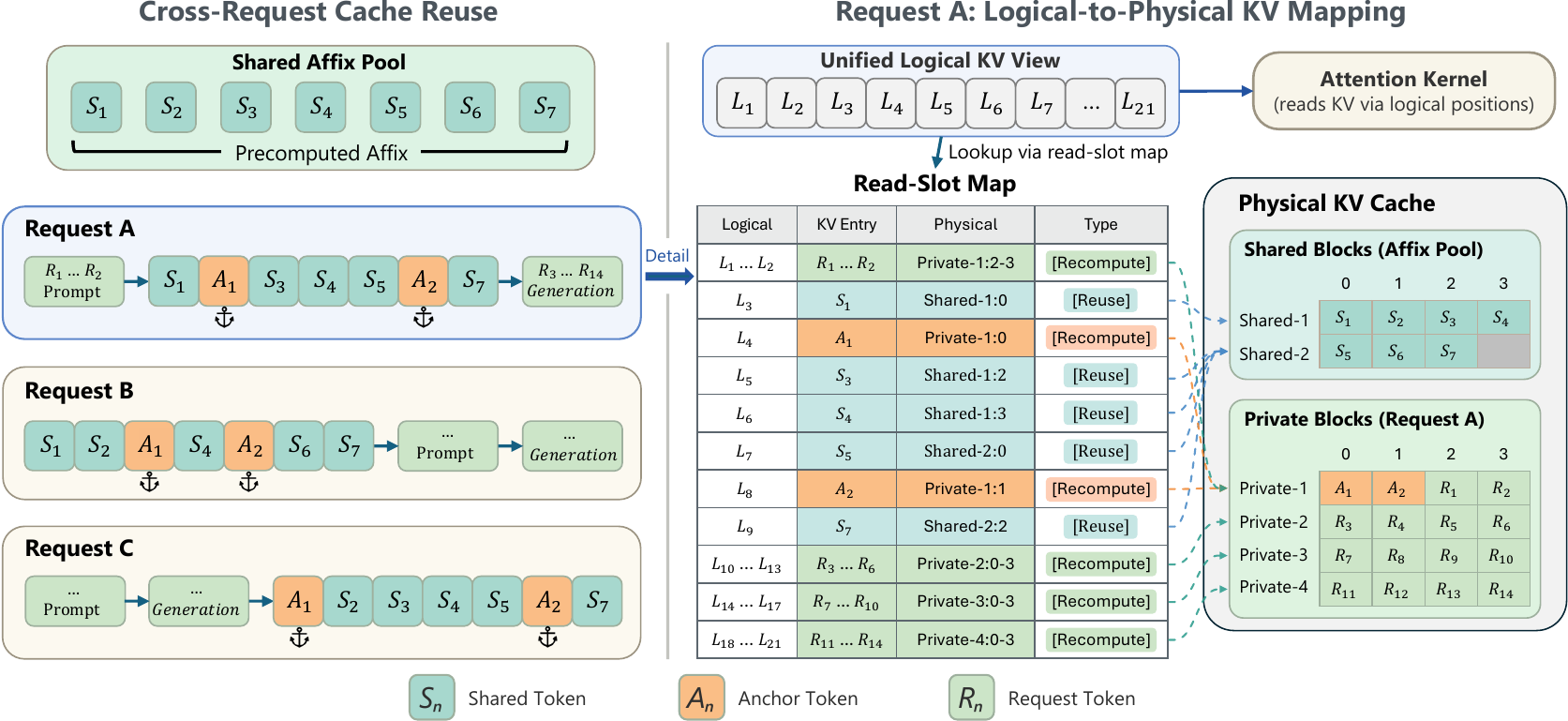}
\caption{\textbf{System architecture of ACache.} \textit{Left}: requests reuse one precomputed affix cache as an infix (A), prefix (B), or suffix (C), selecting request-specific Anchor Tokens. \textit{Right}: one request's read-slot map, resolving each logical position to a private or shared KV slot.
}
\label{fig:system-design}
\end{figure*}

\section{ACache System Support}
\label{sec:system-design}

To make ACache's affix cache reuse pay off, the runtime can rely on none of the three assumptions behind AR prefix caching: every request reuses the whole shared span; every request finds it at offset zero; and no request rewrites its KV once prefill ends. We build that runtime on top of Nano-vLLM~\citep{geeeekexplorer2025NanoVLLM}, a lightweight version of the vLLM engine~\citep{kwon2023vllm}, layering ACache's cache management and admission logic over intra-request caching mechanisms.

\subsection{KV Layout and Read-Slot Map}

The system partitions the paged KV cache into reserved shared blocks and the request-private blocks outside them. The reserved blocks hold the precomputed affix KV cache, stored once rather than per request; each request's private blocks then store the KV of two groups: the Anchor Tokens selected from that affix, and the request-owned prompt and generation-region tokens. Figure~\ref{fig:system-design} shows shared affix cache reuse across requests on the left, and one request's layout on the right.

A simpler design would materialize a blended cache per request: copy the shared affix cache, patch in the recomputed Anchor KV, and run standard attention. Selective-recompute systems for AR inference can afford exactly that~\citep{yao2025cache-blend, hu2025epic, wang2026prophetkv, zhang2026sparsex}, because the blend duplicates nothing that concurrent requests were sharing in memory. ACache's affix cache is shared in memory by every request that uses it: copying fills each request's private blocks with the whole affix cache rather than the Anchors alone, so fewer requests fit in the KV cache concurrently, and restricting the copy to the blocks the Anchors land in barely helps, since the selection criterion scatters Anchors across blocks that hold hundreds of token slots each. In a DLLM, mixed ownership is also the steady state rather than a transient: AR inference freezes context KV at the end of prefill, but Anchor and request-owned KV is rewritten at every recomputation point.

ACache therefore composes shared and private KV in the read path. A per-request \emph{read-slot map} (Figure~\ref{fig:system-design}, right) binds every logical position to the physical KV slot to read from, sending Anchors to private slots and the rest of the affix to shared ones, so attention sees one logical full-context view without a blended copy. Appendix~\ref{sec:appendix-kernel-integration} details the kernel built around this map (Algorithm~\ref{alg:acache-kernel}).

\subsection{Shared-Affix Cache Rotation}

The shared affix cache is stored after positional rotation, as though every affix began at position zero. Rotary embeddings compose additively---rotating by two offsets in sequence equals rotating once by their sum~\citep{su2024roformer}---so that one copy can be used at any starting offset with no approximation (Figure~\ref{fig:system-design}, left). SparseX exploits the same property by re-basing the stored keys, which costs a rotated copy per request~\citep{zhang2026sparsex}. ACache instead keeps the correction on the query side: the kernel counter-rotates the query by the request's offset before scoring it against shared slots, so the shared cache is never rewritten and serves requests at different offsets concurrently. Affixes already at position zero skip the correction altogether. Appendix~\ref{sec:appendix-kernel-integration} details this correction.

\subsection{Admission and Inference Loop}

\setlength{\intextsep}{6pt}
\setlength{\columnsep}{20pt}
\begin{wrapfigure}[23]{r}{0.53\textwidth}
\vspace{-6pt}
\begin{minipage}{\linewidth}
\refstepcounter{algorithm}
\hrule height 1pt
\vspace{2pt}
\noindent\textbf{Algorithm~\thealgorithm} ACache request lifecycle
\label{alg:request-lifecycle}
\vspace{2pt}
\hrule height 0.5pt
\small
\begin{algorithmic}[1]
\Procedure{Admit}{$r$}
\State Reuse $r$'s shared affix cache
\State Select $r$'s Anchor Tokens
\State Allocate private slots for $r$'s Anchor and request-owned tokens
\State Build $r$'s read-slot map
\EndProcedure
\Procedure{InferenceLoop}{}
\While{requests remain}
    \State Advance or finish sequences
    \If{any sequence just finished}
        \State \Call{Admit}{$r$} for each newly admissible $r$
    \EndIf
    \State $R \gets$ sequences that need recomputation
    \If{$R \neq \emptyset$}
        \State Recompute KV states at each $r \in R$'s positions awaiting recomputation, via its read-slot map
        \State \textbf{continue} \Comment{skip decode this step}
    \Else
        \State Decode one step for the active sequences
    \EndIf
\EndWhile
\EndProcedure
\end{algorithmic}
\hrule height 1pt
\end{minipage}
\end{wrapfigure}
On admission, the runtime selects a request's Anchor Tokens by mask-to-affix importance and allocates private slots whose positions start out awaiting recomputation (Algorithm~\ref{alg:request-lifecycle}, \textsc{Admit}). During KV recomputation, the runtime forms queries only for those positions, reading keys and values through the read-slot map. The full-context view seen by attention is unchanged, but the query work shrinks, as Section~\ref{sec:system-efficiency} quantifies. During decoding, queries are instead formed only for the request's currently active region: the resulting KV states are written to the private slots already assigned by the read-slot map, and attention again reads the full-context view through the same map.

\begin{figure*}[t]
\centering
\includegraphics[width=0.994\textwidth]{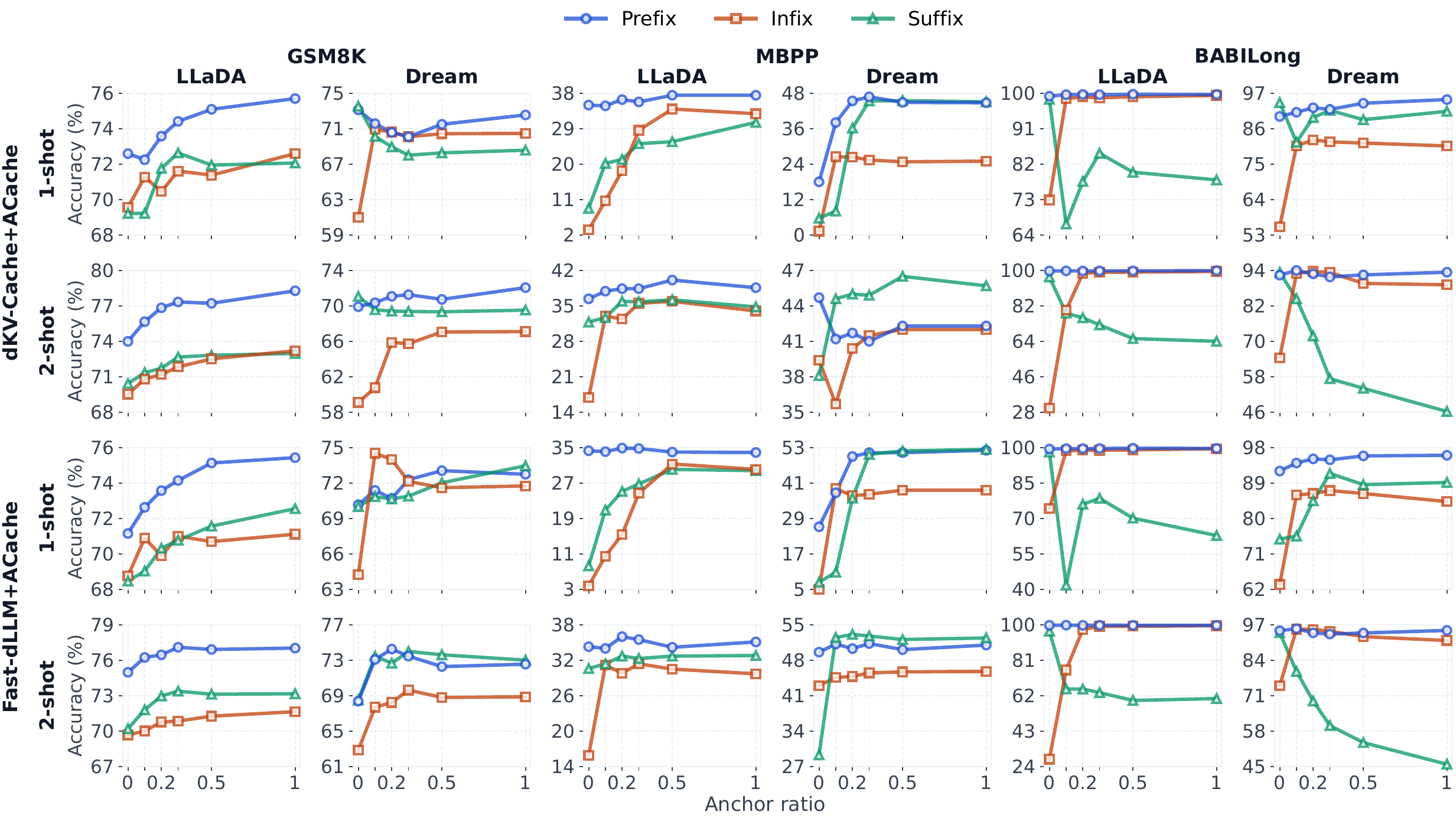}
\caption{\textbf{Accuracy with ACache.} Accuracy increases with the Anchor ratio (0: direct reuse; 1: full recomputation), except BABILong suffix, which declines overall.
}
\label{fig:affix-accuracy}
\end{figure*}

Recomputation points are per-request and unsynchronized, so sequences in a batch reach them at different steps; that is what makes it worth scheduling at step granularity. Algorithm~\ref{alg:request-lifecycle} (\textsc{InferenceLoop}) puts recomputation ahead of the next decode step, a continuous batching pattern that lets a request finishing recomputation join the very next decode batch, in the spirit of ORCA and vLLM-Omni~\citep{yu2022orca, vllm-omni2026}. Appendix~\ref{sec:appendix-continuous-batching} describes how each underlying caching mechanism refills the batch around this loop.
\setlength{\columnsep}{10pt}

%% file: sections/5_Evaluation.tex
\section{Evaluation}
\label{sec:evaluation}

We test the three claims ACache's case rests on: that recomputing a small affix subset restores most of the accuracy direct reuse loses; that the recomputation it saves survives into end-to-end gains; and that its choice of what to recompute holds up against the closest baselines at a matched budget.

We describe the experimental setup as follows:
\begin{smartlist}
\item \textbf{Datasets.} We evaluate math reasoning on GSM8K~\citep{cobbe2021gsm8k}, code generation on MBPP~\citep{austin2021mbpp}, and context retrieval on BABILong-\texttt{0k}/\texttt{qa1}~\citep{kuratov2024babilong}.

\item \textbf{Models.} Following prior work~\citep{kong2025localleap, liu2026dllm-cache, ma2025dkv-cache, song2026sparse-dllm, jiang2026d2cache, luo2026dsb, luo2026dawn}, we evaluate two DLLMs: LLaDA-8B-Instruct (LLaDA)~\citep{nie2025llada} and Dream-v0-Instruct-7B (Dream)~\citep{ye2025dream}.

\item \textbf{Hardware.} We conduct all experiments on NVIDIA A100-SXM4-40GB GPUs.

\item \textbf{Frameworks.} Quality experiments use Hugging Face Transformers; efficiency experiments use the system described in Section~\ref{sec:system-design}. Both place affixes at any position exactly (Appendix~\ref{sec:appendix-experimental-setup}), so the stale-cache effect studied below is the only approximation.

\item \textbf{Default settings.} We use a generation length of 256 and a block length of 32 tokens, following prior work~\citep{jiang2026d2cache, xiao2026streaming-dllm, luo2026dsb, luo2026dawn, kasa2026fast-dllm-pp, lee2026polestar}, and 8 tokens for both on BABILong, to accommodate its short answers. Parallel decoding uses Fast-dLLM's confidence threshold of 0.9~\citep{wu2026fast-dllm}. Quality evaluation sweeps Anchor ratios in $\{0,0.1,0.2,0.3,0.5,1.0\}$. All results are averaged over two runs.

\item \textbf{Affix construction.} For each task and shot count, we place the same examples before the query (prefix), between the query header and answer prompt (infix), or after the masked generation region (suffix). Appendix~\ref{sec:prompt-layouts} gives the templates; Appendix~\ref{sec:appendix-system-prompt} reuses a system prompt.
\end{smartlist}

\makeatletter
\newcommand{\EvalTableFont}{%
  \@setfontsize\EvalTableFont{8}{9}%
  \setlength{\aboverulesep}{0.4ex}%
  \setlength{\belowrulesep}{0.65ex}%
  \setlength{\heavyrulewidth}{0.08em}%
  \setlength{\lightrulewidth}{0.05em}%
  \setlength{\cmidrulewidth}{0.03em}%
}
\makeatother

\newcommand{\SystemPerfHead}[1]{\makebox[0.6in][c]{#1}}
\newcommand{\SystemPerfCell}[1]{\makebox[0.6in][c]{#1}}
\newcommand{\SystemPerfSpeedup}[1]{($#1\times$)}
\newcommand{\SystemPerfGain}[1]{\textcolor{ACacheDarkGreen}{($#1\times$)}}
\newcommand{\RecomputePerfHead}[1]{\makebox[0.65in][c]{#1}}
\newcommand{\RecomputePerfCell}[1]{\makebox[0.65in][c]{#1}}
\newcommand{\RecomputePerfDrop}[1]{\textcolor{ACacheDarkGreen}{($#1\%\downarrow$)}}
\newcommand{\RecomputePerfRise}[1]{($#1\%\uparrow$)}
\newcolumntype{L}[1]{>{\raggedright\arraybackslash}p{#1}}
\newcolumntype{Z}[1]{>{\centering\arraybackslash}p{#1}}

\newlength{\GridColModel}
\newlength{\GridColDataset}
\newlength{\GridColBatch}
\setlength{\GridColModel}{0.4in}
\setlength{\GridColDataset}{0.4in}
\setlength{\GridColBatch}{0.3in}

\begin{table*}[t]
\centering
\caption{\textbf{LLaDA recompute latency (ms) under prefix reuse.} Dream: Table~\ref{tab:dream-system-recompute-grid}.}
\label{tab:system-recompute-grid}
\EvalTableFont
\setlength{\tabcolsep}{2.5pt}
\begin{tabular}{L{\GridColDataset}Z{\GridColBatch}rrrrrr}
\toprule
\TableHeadMultirow{Dataset} & \TableHeadMultirow{Batch} & \multicolumn{2}{c}{1-shot} & \multicolumn{2}{c}{2-shot} & \multicolumn{2}{c}{4-shot} \\
\cmidrule(lr){3-4}\cmidrule(lr){5-6}\cmidrule(lr){7-8}
 & & \RecomputePerfHead{Fast-dLLM} & \RecomputePerfHead{ACache} & \RecomputePerfHead{Fast-dLLM} & \RecomputePerfHead{ACache} & \RecomputePerfHead{Fast-dLLM} & \RecomputePerfHead{ACache} \\
\midrule
\multirow[c]{3}{*}{GSM8K} & 1 & \RecomputePerfCell{56.3} & \RecomputePerfCell{38.9 \RecomputePerfDrop{30.8}} & \RecomputePerfCell{91.2} & \RecomputePerfCell{46.1 \RecomputePerfDrop{49.5}} & \RecomputePerfCell{124.7} & \RecomputePerfCell{60.0 \RecomputePerfDrop{51.9}} \\
 & 4 & \RecomputePerfCell{62.8} & \RecomputePerfCell{43.5 \RecomputePerfDrop{30.7}} & \RecomputePerfCell{100.6} & \RecomputePerfCell{52.3 \RecomputePerfDrop{48.0}} & \RecomputePerfCell{144.1} & \RecomputePerfCell{65.5 \RecomputePerfDrop{54.5}} \\
 & 16 & \RecomputePerfCell{87.9} & \RecomputePerfCell{61.9 \RecomputePerfDrop{29.5}} & \RecomputePerfCell{147.2} & \RecomputePerfCell{76.9 \RecomputePerfDrop{47.8}} & \RecomputePerfCell{213.9} & \RecomputePerfCell{92.6 \RecomputePerfDrop{56.7}} \\
\midrule
\multirow[c]{3}{*}{MBPP} & 1 & \RecomputePerfCell{58.2} & \RecomputePerfCell{44.3 \RecomputePerfDrop{23.8}} & \RecomputePerfCell{91.7} & \RecomputePerfCell{50.6 \RecomputePerfDrop{44.8}} & \RecomputePerfCell{133.0} & \RecomputePerfCell{65.7 \RecomputePerfDrop{50.6}} \\
 & 4 & \RecomputePerfCell{61.7} & \RecomputePerfCell{47.1 \RecomputePerfDrop{23.6}} & \RecomputePerfCell{96.8} & \RecomputePerfCell{54.4 \RecomputePerfDrop{43.8}} & \RecomputePerfCell{145.9} & \RecomputePerfCell{69.9 \RecomputePerfDrop{52.1}} \\
 & 16 & \RecomputePerfCell{75.1} & \RecomputePerfCell{59.2 \RecomputePerfDrop{21.1}} & \RecomputePerfCell{112.2} & \RecomputePerfCell{65.2 \RecomputePerfDrop{41.9}} & \RecomputePerfCell{176.6} & \RecomputePerfCell{82.8 \RecomputePerfDrop{53.1}} \\
\bottomrule
\end{tabular}
\end{table*}

\begin{table*}[t]
\centering
\caption{\textbf{End-to-end throughput under prefix reuse.} Measured in generated tokens per second.}
\label{tab:system-throughput-grid}
\EvalTableFont
\setlength{\tabcolsep}{2.5pt}
\begin{tabular}{L{\GridColModel}L{\GridColDataset}Z{\GridColBatch}rrrrrr}
\toprule
\TableHeadMultirow{Model} & \TableHeadMultirow{Dataset} & \TableHeadMultirow{Batch} & \multicolumn{2}{c}{1-shot} & \multicolumn{2}{c}{2-shot} & \multicolumn{2}{c}{4-shot} \\
\cmidrule(lr){4-5}\cmidrule(lr){6-7}\cmidrule(lr){8-9}
 & & & \SystemPerfHead{Fast-dLLM} & \SystemPerfHead{ACache} & \SystemPerfHead{Fast-dLLM} & \SystemPerfHead{ACache} & \SystemPerfHead{Fast-dLLM} & \SystemPerfHead{ACache} \\
\midrule
\TableModelMultirow{LLaDA} & \multirow[c]{3}{*}{GSM8K} & 1 & \SystemPerfCell{104} & \SystemPerfCell{99 \SystemPerfSpeedup{0.96}} & \SystemPerfCell{94} & \SystemPerfCell{98 \SystemPerfGain{1.04}} & \SystemPerfCell{85} & \SystemPerfCell{92 \SystemPerfGain{1.09}} \\
 & & 4 & \SystemPerfCell{263} & \SystemPerfCell{274 \SystemPerfGain{1.04}} & \SystemPerfCell{206} & \SystemPerfCell{253 \SystemPerfGain{1.23}} & \SystemPerfCell{166} & \SystemPerfCell{229 \SystemPerfGain{1.37}} \\
 & & 16 & \SystemPerfCell{394} & \SystemPerfCell{433 \SystemPerfGain{1.10}} & \SystemPerfCell{278} & \SystemPerfCell{378 \SystemPerfGain{1.36}} & \SystemPerfCell{204} & \SystemPerfCell{329 \SystemPerfGain{1.61}} \\
\cmidrule(lr){2-9}
 & \multirow[c]{3}{*}{MBPP} & 1 & \SystemPerfCell{209} & \SystemPerfCell{213 \SystemPerfGain{1.02}} & \SystemPerfCell{168} & \SystemPerfCell{194 \SystemPerfGain{1.16}} & \SystemPerfCell{123} & \SystemPerfCell{155 \SystemPerfGain{1.26}} \\
 & & 4 & \SystemPerfCell{380} & \SystemPerfCell{401 \SystemPerfGain{1.06}} & \SystemPerfCell{272} & \SystemPerfCell{366 \SystemPerfGain{1.34}} & \SystemPerfCell{189} & \SystemPerfCell{289 \SystemPerfGain{1.53}} \\
 & & 16 & \SystemPerfCell{472} & \SystemPerfCell{507 \SystemPerfGain{1.07}} & \SystemPerfCell{325} & \SystemPerfCell{446 \SystemPerfGain{1.37}} & \SystemPerfCell{210} & \SystemPerfCell{353 \SystemPerfGain{1.68}} \\
\midrule
\TableModelMultirow{Dream} & \multirow[c]{3}{*}{GSM8K} & 1 & \SystemPerfCell{129} & \SystemPerfCell{125 \SystemPerfSpeedup{0.97}} & \SystemPerfCell{120} & \SystemPerfCell{120 \SystemPerfSpeedup{1.00}} & \SystemPerfCell{103} & \SystemPerfCell{117 \SystemPerfGain{1.13}} \\
 & & 4 & \SystemPerfCell{311} & \SystemPerfCell{322 \SystemPerfGain{1.04}} & \SystemPerfCell{255} & \SystemPerfCell{298 \SystemPerfGain{1.17}} & \SystemPerfCell{195} & \SystemPerfCell{276 \SystemPerfGain{1.42}} \\
 & & 16 & \SystemPerfCell{444} & \SystemPerfCell{486 \SystemPerfGain{1.09}} & \SystemPerfCell{323} & \SystemPerfCell{434 \SystemPerfGain{1.34}} & \SystemPerfCell{231} & \SystemPerfCell{376 \SystemPerfGain{1.63}} \\
\cmidrule(lr){2-9}
 & \multirow[c]{3}{*}{MBPP} & 1 & \SystemPerfCell{195} & \SystemPerfCell{190 \SystemPerfSpeedup{0.98}} & \SystemPerfCell{175} & \SystemPerfCell{183 \SystemPerfGain{1.05}} & \SystemPerfCell{137} & \SystemPerfCell{171 \SystemPerfGain{1.25}} \\
 & & 4 & \SystemPerfCell{402} & \SystemPerfCell{410 \SystemPerfGain{1.02}} & \SystemPerfCell{320} & \SystemPerfCell{381 \SystemPerfGain{1.19}} & \SystemPerfCell{217} & \SystemPerfCell{331 \SystemPerfGain{1.53}} \\
 & & 16 & \SystemPerfCell{520} & \SystemPerfCell{553 \SystemPerfGain{1.06}} & \SystemPerfCell{392} & \SystemPerfCell{504 \SystemPerfGain{1.29}} & \SystemPerfCell{246} & \SystemPerfCell{421 \SystemPerfGain{1.71}} \\
\bottomrule
\end{tabular}
\end{table*}

\subsection{Accuracy Under Shared Affixes}
\label{sec:accuracy-shared-affixes}

Figure~\ref{fig:affix-accuracy} plots accuracy as the Anchor ratio varies from 0 (direct reuse) to 1 (full recomputation) for prefix, infix, and suffix reuse. We omit 4-shot affixes, the longest we use, from the figure: in our runs, they match full recomputation at a lower Anchor ratio than the 1- and 2-shot settings shown. Direct reuse is often inaccurate, confirming that a DLLM KV cache for shared text needs request-specific adaptation to remain accurate across requests. Affix position also affects accuracy, consistent with the finding that query position is a first-order variable in bidirectional DLLM decoding~\citep{li2026positional-bias}. BABILong suffix goes further: accuracy declines overall as recomputation increases, unlike every other position-dataset pair. We attribute this reversal to BABILong's answer format: the answer is a bare location, and the few-shot examples end in locations of their own; under suffix placement these sit right next to the masked generation and can interfere with the model's answer. We exclude it from the averages below, which would otherwise count the reversal as recovered accuracy.

The remaining loss depends on dataset and position jointly: averaged over models, shot counts, and frameworks, direct reuse costs 1.73--5.25 points on GSM8K across positions, while MBPP rises from 7.10 at prefix to 18.68 at infix and 20.45 at suffix, and BABILong from 1.40 at prefix to 35.03 at infix. As the ratio increases, ACache restores accuracy across settings; ratio 0.2 recovers about 87\% of the accuracy lost to direct reuse. By shot count, ratio 0.2 raises 1-shot accuracy from 51.11\% to 62.21\%, vs. 64.88\% of full recomputation; in 2-shot settings direct reuse starts higher at 56.96\%, and ratio 0.2 nearly matches full recomputation (66.04\% vs. 66.28\%). Few-shot examples become more redundant, so accuracy becomes less sensitive to the Anchor ratio and to the selection criterion.

\newcommand{\MemoryPerfHead}[1]{\makebox[0.65in][c]{#1}}
\newcommand{\MemoryPerfCell}[1]{\makebox[0.65in][c]{#1}}
\newcommand{\MemoryPerfSame}[1]{($#1\%$)}
\newcommand{\MemoryPerfDrop}[1]{\textcolor{ACacheDarkGreen}{($#1\%\downarrow$)}}
\newcommand{\MemoryPerfRise}[1]{($#1\%\uparrow$)}

\begin{table*}[t]
\centering
\caption{\textbf{LLaDA peak KV cache memory usage (GB) under prefix reuse.} Dream: Table~\ref{tab:dream-peak-kv-memory-grid}.}
\label{tab:peak-kv-memory-grid}
\EvalTableFont
\setlength{\tabcolsep}{2.5pt}
\begin{tabular}{L{\GridColDataset}Z{\GridColBatch}rrrrrr}
\toprule
\TableHeadMultirow{Dataset} & \TableHeadMultirow{Batch} & \multicolumn{2}{c}{1-shot} & \multicolumn{2}{c}{2-shot} & \multicolumn{2}{c}{4-shot} \\
\cmidrule(lr){3-4}\cmidrule(lr){5-6}\cmidrule(lr){7-8}
 & & \MemoryPerfHead{Fast-dLLM} & \MemoryPerfHead{ACache} & \MemoryPerfHead{Fast-dLLM} & \MemoryPerfHead{ACache} & \MemoryPerfHead{Fast-dLLM} & \MemoryPerfHead{ACache} \\
\midrule
\multirow[c]{3}{*}{GSM8K} & 1 & \MemoryPerfCell{0.38} & \MemoryPerfCell{0.38 \MemoryPerfSame{0.0}} & \MemoryPerfCell{0.50} & \MemoryPerfCell{0.75 \MemoryPerfRise{50.0}} & \MemoryPerfCell{0.75} & \MemoryPerfCell{0.88 \MemoryPerfRise{16.7}} \\
 & 4 & \MemoryPerfCell{1.50} & \MemoryPerfCell{1.12 \MemoryPerfDrop{25.0}} & \MemoryPerfCell{2.00} & \MemoryPerfCell{1.50 \MemoryPerfDrop{25.0}} & \MemoryPerfCell{3.00} & \MemoryPerfCell{2.00 \MemoryPerfDrop{33.3}} \\
 & 16 & \MemoryPerfCell{6.00} & \MemoryPerfCell{4.12 \MemoryPerfDrop{31.2}} & \MemoryPerfCell{8.00} & \MemoryPerfCell{4.62 \MemoryPerfDrop{42.2}} & \MemoryPerfCell{12.00} & \MemoryPerfCell{6.50 \MemoryPerfDrop{45.8}} \\
\midrule
\multirow[c]{3}{*}{MBPP} & 1 & \MemoryPerfCell{2.12} & \MemoryPerfCell{2.12 \MemoryPerfSame{0.0}} & \MemoryPerfCell{2.25} & \MemoryPerfCell{2.25 \MemoryPerfSame{0.0}} & \MemoryPerfCell{2.50} & \MemoryPerfCell{2.62 \MemoryPerfRise{5.0}} \\
 & 4 & \MemoryPerfCell{3.25} & \MemoryPerfCell{3.00 \MemoryPerfDrop{7.7}} & \MemoryPerfCell{3.75} & \MemoryPerfCell{3.25 \MemoryPerfDrop{13.3}} & \MemoryPerfCell{4.75} & \MemoryPerfCell{3.75 \MemoryPerfDrop{21.1}} \\
 & 16 & \MemoryPerfCell{8.12} & \MemoryPerfCell{6.25 \MemoryPerfDrop{23.1}} & \MemoryPerfCell{10.12} & \MemoryPerfCell{6.50 \MemoryPerfDrop{35.8}} & \MemoryPerfCell{13.75} & \MemoryPerfCell{8.25 \MemoryPerfDrop{40.0}} \\
\bottomrule
\end{tabular}
\end{table*}

\subsection{System Efficiency}
\label{sec:system-efficiency}

Here we fix the Anchor ratio at 0.2, which on average recovers most of the lost accuracy at 1 shot and matches full recomputation at 2 and 4 shots. We use Fast-dLLM, the more efficient of the two caching frameworks; Appendix~\ref{sec:appendix-dkv-system-efficiency} reports dKV-Cache. We focus on prefix reuse; Appendix~\ref{sec:appendix-infix-suffix-system-efficiency} reports infix and suffix, with similar results. We omit BABILong for its short generation length. Table~\ref{tab:system-recompute-grid} isolates ACache's direct effect on LLaDA recomputation. ACache reduces recompute latency in every setting, by 21.1\% to 56.7\%. Table~\ref{tab:system-throughput-grid} reports end-to-end throughput. At batch size 1 with the shortest, 1-shot affix, throughput ranges from $0.96\times$ to $1.02\times$: Anchor selection overhead is least amortized there, and Appendix~\ref{sec:appendix-anchor-selection-overhead} breaks it down. We include batch size 1 to bound that cost, though it is not the regime a cross-request mechanism targets. Recomputation savings outweigh the overhead in every other setting, reaching up to $1.71\times$ at 4-shot, where ratio 0.2 matches full recomputation.

Table~\ref{tab:peak-kv-memory-grid} reports LLaDA peak KV cache memory for the same settings. ACache reduces peak KV cache usage as batch size and shared-affix length increase, reaching the largest reductions at batch size 16 and 4-shot: on GSM8K from 12.00 GB to 6.50 GB, a 45.8\% drop, and on MBPP from 13.75 GB to 8.25 GB, a 40.0\% drop. These savings come from the read-slot map, which keeps one copy of the affix cache where the baseline, like a per-request blended copy (Section~\ref{sec:system-design}), keeps one per request, leaving headroom for more concurrent requests. At batch size 1, ACache uses up to 0.25 GB more, as shared and private blocks are allocated separately at block granularity.

\newlength{\AnchorColModel}
\newlength{\AnchorColRatio}
\newlength{\AnchorColMethod}
\newlength{\AnchorColData}
\setlength{\AnchorColModel}{0.4in}
\setlength{\AnchorColRatio}{0.25in}
\setlength{\AnchorColMethod}{0.72in}
\setlength{\AnchorColData}{0.3in}
\begin{table*}[t]
\centering
\caption{\textbf{ACache's criterion compared to adapted baselines, 1-shot.} Accuracy (\%) on Fast-dLLM. Bold marks the best value per column; underline, the second-best. BiCache$^\dagger$ is the official implementation; Ratio is the mean equivalent Anchor ratio. Avg.\ omits BABILong suffix (Section~\ref{sec:accuracy-shared-affixes}).}
\label{tab:anchor-selection-comparison}
\EvalTableFont
\setlength{\tabcolsep}{3pt}
\begin{tabular}{L{\AnchorColModel}L{\AnchorColRatio}L{\AnchorColMethod}*{9}{Z{\AnchorColData}}Z{\AnchorColData}}
\toprule
\TableHeadMultirow{Model} & \TableHeadMultirow{Ratio} & \TableHeadMultirow{Criterion} & \multicolumn{3}{c}{GSM8K} & \multicolumn{3}{c}{MBPP} & \multicolumn{3}{c}{BABILong} & \TableHeadMultirow{Avg.} \\
\cmidrule(lr){4-6}\cmidrule(lr){7-9}\cmidrule(lr){10-12}
 & & & Pfx. & Inf. & Suf. & Pfx. & Inf. & Suf. & Pfx. & Inf. & Suf. & \\
\midrule
\multirow[c]{9}{*}{\raisebox{-0.35ex}{LLaDA}} & \multirow[c]{4}{*}{0.2} & CacheBlend & 71.6 & 69.6 & 69.8 & \textbf{35.5} & 7.8 & 22.0 & 99.5 & 82.8 & 63.6 & 57.3 \\
 & & ProphetKV & \textbf{75.6} & \underline{69.8} & \textbf{70.8} & 33.5 & \underline{13.1} & \textbf{25.8} & \textbf{99.7} & \underline{98.6} & 43.4 & \underline{60.9} \\
 & & BiCache & 71.4 & 69.1 & 68.2 & 33.2 & 3.4 & 8.4 & 99.4 & 73.9 & \textbf{98.4} & 53.4 \\
 & & ACache & \underline{73.6} & \textbf{69.9} & \underline{70.3} & \underline{34.9} & \textbf{15.4} & \underline{25.1} & \underline{99.6} & \textbf{98.9} & \underline{76.1} & \textbf{61.0} \\
\cmidrule(lr){2-13}
 & \multirow[c]{4}{*}{0.3} & CacheBlend & 71.9 & 70.4 & \underline{70.9} & \textbf{35.6} & 12.3 & 15.2 & \underline{99.4} & 94.7 & 27.8 & \underline{58.8} \\
 & & ProphetKV & \textbf{75.3} & \underline{70.5} & \textbf{72.1} & \underline{34.9} & \underline{23.0} & \textbf{27.0} & \underline{99.4} & \underline{98.8} & 43.8 & \textbf{62.6} \\
 & & BiCache & 71.0 & 67.9 & 67.6 & 33.8 & 3.1 & 10.3 & \underline{99.4} & 68.1 & \textbf{98.2} & 52.6 \\
 & & ACache & \underline{74.1} & \textbf{71.0} & 70.8 & 34.8 & \textbf{24.7} & \underline{26.8} & \textbf{99.7} & \textbf{98.9} & \underline{78.6} & \textbf{62.6} \\
\cmidrule(lr){2-13}
 & 0.67 & BiCache$^\dagger$ & 76.1 & -- & -- & 34.6 & -- & -- & 99.5 & -- & -- & -- \\
\cmidrule(lr){1-13}
\multirow[c]{9}{*}{\raisebox{-0.35ex}{Dream}} & \multirow[c]{4}{*}{0.2} & CacheBlend & \textbf{73.4} & 68.2 & \textbf{73.5} & 38.6 & 4.8 & 4.9 & \textbf{96.4} & \underline{84.3} & 80.5 & 55.5 \\
 & & ProphetKV & 70.4 & \underline{73.5} & 69.3 & \underline{49.9} & \underline{32.3} & \underline{27.4} & \underline{95.9} & 82.3 & \textbf{91.5} & \underline{62.6} \\
 & & BiCache & \underline{70.9} & 64.3 & \underline{71.4} & 26.8 & 4.8 & 7.0 & 92.1 & 55.0 & 77.2 & 49.0 \\
 & & ACache & 70.7 & \textbf{74.0} & 70.7 & \textbf{50.0} & \textbf{36.7} & \textbf{35.9} & 95.2 & \textbf{86.4} & \underline{84.5} & \textbf{64.9} \\
\cmidrule(lr){2-13}
 & \multirow[c]{4}{*}{0.3} & CacheBlend & \underline{73.5} & 71.1 & \textbf{73.6} & \textbf{52.2} & 14.4 & 18.0 & \textbf{95.6} & \textbf{87.1} & \underline{86.3} & 60.7 \\
 & & ProphetKV & 72.2 & \underline{71.6} & \underline{71.1} & 49.8 & \textbf{37.7} & \underline{39.3} & \underline{95.3} & \underline{83.4} & 83.4 & \underline{65.0} \\
 & & BiCache & \textbf{73.7} & 66.1 & 70.5 & 41.1 & 3.5 & 9.6 & 90.6 & 45.6 & 77.5 & 50.1 \\
 & & ACache & 72.3 & \textbf{72.1} & 70.9 & \underline{51.3} & \underline{37.2} & \textbf{50.7} & 94.9 & \textbf{87.1} & \textbf{91.5} & \textbf{67.1} \\
\cmidrule(lr){2-13}
 & 0.77 & BiCache$^\dagger$ & 71.4 & -- & -- & 51.9 & -- & -- & 91.0 & -- & -- & -- \\
\bottomrule
\end{tabular}
\end{table*}

\newlength{\BiColModel}
\newlength{\BiColDataset}
\newlength{\BiColData}
\setlength{\BiColModel}{0.4in}
\setlength{\BiColDataset}{0.4in}
\setlength{\BiColData}{0.6in}
\begin{table*}[t]
\centering
\caption{\textbf{End-to-end throughput at batch size 1 under prefix reuse (tokens/s).} Both run on Fast-dLLM: BiCache from its official implementation (Appendix~\ref{sec:appendix-baseline-selector-adaptations}), ACache at Anchor ratio 0.2.}
\label{tab:bicache-throughput-grid}
\EvalTableFont
\setlength{\tabcolsep}{2.5pt}
\begin{tabular}{L{\BiColModel}L{\BiColDataset}*{6}{Z{\BiColData}}}
\toprule
\TableHeadMultirow{Model} & \TableHeadMultirow{Dataset} & \multicolumn{2}{c}{1-shot} & \multicolumn{2}{c}{2-shot} & \multicolumn{2}{c}{4-shot} \\
\cmidrule(lr){3-4}\cmidrule(lr){5-6}\cmidrule(lr){7-8}
 & & BiCache & ACache & BiCache & ACache & BiCache & ACache \\
\midrule
\multirow[c]{2}{*}{LLaDA} & GSM8K & 96 & 99 \SystemPerfGain{1.03} & 84 & 98 \SystemPerfGain{1.18} & 71 & 92 \SystemPerfGain{1.30} \\
 & MBPP & 192 & 213 \SystemPerfGain{1.11} & 165 & 194 \SystemPerfGain{1.18} & 118 & 155 \SystemPerfGain{1.31} \\
\midrule
\multirow[c]{2}{*}{Dream} & GSM8K & 87 & 125 \SystemPerfGain{1.44} & 93 & 120 \SystemPerfGain{1.30} & 83 & 117 \SystemPerfGain{1.41} \\
 & MBPP & 145 & 190 \SystemPerfGain{1.31} & 136 & 183 \SystemPerfGain{1.35} & 119 & 171 \SystemPerfGain{1.44} \\
\bottomrule
\end{tabular}
\end{table*}

\subsection{Comparison to Closest Baselines}
\label{sec:anchor-selection-comparison}

None of the three systems closest to ACache applies directly to DLLM affix reuse. CacheBlend~\citep{yao2025cache-blend} and ProphetKV~\citep{wang2026prophetkv} are autoregressive: both recompute once during prefill and overwrite the selected entries in a per-request copy of the cache. A DLLM rewrites affix KV at every recomputation point, so it would repeat that overwrite throughout decoding, each request holding its own copy of the shared affix cache (Section~\ref{sec:system-design}). BiCache~\citep{go2026bicache} targets DLLMs, but only shared prefixes, and reads reuse depth from an offline profile rather than a budget we can match. We therefore compare criteria, not whole systems, porting each into ACache's pipeline at a matched Anchor ratio $\rho$. CacheBlend scores tokens by KV deviation, ProphetKV by request-prompt attention, and BiCache recomputes the affix in the last $\lceil \rho N \rceil$ of $N$ layers (Appendix~\ref{sec:appendix-baseline-selector-adaptations}).

Table~\ref{tab:anchor-selection-comparison} reports this comparison at Anchor ratios 0.2 and 0.3 in the 1-shot setting, where accuracy is most sensitive to the selection criterion (Section~\ref{sec:accuracy-shared-affixes}). ACache leads the next-best criterion, ProphetKV, by 2.0 to 2.3 points of average accuracy on Dream and matches it on LLaDA, within 0.1 points; it leads CacheBlend, likewise token-level, by 3.7 to 9.4. Which tokens are recomputed thus matters at a fixed budget, and mask-to-affix attention importance is a well-suited criterion for DLLM affix reuse, with the advantage most visible on infix and suffix, where direct reuse costs the most accuracy. Appendix~\ref{sec:appendix-random-bottom-selection} further swaps ACache's ranking for random and bottom selection. BiCache places last on both models, at both ratios. Averaged over the three prefix settings, its official implementation reaches 70.1\% on LLaDA and 71.4\% on Dream, against ACache's 69.4\% and 72.0\% at ratio 0.2, but at a mean equivalent Anchor ratio of 0.67 and 0.77. Whole layers thus seem a coarser unit than tokens for preserving request-specific information, though a criterion could select on both.

For system efficiency, the four matched criteria all spend the same recomputation budget on every request that shares an affix. BiCache's profiled depth, as originally designed, instead varies with each request's shared-prefix fraction, so recomputation cost differs across a batch---which is hard for a continuous batching loop to absorb~\citep{yu2022orca}. Table~\ref{tab:bicache-throughput-grid} compares official BiCache with ACache at batch size 1, the only one BiCache's open-source engine supports: ACache is faster in all twelve settings, by $1.03\times$ to $1.44\times$. There the advantage comes instead from the recomputation ACache saves and from more efficient system support; ACache also batches requests, widening the gap.

%% file: sections/6_Conclusion.tex
\section{Conclusion}

We presented ACache, in which one Anchor ratio governs how much of a shared span's KV cache a DLLM reuses, and one copy serves every request whether it places that span as a prefix, infix, or suffix. Mask-to-affix attention importance concentrates on a small, stable subset of affix tokens, so recomputing around 20\% of them---the Anchor Tokens---recovers most of the accuracy direct reuse loses, and at a matched budget this criterion selects better than those adapted from the closest baselines. The recomputation and memory ACache saves grow with both batch size and shared-span length, so the gains are largest in exactly the regime cross-request reuse targets. In the future, we hope the affix can do for DLLM inference stacks what cross-request reuse has done for AR ones.

%% file: sections/7_Statements.tex
\subsection*{AI Use Statement}

In this work, we used generative AI tools to assist with implementing parts of the codebase, including evaluation scripts. We have not used generative AI tools to propose or refine hypotheses, design research methodology or experiments. Additionally, we used generative AI tools to assist with polishing the manuscript for wording, grammar, and clarity, and to support literature search during related-work review. We have reviewed all AI-assisted work and take responsibility for the final content of this work, including text, claims, and artifacts produced with the aid of generative AI.

\subsection*{Reproducibility Statement}

We describe the ACache design and Anchor Token selection procedure in Section~\ref{sec:design}, and the system design and implementation details in Section~\ref{sec:system-design} and Appendix~\ref{sec:appendix-system-design}. Evaluation settings are reported under ``Default settings'' in Section~\ref{sec:evaluation}; exact prompt templates for each affix position are given in Appendix~\ref{sec:prompt-layouts}. All datasets (GSM8K, MBPP, BABILong) and base models (LLaDA, Dream) are publicly available, and all measurements were taken on the GPUs reported under ``Hardware'' in Section~\ref{sec:evaluation}. Because ACache adds a custom Triton attention kernel, the code pins its package versions and documents the PyTorch, CUDA, and Triton installation it expects. The ACache implementation, including the Nano-vLLM-based system and the scripts used to reproduce our experiments, \ifarxiv is publicly available at \url{https://github.com/sands-lab/ACache}.\fi

%% file: sections/8A_Appendix_Discussion.tex
\section{Related Work}
\label{sec:related-work}

\subsection{Development of Diffusion-Based Language Models}
\label{sec:related-work-dllm}

Diffusion-based language models adapt diffusion probabilistic models~\citep{sohl-dickstein2015diffusion, ho2020ddpm} to text by iteratively denoising a corrupted sequence: the whole sequence is laid out from the first step, and predictions at masked positions are conditioned on global context rather than a prefix. This non-causal formulation appears in continuous-embedding models such as Diffusion-LM~\citep{li2022continuous}, and discrete-state models such as D3PM~\citep{austin2021d3pm} and Multinomial Diffusion~\citep{hoogeboom2021multinomial}. Subsequent work such as SEDD~\citep{lou2024sedd}, MDLM~\citep{sahoo2024mdlm}, MD4~\citep{shi2024md4}, and RADD~\citep{ou2025radd} refined this formulation with simpler objectives and stronger training, narrowing the gap to AR models.

\subsection{DLLM Acceleration}

ACache is orthogonal to most DLLM acceleration techniques. Many reduce intra-request cost through cache reuse and recomputation~\citep{ma2025dkv-cache, liu2026dllm-cache, wu2026fast-dllm, hu2026flash-dlm, ma2025dinfer}. Inference systems interleave recomputation and decoding across concurrent requests~\citep{fan2025dllm-serve, kedia2026sangam}, and offload KV state to host memory to retrieve a sparse subset on demand~\citep{kwon2026herald, wang2026groupkv}. Other work shrinks the context each step attends to, pruning distant suffix tokens or compressing the masked region~\citep{zuo2026window-diffusion, xiao2026streaming-dllm, chen2026dpad, wu2026elastic-dllm}, and for long contexts sparsifies attention or filters what it caches~\citep{long2026focus-dllm, xiong2026prefilling-dllm, yang2026wavefilter}. Some adjust the decoding schedule, varying block size or tokens committed per step~\citep{kong2025localleap, lu2026adablock-dllm, luo2026dsb}. Others decide at runtime which cached state merits reuse~\citep{cheong2026entropy-cache, sun2026spa-cache, wu2026dynamic-dllm, frumkin2026dare, he2026archer, lee2026polestar}, or compute only the tokens that still change~\citep{zhu2026es-dllm, lee2026dyllm}. Several read that decision off attention~\citep{huang2026maskkv, song2026sparse-dllm, jiang2026d2cache, nguyen-tri2026attention, liang2026focus, luo2026dawn}. ACache also uses attention, but to select request-specific Anchor Tokens, keeping the rest of the cache (Appendix~\ref{sec:appendix-non-anchor-cache}). ASRD uses the term for another object: generated tokens that stay stable across iterations, used to revoke earlier predictions~\citep{yao2026follow}. ACache targets cross-request reuse; BiCache shares that target but resolves reuse by depth rather than by token, and as originally designed covers prefixes alone~\citep{go2026bicache}. ACache and these intra-request optimizations compose: it exposes reuse across requests, they cut the cost that remains.

\subsection{Inference Abstractions}

In AR LLM inference, prefix-cache reuse has shaped runtime abstractions at every layer of the inference stack: cache layout and prefix indexing~\citep{kwon2023vllm, zheng2024sglang}, prefix-aware attention kernels~\citep{ye2024chunkattention, pan2025fasttree, ye2025flashinfer}, eviction policies~\citep{ouyang2026unicache}, request scheduling and cache affinity~\citep{srivatsa2025preble, yuan2026dualmap}, cluster-level cache pooling~\citep{qin2025mooncake}, and modular reuse abstractions~\citep{gim2024prompt_cache}. CacheBlend studies non-prefix chunk reuse for AR RAG inference~\citep{yao2025cache-blend}, and EPIC formalizes it as position-independent caching~\citep{hu2025epic}; both settle a reused span before generation begins, which a DLLM cannot do (Section~\ref{sec:system-design}). Subsequent AR systems push this direction further, sharing KV at segment granularity across requests~\citep{zhang2026sparsex}, reusing it per attention head~\citep{liu2026redknot}, spending the recomputation budget on the entries the query actually needs~\citep{wang2026prophetkv}, and tuning how much to recompute against hardware cost~\citep{li2026cachetune}. By treating shared context as an affix-level object whose consistency can be managed across requests, ACache points to an inference research direction for DLLMs analogous to prefix caching in AR LLMs, and one where bidirectional modeling may organize shared context more flexibly.

%% file: sections/8B_Appendix_SystemDesign.tex
\section{System Design Details}
\label{sec:appendix-system-design}

\subsection{Kernel Integration}
\label{sec:appendix-kernel-integration}

To support the per-request read-slot map, the system implements a customized Triton~\citep{tillet2019triton} attention kernel for ACache. In addition to the paged KV cache and ragged query/KV sequence lengths, the kernel takes a flattened \texttt{read\_slot\_map}. For each KV tile, it loads the physical slots named by this map before gathering keys and values from the paged cache. Thus logical KV positions can point either to shared affix slots or to request-private slots, while the query tile remains a standard dense tile. Algorithm~\ref{alg:acache-kernel} details the per-tile loop, which keeps the tiled online-softmax accumulation of FlashAttention~\citep{dao2022flashattention} and changes only how each tile is addressed: unlike a standard paged-attention kernel, which resolves each KV tile through a block table (\texttt{block\_id}, \texttt{offset})~\citep{kwon2023vllm}, this kernel gathers keys and values directly by physical slot, so shared and private KV states are read through the same indirection.

FlashInfer already represents KV caches with block-sparse formats, including fine-grained blocks, and composes attention over shared-prefix and request-specific KV by merging their partial attention states through log-sum-exp~\citep{juravsky2024hydragen, ye2025flashinfer}. Its shared-prefix use case does not need to address relocating a cached span across request-specific positions, since the shared prefix occupies the same initial positions in every request. ACache reuses one affix cache even when its reusable slots are interleaved with request-private Anchor slots and the affix begins at different offsets across requests; the read-slot map exposes these slots as one logical KV view, while the positional correction in Algorithm~\ref{alg:rotation-dispatch} lets the shared cache stay unchanged across offsets.

\begin{algorithm}[t]
\caption{ACache attention kernel (one query tile, one head)}
\label{alg:acache-kernel}
\small
\begin{algorithmic}[1]
\Require query tile $Q$; \texttt{read\_slot\_map}; flat per-slot caches $K$, $V$; KV length $\mathit{kv\_len}$
\Ensure attention output $O$ for this tile
\State $m \gets -\infty$, $\ell \gets 0$, $\mathit{acc} \gets 0$ \Comment{online-softmax state}
\For{$t \gets 0$ to $\mathit{kv\_len} - 1$ step $\mathit{BLOCK\_N}$}
    \State $\mathit{slots} \gets \texttt{read\_slot\_map}[t : t + \mathit{BLOCK\_N}]$ \Comment{physical slots for this KV tile}
    \State $K_{\mathrm{tile}} \gets K[\mathit{slots}]$, $V_{\mathrm{tile}} \gets V[\mathit{slots}]$ \Comment{direct slot gather, no block table}
    \State $\mathit{scores} \gets Q \cdot K_{\mathrm{tile}}^\top \cdot \mathit{sm\_scale}$
    \State $m_{\mathrm{new}} \gets \max(m, \mathrm{rowmax}(\mathit{scores}))$
    \State $p \gets \exp(\mathit{scores} - m_{\mathrm{new}})$, $\alpha \gets \exp(m - m_{\mathrm{new}})$
    \State $\ell \gets \alpha \cdot \ell + \mathrm{rowsum}(p)$, $\mathit{acc} \gets \alpha \cdot \mathit{acc} + p \cdot V_{\mathrm{tile}}$
    \State $m \gets m_{\mathrm{new}}$
\EndFor
\State $O \gets \mathit{acc} / \ell$
\end{algorithmic}
\end{algorithm}

The runtime dispatches the shared-affix rotation correction (Section~\ref{sec:system-design}) based on the engine's configured maximum batch size, not on how many sequences a given batch actually holds. Algorithm~\ref{alg:rotation-dispatch} summarizes these cases. In the batched case, the runtime first groups a request's shared slots ahead of its private ones in the read-slot map---exact, since bidirectional attention imposes no order on the KV axis---so every KV tile is either all shared or all private and the counter-rotation applies once to the query tile rather than once per key slot. The other cases need no grouping: a materialized copy is already positioned for the request, and an affix starting at position zero takes no correction at all. A single-sequence request that cannot reserve the workspace needs no grouping either: the kernel rotates the shared keys inline as it reads them, leaving the stored cache untouched.

\begin{algorithm}[t]
\caption{Shared-affix rotation dispatch}
\label{alg:rotation-dispatch}
\small
\begin{algorithmic}[1]
\Require engine configuration $\mathit{max\_num\_seqs}$; a shared affix whose $K$ was rotated for position zero but now starts at offset $\mathit{shift} \neq 0$
\Ensure correctly positioned $K$ for the shared slots, as seen by attention
\If{$\mathit{max\_num\_seqs} \neq 1$} \Comment{configured for batching}
    \State In Algorithm~\ref{alg:acache-kernel}, score the shared-slot tiles with $Q$ rotated by $-\mathit{shift}$ and the remaining tiles with $Q$ \Comment{one rotation, before the KV loop}
\ElsIf{a workspace can be reserved} \Comment{configured for single-sequence inference}
    \State $K' \gets \mathrm{rotate}(K, \mathit{shift})$, $V' \gets V$ \Comment{$V$ carries no position, so it is copied as is}
    \State Write $K', V'$ into the workspace
    \State Redirect \texttt{read\_slot\_map} to the workspace \Comment{Algorithm~\ref{alg:acache-kernel} then runs unmodified}
\Else \Comment{no workspace available}
    \State In Algorithm~\ref{alg:acache-kernel}, rotate the shared keys by $\mathit{shift}$ as each tile is gathered \Comment{stored cache untouched}
\EndIf
\end{algorithmic}
\end{algorithm}

The ACache path preserves the same store-then-attend interface in both recomputation and block decoding. The runtime first writes newly produced KV states through the normal write-side slot mapping, then invokes the read-slot kernel over the effective KV view. The two phases differ only in the query/write slice: recomputation uses the Anchor Tokens and request-specific positions, whereas decoding uses the current block. This keeps ACache's storage indirection inside the attention read path: neither phase materializes a blended copy of the shared affix cache.

\subsection{Continuous Batching Across Caching Mechanisms}
\label{sec:appendix-continuous-batching}

The inference loop of Algorithm~\ref{alg:request-lifecycle} advances each request's currently active region (Section~\ref{sec:system-design}) without prescribing how the underlying intra-request caching mechanism defines it. In this codebase, both supported mechanisms instantiate it as a fixed-size block of \texttt{block\_length} tokens and require a block to fully unmask before the sequence advances to the next one, but they differ in how they support continuous batching around this boundary. Under Fast-dLLM~\citep{wu2026fast-dllm}, the runtime treats every completed block as a checkpoint: it advances each sequence whose block has just unmasked, retires those that have finished their last block, and refills the freed slots from the pending queue within the same pass, so newly admitted or newly advanced requests can join the very next decode step. Under dKV-Cache~\citep{ma2025dkv-cache}, each step commits a fixed number of tokens rather than all those above a confidence threshold, so every sequence in a batch spends the same number of steps on each block. The runtime therefore admits a batch, runs it through all of its blocks, and deallocates it as a whole before admitting the next: requests join and leave only at batch boundaries.

%% file: sections/8C_Appendix_EvaluationDetails.tex
\section{Evaluation Details}
\label{sec:evaluation-details}

\subsection{Experimental Setup}
\label{sec:appendix-experimental-setup}

Quality evaluations use the Language Model Evaluation Harness (\texttt{lm-eval})~\citep{eval-harness} with model-specific wrappers for LLaDA and Dream. On this backend, affix keys are stored before rotary rotation and rotated to the affix's request-time position when read, rather than stored for position zero and corrected on the query side as in the runtime (Section~\ref{sec:system-design}). Both score the affix at its absolute request-time positions, so placement is exact on either backend. We measure accuracy on this backend rather than on the runtime to reduce variance: under continuous batching, a request's output in the runtime can shift with the other requests that share its batch. At batch sizes 1, 4, and 16, runtime accuracy at Anchor ratio 0.2 stays within 2.0 points of the Transformers backend's in all 24 GSM8K and MBPP settings of Figure~\ref{fig:affix-accuracy}, and is 0.1 points higher on average. GSM8K and MBPP use standard task data with the affix serialization in Appendix~\ref{sec:prompt-layouts}. For BABILong, we add a custom task YAML that loads the \texttt{0k}/\texttt{qa1} split of \texttt{RMT-team/babilong-1k-samples} and scores it with exact match.

\subsection{Prompt Construction}
\label{sec:prompt-layouts}

All prompts are serialized with the corresponding model chat template; the templates below show the message contents before tokenizer-specific role and control tokens are added. Let $\mathsf{Ex}$ denote the serialized few-shot examples used as the shared affix, $\mathsf{Qry}$ the task query, $\mathsf{Ans}$ the final prompt used in the infix layout to direct the model back to the initial query, and $\mathsf{Mask}$ the masked generation region. The prefix layout is $\mathsf{Ex} \; \mathsf{Qry} \; \mathsf{Mask}$. The infix layout is $\mathsf{Qry}_{\mathrm{head}} \; \mathsf{Ex} \; \mathsf{Ans} \; \mathsf{Mask}$, where examples are inserted between the query header and this final prompt. The suffix layout is $\mathsf{Qry} \; \mathsf{Mask} \; \mathsf{Ex}$.

\smartparagraph{GSM8K.}
For GSM8K, the query message is:
\begin{quote}
\small\ttfamily
Question: \{question\}\\
Let's think step by step.\\
Answer:
\end{quote}
Few-shot answers are normalized to:
\begin{quote}
\small\ttfamily
\{reasoning\}\\
The answer is \{final\_answer\}
\end{quote}
For infix placement, the query header is:
\begin{quote}
\small\ttfamily
Question: \{question\}\\
\\
Example(s):
\end{quote}
and the final answer prompt is:
\begin{quote}
\small\ttfamily
The examples above are demonstrations only.\\
Let's come back to the question at the beginning and think step by step.\\
Answer:
\end{quote}

\smartparagraph{MBPP.}
For MBPP, the query message is:
\begin{quote}
\small\ttfamily
You are an expert Python programmer, and here is your task: \{question\} Your code should pass these tests:\\
\\
\{test\_1\}\\
\{test\_2\}\\
\{test\_3\}\\
{}[BEGIN]
\end{quote}
Few-shot answers are normalized to:
\begin{quote}
\small\ttfamily
\{code\}\\
{}[DONE]
\end{quote}
For infix placement, the query header replaces the final \texttt{[BEGIN]} marker with a blank line followed by:
\begin{quote}
\small\ttfamily
Example(s):
\end{quote}
and the final answer prompt is:
\begin{quote}
\small\ttfamily
The examples above demonstrate the expected response pattern.\\
Let's come back to the task at the beginning and write the Python code in that pattern.\\
Do not include any explanation, comments outside the code, markdown fences, or test cases.\\
{}[BEGIN]
\end{quote}

\smartparagraph{BABILong-\texttt{0k}/\texttt{qa1}.}
For BABILong, the query message is:
\begin{quote}
\small\ttfamily
Story:\\
\{story\}\\
Question: \{question\}\\
Answer with exactly one location name.\\
Location:
\end{quote}
Few-shot answers are the stripped target location:
\begin{quote}
\small\ttfamily
\{target\}
\end{quote}
For infix placement, the query header is:
\begin{quote}
\small\ttfamily
Story:\\
\{story\}\\
Question: \{question\}\\
\\
Example(s):
\end{quote}
and the final answer prompt is:
\begin{quote}
\small\ttfamily
End of examples.\\
Let's come back to the story and question at the beginning.\\
Answer with exactly one location name.\\
Location:
\end{quote}

\smartparagraph{Few-shot affix serialization.}
Few-shot examples are sampled from the \texttt{main}/\texttt{train} split of \texttt{gsm8k} for GSM8K and the \texttt{full}/\texttt{prompt} split of \texttt{google-research-datasets/mbpp} for MBPP; BABILong has no separate example pool, so its examples come from the same \texttt{0k}/\texttt{qa1} split that is evaluated. In prefix and suffix modes, each sampled few-shot example is serialized as a user query followed by its normalized assistant answer. In infix mode, the shared affix is a plaintext block:
\begin{quote}
\small\ttfamily
Example \{i\}:\\
\{query message\}\\
\{normalized answer\}
\end{quote}
with examples separated by blank lines. The query header, example block, and final answer prompt are joined with blank lines within a single user message.

\subsection{Baseline Criterion Adaptations}
\label{sec:appendix-baseline-selector-adaptations}

Table~\ref{tab:anchor-selection-comparison} matches CacheBlend, ProphetKV, and BiCache to ACache's Anchor ratio $\rho$, so all four criteria spend the same recomputation budget. Each baseline keeps ACache's generation loop and cache layout unchanged, changing only what it recomputes.

CacheBlend and ProphetKV both select at the token level, like ACache: each scores the $L$ affix tokens and recomputes the top-$K = \lceil \rho L \rceil$ by that score, differing only in the scoring signal. CacheBlend~\citep{yao2025cache-blend} scores each affix token by high KV deviation: the $\ell_2$ distance between its key and value entries in the affix-only cache and in the full-prompt cache, summed across layers, without using which masked generation positions attend to it. We obtain that full-prompt cache from an actual prefill at every layer, whereas CacheBlend prefills only the first layer and narrows the token set layer by layer, so this criterion is given information the original cannot obtain. ProphetKV~\citep{wang2026prophetkv} instead scores each affix token by the attention that the request-owned prompt places on it, rather than the masked generation positions ACache uses, averaged across layers, heads, and query positions. Its original criterion scores by attention from the user query, which its retrieval-augmented setting places at the end of the prompt; since a DLLM affix can sit before, inside, or after the request-owned tokens, we anchor the signal to them wherever they fall.

BiCache~\citep{go2026bicache} selects on a different axis: it decides which \emph{layers}, not which tokens, keep a stale cache. We match it to the same ratio $\rho$ by recomputing the affix jointly with the request in the last $\lceil \rho N \rceil$ of the model's $N$ layers, while the remaining, shallower layers reuse the shared affix cache untouched: request tokens alone traverse the shallow layers, and the affix rejoins the sequence only at the boundary layer. This adapts BiCache's layer-selection idea to ACache's ratio-based budget, without BiCache's own offline shared-prefix-ratio lookup or its periodic recompute interval. Where Table~\ref{tab:anchor-selection-comparison} and Table~\ref{tab:bicache-throughput-grid} instead report BiCache's own implementation, we run the released code at commit \texttt{82a8d0122807daf58ceacd6a121201b8355f46c1} of \url{https://github.com/OSSS-KU/BiCache}, changing only its fixed per-step decoding schedule to the 0.9 confidence threshold of Section~\ref{sec:evaluation}. The release supports LLaDA alone; we port it to Dream, reprofiling its layer policy there.

%% file: sections/8D_Appendix_AdditionalEvaluation.tex
\section{Additional Evaluation}
\label{sec:additional-evaluation}

\subsection{Anchor Stability}
\label{sec:appendix-anchor-stability}

\newlength{\StabColData}
\setlength{\StabColData}{0.38in}

\begin{table*}[t]
\centering
\caption{\textbf{Anchor stability at Anchor ratio 0.2.} Fast-dLLM, full recomputation; all values in \%, means over each test set. At each block boundary, a set is compared with the current top-$K$ set: Random, Fresh (the first's), or ACache's Anchors (averaged). Share: the first top-$K$ set's importance.}
\label{tab:anchor-stability}
\EvalTableFont
\setlength{\tabcolsep}{2.5pt}
\begin{tabular}{L{\GridColDataset}L{\GridColDataset}Z{\GridColBatch}L{\AnchorColModel}*{8}{Z{\StabColData}}}
\toprule
\multirow[c]{3}{*}{Dataset} & \multirow[c]{3}{*}{Affix} & \multirow[c]{3}{*}{Shots} & \multirow[c]{3}{*}{Model} & \multirow[c]{3}{*}{Share} & \multicolumn{4}{c}{Jaccard similarity} & \multicolumn{3}{c}{Mass recall} \\
\cmidrule(lr){6-9}\cmidrule(lr){10-12}
 & & & & & \TableHeadMultirow{Random} & \multicolumn{2}{c}{Fresh} & \TableHeadMultirow{ACache} & \multicolumn{2}{c}{Fresh} & \TableHeadMultirow{ACache} \\
\cmidrule(lr){7-8}\cmidrule(lr){10-11}
 & & & & & & 2nd & Last & & 2nd & Last & \\
\midrule
\multirow[c]{12}{*}{GSM8K} & \multirow[c]{4}{*}{Prefix} & \multirow[c]{2}{*}{1} & LLaDA & 62.7 & 11.3 & 71.6 & 56.8 & 38.9 & 98.5 & 95.8 & 76.7 \\
 &  &  & Dream & 71.8 & 11.3 & 78.8 & 57.4 & 54.4 & 97.8 & 89.5 & 89.3 \\
 &  & \multirow[c]{2}{*}{2} & LLaDA & 62.6 & 11.2 & 73.4 & 52.5 & 35.2 & 98.5 & 95.1 & 91.4 \\
 &  &  & Dream & 73.4 & 11.2 & 84.9 & 58.2 & 62.8 & 99.4 & 93.6 & 94.3 \\
\cmidrule(lr){2-12}
 & \multirow[c]{4}{*}{Infix} & \multirow[c]{2}{*}{1} & LLaDA & 51.3 & 11.3 & 48.8 & 54.9 & 33.9 & 91.2 & 92.4 & 84.7 \\
 &  &  & Dream & 62.2 & 11.5 & 68.7 & 69.0 & 64.4 & 94.2 & 95.5 & 94.6 \\
 &  & \multirow[c]{2}{*}{2} & LLaDA & 60.2 & 11.1 & 58.6 & 53.4 & 29.1 & 95.8 & 95.4 & 88.6 \\
 &  &  & Dream & 66.1 & 11.2 & 71.0 & 62.9 & 58.6 & 96.5 & 94.3 & 93.5 \\
\cmidrule(lr){2-12}
 & \multirow[c]{4}{*}{Suffix} & \multirow[c]{2}{*}{1} & LLaDA & 67.4 & 11.3 & 70.8 & 55.5 & 45.8 & 98.4 & 96.3 & 94.6 \\
 &  &  & Dream & 59.8 & 11.3 & 84.5 & 61.4 & 57.1 & 98.3 & 91.6 & 86.8 \\
 &  & \multirow[c]{2}{*}{2} & LLaDA & 66.8 & 11.2 & 81.9 & 63.1 & 50.1 & 99.1 & 97.3 & 93.7 \\
 &  &  & Dream & 76.7 & 11.2 & 89.4 & 68.8 & 69.7 & 99.3 & 96.8 & 91.6 \\
\midrule
\multirow[c]{12}{*}{MBPP} & \multirow[c]{4}{*}{Prefix} & \multirow[c]{2}{*}{1} & LLaDA & 59.0 & 11.4 & 70.2 & 68.8 & 63.1 & 97.9 & 97.6 & 95.4 \\
 &  &  & Dream & 81.2 & 11.5 & 79.0 & 65.5 & 69.9 & 99.5 & 98.5 & 98.7 \\
 &  & \multirow[c]{2}{*}{2} & LLaDA & 58.2 & 11.2 & 68.0 & 62.9 & 57.7 & 97.4 & 96.2 & 95.7 \\
 &  &  & Dream & 81.4 & 11.2 & 75.7 & 64.1 & 68.5 & 99.3 & 98.3 & 98.8 \\
\cmidrule(lr){2-12}
 & \multirow[c]{4}{*}{Infix} & \multirow[c]{2}{*}{1} & LLaDA & 47.2 & 11.2 & 62.1 & 58.0 & 55.9 & 95.2 & 94.9 & 92.2 \\
 &  &  & Dream & 48.6 & 11.3 & 69.6 & 54.4 & 68.9 & 95.8 & 90.9 & 96.3 \\
 &  & \multirow[c]{2}{*}{2} & LLaDA & 56.2 & 11.2 & 62.6 & 55.5 & 51.7 & 95.8 & 95.3 & 93.2 \\
 &  &  & Dream & 56.3 & 11.2 & 71.3 & 62.6 & 66.8 & 94.6 & 93.1 & 93.4 \\
\cmidrule(lr){2-12}
 & \multirow[c]{4}{*}{Suffix} & \multirow[c]{2}{*}{1} & LLaDA & 63.9 & 11.4 & 76.7 & 65.3 & 56.6 & 99.1 & 96.7 & 95.0 \\
 &  &  & Dream & 67.3 & 11.4 & 73.1 & 64.7 & 63.8 & 97.9 & 95.9 & 95.5 \\
 &  & \multirow[c]{2}{*}{2} & LLaDA & 64.1 & 11.2 & 76.9 & 66.8 & 59.8 & 98.7 & 97.6 & 96.0 \\
 &  &  & Dream & 69.1 & 11.2 & 77.9 & 64.5 & 65.5 & 98.1 & 96.7 & 95.1 \\
\bottomrule
\end{tabular}
\end{table*}

Figure~\ref{fig:anchor-observation} illustrates on three samples that importance concentrates on a small subset of affix tokens and that this subset stays stable as decoding proceeds. Table~\ref{tab:anchor-stability} measures both over the full test sets of GSM8K (1,319 questions) and MBPP (500), with 1- and 2-shot affixes at every position, on Fast-dLLM at Anchor ratio 0.2. We omit BABILong, whose 8-token answer fits one block and so leaves no drift to measure. Decoding runs with full recomputation, so ACache does not shape the trajectory. At each of the 8 block boundaries $b$, Fast-dLLM already runs a full-sequence forward pass; from it we compute each affix token's importance $s_j$ (Section~\ref{sec:design}) over the positions still masked, and take the top $K = \lceil \rho L \rceil$ tokens as the current top-$K$ set. We compare that set with two reference sets and, as a baseline, a random set of the same size: Fresh, the top-$K$ set at $b=1$, which isolates drift during decoding; and ACache's Anchors, selected against the precomputed affix cache (Section~\ref{sec:design}). Jaccard similarity is the fraction of positions two sets have in common (intersection over union). Mass recall divides the importance a reference set captures at $b$ by that of the current top-$K$ set, so swapping tokens of near-equal importance barely lowers it. One MBPP question (infix, 2-shot, LLaDA) is excluded because its attention scores at some boundary were not finite.

Importance concentrates: the top 20\% of affix tokens receive 47--81\% of the mask-to-affix attention, against 20\% under uniform importance. The Fresh set's membership drifts: by the last boundary its Jaccard similarity is 52--69\%, still well above the 11--12\% of random sets. Its importance drifts far less: mass recall is at least 91\% at the second boundary and 89\% at the last, whereas a random set reaches about $\rho/\text{Share}$, or 25--42\%. A fresh selection at request start thus captures nearly as much importance as one made later, so one selection per request suffices. ACache's Anchors share fewer positions with the current top-$K$ set, since they are scored against the stale precomputed affix cache: their Jaccard similarity is 29--70\%, at least 2.6 times a random set's. Their mass recall averages 77--99\% and falls below 84\% only for LLaDA on 1-shot GSM8K prefix, where accuracy stays within 1.9 points of full recomputation (Figure~\ref{fig:affix-accuracy}). Scoring with fresh affix KV would require recomputing the whole affix, the cost ACache avoids. At ratio 0.3, on the same runs, the same holds: the top 30\% capture 58--86\%, Fresh mass recall is at least 93\%, and the Anchors' average mass recall is 79--99\%.

\subsection{Anchor Selection Overhead}
\label{sec:appendix-anchor-selection-overhead}

\begin{table*}[t]
\centering
\caption{\textbf{Anchor selection cost per request (ms) under prefix reuse, batch size 1.} Profiled on the runs of Table~\ref{tab:system-recompute-grid} (LLaDA) and Table~\ref{tab:dream-system-recompute-grid} (Dream). Affix reports the shared span's length in tokens.}
\label{tab:anchor-selection-overhead}
\EvalTableFont
\setlength{\tabcolsep}{4pt}
\begin{tabular}{L{\GridColModel}L{\GridColDataset}crrrrrr}
\toprule
Model & Dataset & Shot & Affix & Preparation & Forward & Top-$K$ & Total & Forward share \\
\midrule
\multirow[c]{6}{*}{LLaDA} & \multirow[c]{3}{*}{GSM8K} & 1 & 213 & 0.49 & 50.05 & 0.16 & 50.69 & 98.7\% \\
 & & 2 & 549 & 0.52 & 56.46 & 0.16 & 57.14 & 98.8\% \\
 & & 4 & 981 & 0.57 & 66.02 & 0.18 & 66.76 & 98.9\% \\
\cmidrule(lr){2-9}
 & \multirow[c]{3}{*}{MBPP} & 1 & 178 & 0.55 & 54.70 & 0.16 & 55.40 & 98.7\% \\
 & & 2 & 455 & 0.55 & 61.40 & 0.19 & 62.14 & 98.8\% \\
 & & 4 & 1016 & 0.60 & 71.57 & 0.21 & 72.38 & 98.9\% \\
\midrule
\multirow[c]{6}{*}{Dream} & \multirow[c]{3}{*}{GSM8K} & 1 & 213 & 0.49 & 44.26 & 0.16 & 44.90 & 98.6\% \\
 & & 2 & 538 & 0.51 & 46.42 & 0.16 & 47.09 & 98.6\% \\
 & & 4 & 957 & 0.55 & 49.95 & 0.18 & 50.68 & 98.6\% \\
\cmidrule(lr){2-9}
 & \multirow[c]{3}{*}{MBPP} & 1 & 172 & 0.52 & 47.51 & 0.18 & 48.21 & 98.5\% \\
 & & 2 & 434 & 0.53 & 49.27 & 0.20 & 50.01 & 98.5\% \\
 & & 4 & 951 & 0.59 & 53.91 & 0.20 & 54.70 & 98.5\% \\
\bottomrule
\end{tabular}
\end{table*}

Table~\ref{tab:anchor-selection-overhead} separates ACache's one-time Anchor selection cost into preparation, the mask-to-affix attention forward pass, and top-$K$ selection. Selection costs 44.9--72.4 ms per request. The forward pass accounts for 98.5--98.9\% of it, while preparation costs at most 0.60 ms and top-$K$ at most 0.21 ms. The cost tracks the affix length rather than the decoding loop: it grows from 44.9--55.4 ms at 1-shot to 50.7--72.4 ms at 4-shot, as the shared span grows from 172--213 to 951--1016 tokens. The selector is thus a single extra forward pass per request, forming queries only for request-owned tokens, never for the $L$ affix positions (Section~\ref{sec:design}).

We report batch size 1 because Anchor selection runs one request at a time, so the inference batch size does not change what the selector does. Measurement agrees: holding everything else fixed, per-request selection cost varies by at most 1.8\% across batch sizes 1, 4, and 16, except for Dream on 1-shot GSM8K infix, where it varies by 4.3\%. Affix position matters more. Relative to prefix, suffix reuse costs 0.8--4.7\% more and infix reuse 2.1--15.6\% more. Preparation, which builds the read-slot map for a relocated span, rises from about 0.5 ms to 1.4--2.7 ms and covers the smaller gaps; the widest infix gaps come from the forward pass, because the infix layout adds a query header and answer prompt to the tokens the pass forms queries for (Appendix~\ref{sec:prompt-layouts}). Precomputing the shared affix cache is a separate cost, paid once per shared span rather than per request: 0.76--1.24 s, amortized across the requests that reuse it. A $4.5\times$ longer affix takes only $1.1\times$ longer to precompute, so most of that time is fixed setup rather than the forward pass itself.

Both the selector's cost and the recomputation it saves scale with the workload. Anchor selection is 2.0--4.8\% of ACache's per-request end-to-end time at batch size 1, rising to 7.5--11.1\% at batch size 16, where the decode work attributable to each request shrinks while the selector's one pass does not. The recompute savings grow faster: across both models, ACache reduces recompute latency by 17.0--56.7\% and improves throughput in every tested batch-4 and batch-16 setting (Table~\ref{tab:system-throughput-grid}). At batch size 1, where the shortest affix leaves the least to save, selection is repaid in seven of the eight settings with more than one shot ($1.04$--$1.26\times$); the exception, Dream on GSM8K at 2-shot, breaks even. One caveat applies to all of these measurements: the selection pass runs with module compilation disabled, since its attention hook has to observe uncompiled forwards, while ACache's decode and recomputation passes run compiled. The reported selector cost is therefore an upper bound.

\subsection{Dream System Efficiency}
\label{sec:appendix-dream-system-efficiency}

\begin{table*}[t]
\centering
\caption{\textbf{Dream recompute latency (ms) under prefix reuse.} LLaDA: Table~\ref{tab:system-recompute-grid}.}
\label{tab:dream-system-recompute-grid}
\EvalTableFont
\setlength{\tabcolsep}{2.5pt}
\begin{tabular}{L{\GridColDataset}Z{\GridColBatch}rrrrrr}
\toprule
\TableHeadMultirow{Dataset} & \TableHeadMultirow{Batch} & \multicolumn{2}{c}{1-shot} & \multicolumn{2}{c}{2-shot} & \multicolumn{2}{c}{4-shot} \\
\cmidrule(lr){3-4}\cmidrule(lr){5-6}\cmidrule(lr){7-8}
 & & \RecomputePerfHead{Fast-dLLM} & \RecomputePerfHead{ACache} & \RecomputePerfHead{Fast-dLLM} & \RecomputePerfHead{ACache} & \RecomputePerfHead{Fast-dLLM} & \RecomputePerfHead{ACache} \\
\midrule
\multirow[c]{3}{*}{GSM8K} & 1 & \RecomputePerfCell{52.8} & \RecomputePerfCell{41.6 \RecomputePerfDrop{21.3}} & \RecomputePerfCell{72.5} & \RecomputePerfCell{46.4 \RecomputePerfDrop{35.9}} & \RecomputePerfCell{111.7} & \RecomputePerfCell{54.8 \RecomputePerfDrop{50.9}} \\
 & 4 & \RecomputePerfCell{60.7} & \RecomputePerfCell{46.6 \RecomputePerfDrop{23.3}} & \RecomputePerfCell{87.0} & \RecomputePerfCell{52.3 \RecomputePerfDrop{39.9}} & \RecomputePerfCell{132.4} & \RecomputePerfCell{63.5 \RecomputePerfDrop{52.0}} \\
 & 16 & \RecomputePerfCell{96.5} & \RecomputePerfCell{70.2 \RecomputePerfDrop{27.3}} & \RecomputePerfCell{147.0} & \RecomputePerfCell{80.4 \RecomputePerfDrop{45.3}} & \RecomputePerfCell{233.5} & \RecomputePerfCell{102.3 \RecomputePerfDrop{56.2}} \\
\midrule
\multirow[c]{3}{*}{MBPP} & 1 & \RecomputePerfCell{54.4} & \RecomputePerfCell{45.2 \RecomputePerfDrop{17.0}} & \RecomputePerfCell{72.7} & \RecomputePerfCell{49.2 \RecomputePerfDrop{32.4}} & \RecomputePerfCell{119.0} & \RecomputePerfCell{60.3 \RecomputePerfDrop{49.4}} \\
 & 4 & \RecomputePerfCell{70.9} & \RecomputePerfCell{56.0 \RecomputePerfDrop{21.0}} & \RecomputePerfCell{96.7} & \RecomputePerfCell{62.2 \RecomputePerfDrop{35.7}} & \RecomputePerfCell{162.5} & \RecomputePerfCell{76.7 \RecomputePerfDrop{52.8}} \\
 & 16 & \RecomputePerfCell{130.2} & \RecomputePerfCell{98.6 \RecomputePerfDrop{24.3}} & \RecomputePerfCell{184.3} & \RecomputePerfCell{110.2 \RecomputePerfDrop{40.2}} & \RecomputePerfCell{310.6} & \RecomputePerfCell{142.8 \RecomputePerfDrop{54.0}} \\
\bottomrule
\end{tabular}
\end{table*}

Table~\ref{tab:dream-system-recompute-grid} complements the LLaDA recomputation results in Table~\ref{tab:system-recompute-grid}; Table~\ref{tab:system-throughput-grid} already reports throughput for both models. ACache reduces Dream recompute latency in every tested setting, by 17.0\% to 56.2\%. These gains are largest with the longest shared affix: at 4-shot, the reduction is 49.4--56.2\% across both datasets and all batch sizes, agreeing with LLaDA's to within 2.5 points. At 1- and 2-shot, Dream's reductions are lower than LLaDA's in 11 of the 12 settings, by up to 13.6 points.

\begin{table*}[t]
\centering
\caption{\textbf{Dream peak KV cache memory usage (GB) under prefix reuse.} LLaDA: Table~\ref{tab:peak-kv-memory-grid}.}
\label{tab:dream-peak-kv-memory-grid}
\EvalTableFont
\setlength{\tabcolsep}{2.5pt}
\begin{tabular}{L{\GridColDataset}Z{\GridColBatch}rrrrrr}
\toprule
\TableHeadMultirow{Dataset} & \TableHeadMultirow{Batch} & \multicolumn{2}{c}{1-shot} & \multicolumn{2}{c}{2-shot} & \multicolumn{2}{c}{4-shot} \\
\cmidrule(lr){3-4}\cmidrule(lr){5-6}\cmidrule(lr){7-8}
 & & \MemoryPerfHead{Fast-dLLM} & \MemoryPerfHead{ACache} & \MemoryPerfHead{Fast-dLLM} & \MemoryPerfHead{ACache} & \MemoryPerfHead{Fast-dLLM} & \MemoryPerfHead{ACache} \\
\midrule
\multirow[c]{3}{*}{GSM8K} & 1 & \MemoryPerfCell{0.04} & \MemoryPerfCell{0.04 \MemoryPerfSame{0.0}} & \MemoryPerfCell{0.05} & \MemoryPerfCell{0.08 \MemoryPerfRise{50.0}} & \MemoryPerfCell{0.08} & \MemoryPerfCell{0.10 \MemoryPerfRise{16.7}} \\
 & 4 & \MemoryPerfCell{0.16} & \MemoryPerfCell{0.12 \MemoryPerfDrop{25.0}} & \MemoryPerfCell{0.22} & \MemoryPerfCell{0.16 \MemoryPerfDrop{25.0}} & \MemoryPerfCell{0.33} & \MemoryPerfCell{0.22 \MemoryPerfDrop{33.3}} \\
 & 16 & \MemoryPerfCell{0.66} & \MemoryPerfCell{0.45 \MemoryPerfDrop{31.2}} & \MemoryPerfCell{0.88} & \MemoryPerfCell{0.51 \MemoryPerfDrop{42.2}} & \MemoryPerfCell{1.31} & \MemoryPerfCell{0.71 \MemoryPerfDrop{45.8}} \\
\midrule
\multirow[c]{3}{*}{MBPP} & 1 & \MemoryPerfCell{0.23} & \MemoryPerfCell{0.23 \MemoryPerfSame{0.0}} & \MemoryPerfCell{0.25} & \MemoryPerfCell{0.25 \MemoryPerfSame{0.0}} & \MemoryPerfCell{0.27} & \MemoryPerfCell{0.29 \MemoryPerfRise{5.0}} \\
 & 4 & \MemoryPerfCell{0.36} & \MemoryPerfCell{0.33 \MemoryPerfDrop{7.7}} & \MemoryPerfCell{0.41} & \MemoryPerfCell{0.36 \MemoryPerfDrop{13.3}} & \MemoryPerfCell{0.52} & \MemoryPerfCell{0.41 \MemoryPerfDrop{21.1}} \\
 & 16 & \MemoryPerfCell{0.83} & \MemoryPerfCell{0.68 \MemoryPerfDrop{18.0}} & \MemoryPerfCell{1.07} & \MemoryPerfCell{0.71 \MemoryPerfDrop{33.3}} & \MemoryPerfCell{1.50} & \MemoryPerfCell{0.90 \MemoryPerfDrop{40.0}} \\
\bottomrule
\end{tabular}
\end{table*}

Table~\ref{tab:dream-peak-kv-memory-grid} complements the LLaDA memory results in Table~\ref{tab:peak-kv-memory-grid}. Absolute KV cache memory is lower for Dream because it has 4 grouped-query KV heads in each of 28 layers where LLaDA has 32 in each of 32, about $9\times$ smaller per token; the meaningful comparison is therefore the within-model Fast-dLLM-to-ACache reduction. Those reductions are identical to LLaDA's in 16 of the 18 settings, differing only on MBPP at batch size 16, where the baseline occupies a different number of cache blocks. The batch-size-1 increases come from the same block-granularity effect described in Section~\ref{sec:system-efficiency}, and the largest increase is 0.03 GB, against savings that reach 0.60 GB at batch size 16.

\subsection{dKV-Cache System Efficiency}
\label{sec:appendix-dkv-system-efficiency}

\newcommand{\ApxTwoModelHead}[1]{\makebox[0.6in][c]{#1}}
\newcommand{\ApxTwoModelCell}[1]{\makebox[0.6in][c]{#1}}

\begin{table*}[t]
\centering
\caption{\textbf{dKV-Cache periodic full-cache recompute latency (ms) under prefix reuse.} OOM marks a run that exceeded GPU memory.}
\label{tab:dkv-system-recompute-grid}
\EvalTableFont
\setlength{\tabcolsep}{2.5pt}
\begin{tabular}{L{\GridColModel}L{\GridColDataset}Z{\GridColBatch}rrrrrr}
\toprule
\TableHeadMultirow{Model} & \TableHeadMultirow{Dataset} & \TableHeadMultirow{Batch} & \multicolumn{2}{c}{1-shot} & \multicolumn{2}{c}{2-shot} & \multicolumn{2}{c}{4-shot} \\
\cmidrule(lr){4-5}\cmidrule(lr){6-7}\cmidrule(lr){8-9}
 & & & \ApxTwoModelHead{dKV-Cache} & \ApxTwoModelHead{ACache} & \ApxTwoModelHead{dKV-Cache} & \ApxTwoModelHead{ACache} & \ApxTwoModelHead{dKV-Cache} & \ApxTwoModelHead{ACache} \\
\midrule
\TableModelMultirow{LLaDA} & \multirow[c]{3}{*}{GSM8K} & 1 & \ApxTwoModelCell{54.1} & \ApxTwoModelCell{38.1 \RecomputePerfDrop{29.6}} & \ApxTwoModelCell{88.4} & \ApxTwoModelCell{44.9 \RecomputePerfDrop{49.2}} & \ApxTwoModelCell{120.4} & \ApxTwoModelCell{58.3 \RecomputePerfDrop{51.6}} \\
 & & 4 & \ApxTwoModelCell{166.2} & \ApxTwoModelCell{117.4 \RecomputePerfDrop{29.4}} & \ApxTwoModelCell{285.7} & \ApxTwoModelCell{152.1 \RecomputePerfDrop{46.8}} & \ApxTwoModelCell{439.5} & \ApxTwoModelCell{181.9 \RecomputePerfDrop{58.6}} \\
 & & 16 & \ApxTwoModelCell{627.4} & \ApxTwoModelCell{435.2 \RecomputePerfDrop{30.6}} & \ApxTwoModelCell{1053.7} & \ApxTwoModelCell{533.4 \RecomputePerfDrop{49.4}} & \ApxTwoModelCell{1674.0} & \ApxTwoModelCell{672.7 \RecomputePerfDrop{59.8}} \\
\cmidrule(lr){2-9}
 & \multirow[c]{3}{*}{MBPP} & 1 & \ApxTwoModelCell{55.8} & \ApxTwoModelCell{43.1 \RecomputePerfDrop{22.8}} & \ApxTwoModelCell{88.2} & \ApxTwoModelCell{48.8 \RecomputePerfDrop{44.6}} & \ApxTwoModelCell{129.2} & \ApxTwoModelCell{63.9 \RecomputePerfDrop{50.6}} \\
 & & 4 & \ApxTwoModelCell{175.3} & \ApxTwoModelCell{138.5 \RecomputePerfDrop{21.0}} & \ApxTwoModelCell{263.9} & \ApxTwoModelCell{162.3 \RecomputePerfDrop{38.5}} & \ApxTwoModelCell{477.1} & \ApxTwoModelCell{205.1 \RecomputePerfDrop{57.0}} \\
 & & 16 & \ApxTwoModelCell{647.5} & \ApxTwoModelCell{484.8 \RecomputePerfDrop{25.1}} & \ApxTwoModelCell{1019.6} & \ApxTwoModelCell{586.9 \RecomputePerfDrop{42.4}} & \ApxTwoModelCell{1997.7} & \ApxTwoModelCell{784.2 \RecomputePerfDrop{60.7}} \\
\midrule
\TableModelMultirow{Dream} & \multirow[c]{3}{*}{GSM8K} & 1 & \ApxTwoModelCell{52.8} & \ApxTwoModelCell{41.6 \RecomputePerfDrop{21.3}} & \ApxTwoModelCell{72.4} & \ApxTwoModelCell{46.4 \RecomputePerfDrop{35.9}} & \ApxTwoModelCell{111.5} & \ApxTwoModelCell{54.6 \RecomputePerfDrop{51.0}} \\
 & & 4 & \ApxTwoModelCell{168.3} & \ApxTwoModelCell{121.0 \RecomputePerfDrop{28.1}} & \ApxTwoModelCell{270.9} & \ApxTwoModelCell{139.4 \RecomputePerfDrop{48.5}} & \ApxTwoModelCell{423.5} & \ApxTwoModelCell{177.6 \RecomputePerfDrop{58.1}} \\
 & & 16 & \ApxTwoModelCell{622.7} & \ApxTwoModelCell{434.6 \RecomputePerfDrop{30.2}} & \ApxTwoModelCell{1021.0} & \ApxTwoModelCell{523.9 \RecomputePerfDrop{48.7}} & \ApxTwoModelCell{1609.1} & \ApxTwoModelCell{654.6 \RecomputePerfDrop{59.3}} \\
\cmidrule(lr){2-9}
 & \multirow[c]{3}{*}{MBPP} & 1 & \ApxTwoModelCell{54.1} & \ApxTwoModelCell{45.0 \RecomputePerfDrop{16.9}} & \ApxTwoModelCell{72.1} & \ApxTwoModelCell{49.1 \RecomputePerfDrop{31.9}} & \ApxTwoModelCell{119.6} & \ApxTwoModelCell{60.4 \RecomputePerfDrop{49.5}} \\
 & & 4 & \ApxTwoModelCell{178.4} & \ApxTwoModelCell{136.3 \RecomputePerfDrop{23.6}} & \ApxTwoModelCell{259.3} & \ApxTwoModelCell{153.5 \RecomputePerfDrop{40.8}} & \ApxTwoModelCell{439.2} & \ApxTwoModelCell{196.8 \RecomputePerfDrop{55.2}} \\
 & & 16 & \ApxTwoModelCell{642.5} & \ApxTwoModelCell{483.0 \RecomputePerfDrop{24.8}} & \ApxTwoModelCell{956.2} & \ApxTwoModelCell{560.5 \RecomputePerfDrop{41.4}} & \ApxTwoModelCell{\textit{OOM}} & \ApxTwoModelCell{791.7} \\
\bottomrule
\end{tabular}
\end{table*}

\begin{table*}[t]
\centering
\caption{\textbf{End-to-end throughput under prefix reuse on dKV-Cache.} Measured in generated tokens per second. OOM marks a run that exceeded GPU memory.}
\label{tab:dkv-system-throughput-grid}
\EvalTableFont
\setlength{\tabcolsep}{2.5pt}
\begin{tabular}{L{\GridColModel}L{\GridColDataset}Z{\GridColBatch}rrrrrr}
\toprule
\TableHeadMultirow{Model} & \TableHeadMultirow{Dataset} & \TableHeadMultirow{Batch} & \multicolumn{2}{c}{1-shot} & \multicolumn{2}{c}{2-shot} & \multicolumn{2}{c}{4-shot} \\
\cmidrule(lr){4-5}\cmidrule(lr){6-7}\cmidrule(lr){8-9}
 & & & \SystemPerfHead{dKV-Cache} & \SystemPerfHead{ACache} & \SystemPerfHead{dKV-Cache} & \SystemPerfHead{ACache} & \SystemPerfHead{dKV-Cache} & \SystemPerfHead{ACache} \\
\midrule
\TableModelMultirow{LLaDA} & \multirow[c]{3}{*}{GSM8K} & 1 & \SystemPerfCell{33} & \SystemPerfCell{35 \SystemPerfGain{1.08}} & \SystemPerfCell{28} & \SystemPerfCell{33 \SystemPerfGain{1.21}} & \SystemPerfCell{24} & \SystemPerfCell{31 \SystemPerfGain{1.30}} \\
 & & 4 & \SystemPerfCell{56} & \SystemPerfCell{62 \SystemPerfGain{1.11}} & \SystemPerfCell{43} & \SystemPerfCell{56 \SystemPerfGain{1.29}} & \SystemPerfCell{33} & \SystemPerfCell{50 \SystemPerfGain{1.51}} \\
 & & 16 & \SystemPerfCell{68} & \SystemPerfCell{77 \SystemPerfGain{1.14}} & \SystemPerfCell{51} & \SystemPerfCell{69 \SystemPerfGain{1.35}} & \SystemPerfCell{38} & \SystemPerfCell{60 \SystemPerfGain{1.60}} \\
\cmidrule(lr){2-9}
 & \multirow[c]{3}{*}{MBPP} & 1 & \SystemPerfCell{32} & \SystemPerfCell{34 \SystemPerfGain{1.05}} & \SystemPerfCell{28} & \SystemPerfCell{33 \SystemPerfGain{1.18}} & \SystemPerfCell{23} & \SystemPerfCell{30 \SystemPerfGain{1.29}} \\
 & & 4 & \SystemPerfCell{55} & \SystemPerfCell{59 \SystemPerfGain{1.08}} & \SystemPerfCell{45} & \SystemPerfCell{55 \SystemPerfGain{1.22}} & \SystemPerfCell{32} & \SystemPerfCell{48 \SystemPerfGain{1.51}} \\
 & & 16 & \SystemPerfCell{66} & \SystemPerfCell{73 \SystemPerfGain{1.11}} & \SystemPerfCell{52} & \SystemPerfCell{67 \SystemPerfGain{1.27}} & \SystemPerfCell{35} & \SystemPerfCell{56 \SystemPerfGain{1.59}} \\
\midrule
\TableModelMultirow{Dream} & \multirow[c]{3}{*}{GSM8K} & 1 & \SystemPerfCell{32} & \SystemPerfCell{33 \SystemPerfGain{1.03}} & \SystemPerfCell{29} & \SystemPerfCell{32 \SystemPerfGain{1.10}} & \SystemPerfCell{25} & \SystemPerfCell{31 \SystemPerfGain{1.21}} \\
 & & 4 & \SystemPerfCell{59} & \SystemPerfCell{64 \SystemPerfGain{1.09}} & \SystemPerfCell{49} & \SystemPerfCell{61 \SystemPerfGain{1.24}} & \SystemPerfCell{39} & \SystemPerfCell{55 \SystemPerfGain{1.41}} \\
 & & 16 & \SystemPerfCell{72} & \SystemPerfCell{80 \SystemPerfGain{1.11}} & \SystemPerfCell{56} & \SystemPerfCell{72 \SystemPerfGain{1.28}} & \SystemPerfCell{44} & \SystemPerfCell{65 \SystemPerfGain{1.48}} \\
\cmidrule(lr){2-9}
 & \multirow[c]{3}{*}{MBPP} & 1 & \SystemPerfCell{32} & \SystemPerfCell{33 \SystemPerfGain{1.03}} & \SystemPerfCell{30} & \SystemPerfCell{32 \SystemPerfGain{1.08}} & \SystemPerfCell{25} & \SystemPerfCell{30 \SystemPerfGain{1.21}} \\
 & & 4 & \SystemPerfCell{58} & \SystemPerfCell{63 \SystemPerfGain{1.07}} & \SystemPerfCell{50} & \SystemPerfCell{59 \SystemPerfGain{1.19}} & \SystemPerfCell{37} & \SystemPerfCell{52 \SystemPerfGain{1.40}} \\
 & & 16 & \SystemPerfCell{71} & \SystemPerfCell{77 \SystemPerfGain{1.08}} & \SystemPerfCell{59} & \SystemPerfCell{71 \SystemPerfGain{1.22}} & \SystemPerfCell{\textit{OOM}} & \SystemPerfCell{56} \\
\bottomrule
\end{tabular}
\end{table*}

\begin{table*}[t]
\centering
\caption{\textbf{Peak KV cache memory usage (GB) under prefix reuse on dKV-Cache.} OOM marks a run that exceeded GPU memory.}
\label{tab:dkv-system-peak-kv-grid}
\EvalTableFont
\setlength{\tabcolsep}{2.5pt}
\begin{tabular}{L{\GridColModel}L{\GridColDataset}Z{\GridColBatch}rrrrrr}
\toprule
\TableHeadMultirow{Model} & \TableHeadMultirow{Dataset} & \TableHeadMultirow{Batch} & \multicolumn{2}{c}{1-shot} & \multicolumn{2}{c}{2-shot} & \multicolumn{2}{c}{4-shot} \\
\cmidrule(lr){4-5}\cmidrule(lr){6-7}\cmidrule(lr){8-9}
 & & & \ApxTwoModelHead{dKV-Cache} & \ApxTwoModelHead{ACache} & \ApxTwoModelHead{dKV-Cache} & \ApxTwoModelHead{ACache} & \ApxTwoModelHead{dKV-Cache} & \ApxTwoModelHead{ACache} \\
\midrule
\TableModelMultirow{LLaDA} & \multirow[c]{3}{*}{GSM8K} & 1 & \ApxTwoModelCell{0.38} & \ApxTwoModelCell{0.38 \MemoryPerfSame{0.0}} & \ApxTwoModelCell{0.50} & \ApxTwoModelCell{0.75 \MemoryPerfRise{50.0}} & \ApxTwoModelCell{0.75} & \ApxTwoModelCell{0.88 \MemoryPerfRise{16.7}} \\
 & & 4 & \ApxTwoModelCell{1.50} & \ApxTwoModelCell{1.12 \MemoryPerfDrop{25.0}} & \ApxTwoModelCell{2.00} & \ApxTwoModelCell{1.50 \MemoryPerfDrop{25.0}} & \ApxTwoModelCell{3.00} & \ApxTwoModelCell{2.00 \MemoryPerfDrop{33.3}} \\
 & & 16 & \ApxTwoModelCell{6.00} & \ApxTwoModelCell{4.12 \MemoryPerfDrop{31.2}} & \ApxTwoModelCell{8.00} & \ApxTwoModelCell{4.50 \MemoryPerfDrop{43.8}} & \ApxTwoModelCell{12.00} & \ApxTwoModelCell{6.50 \MemoryPerfDrop{45.8}} \\
\cmidrule(lr){2-9}
 & \multirow[c]{3}{*}{MBPP} & 1 & \ApxTwoModelCell{2.12} & \ApxTwoModelCell{2.12 \MemoryPerfSame{0.0}} & \ApxTwoModelCell{2.25} & \ApxTwoModelCell{2.25 \MemoryPerfSame{0.0}} & \ApxTwoModelCell{2.50} & \ApxTwoModelCell{2.62 \MemoryPerfRise{5.0}} \\
 & & 4 & \ApxTwoModelCell{3.25} & \ApxTwoModelCell{2.88 \MemoryPerfDrop{11.5}} & \ApxTwoModelCell{3.75} & \ApxTwoModelCell{3.00 \MemoryPerfDrop{20.0}} & \ApxTwoModelCell{4.75} & \ApxTwoModelCell{3.75 \MemoryPerfDrop{21.1}} \\
 & & 16 & \ApxTwoModelCell{7.50} & \ApxTwoModelCell{6.12 \MemoryPerfDrop{18.3}} & \ApxTwoModelCell{9.75} & \ApxTwoModelCell{6.38 \MemoryPerfDrop{34.6}} & \ApxTwoModelCell{13.75} & \ApxTwoModelCell{8.25 \MemoryPerfDrop{40.0}} \\
\midrule
\TableModelMultirow{Dream} & \multirow[c]{3}{*}{GSM8K} & 1 & \ApxTwoModelCell{0.04} & \ApxTwoModelCell{0.04 \MemoryPerfSame{0.0}} & \ApxTwoModelCell{0.05} & \ApxTwoModelCell{0.08 \MemoryPerfRise{50.0}} & \ApxTwoModelCell{0.08} & \ApxTwoModelCell{0.10 \MemoryPerfRise{16.7}} \\
 & & 4 & \ApxTwoModelCell{0.16} & \ApxTwoModelCell{0.12 \MemoryPerfDrop{25.0}} & \ApxTwoModelCell{0.22} & \ApxTwoModelCell{0.16 \MemoryPerfDrop{25.0}} & \ApxTwoModelCell{0.33} & \ApxTwoModelCell{0.22 \MemoryPerfDrop{33.3}} \\
 & & 16 & \ApxTwoModelCell{0.66} & \ApxTwoModelCell{0.45 \MemoryPerfDrop{31.2}} & \ApxTwoModelCell{0.88} & \ApxTwoModelCell{0.49 \MemoryPerfDrop{43.7}} & \ApxTwoModelCell{1.30} & \ApxTwoModelCell{0.70 \MemoryPerfDrop{46.3}} \\
\cmidrule(lr){2-9}
 & \multirow[c]{3}{*}{MBPP} & 1 & \ApxTwoModelCell{0.23} & \ApxTwoModelCell{0.23 \MemoryPerfSame{0.0}} & \ApxTwoModelCell{0.25} & \ApxTwoModelCell{0.25 \MemoryPerfSame{0.0}} & \ApxTwoModelCell{0.27} & \ApxTwoModelCell{0.29 \MemoryPerfRise{5.0}} \\
 & & 4 & \ApxTwoModelCell{0.34} & \ApxTwoModelCell{0.31 \MemoryPerfDrop{8.0}} & \ApxTwoModelCell{0.41} & \ApxTwoModelCell{0.33 \MemoryPerfDrop{20.0}} & \ApxTwoModelCell{0.52} & \ApxTwoModelCell{0.41 \MemoryPerfDrop{21.1}} \\
 & & 16 & \ApxTwoModelCell{0.79} & \ApxTwoModelCell{0.67 \MemoryPerfDrop{15.5}} & \ApxTwoModelCell{1.04} & \ApxTwoModelCell{0.70 \MemoryPerfDrop{32.9}} & \ApxTwoModelCell{\textit{OOM}} & \ApxTwoModelCell{0.90} \\
\bottomrule
\end{tabular}
\end{table*}

Tables~\ref{tab:dkv-system-recompute-grid}--\ref{tab:dkv-system-peak-kv-grid} repeat the recompute latency, throughput, and peak KV cache measurements of Tables~\ref{tab:system-recompute-grid}--\ref{tab:peak-kv-memory-grid} with ACache built on dKV-Cache, restricted to the prefix affix position. dKV-Cache recomputes its full cache every 8 decoding steps. In every setting that both mechanisms completed, ACache reduces dKV-Cache recompute latency by 16.9\% to 60.7\%, and unlike on Fast-dLLM it improves throughput in all of them, by $1.03\times$ to $1.60\times$, including at batch size 1 with a 1-shot affix. dKV-Cache runs $3.2\times$ to $7.4\times$ slower than Fast-dLLM on the same setting, so absolute rates here are far lower than in Table~\ref{tab:system-throughput-grid}. That gap is also why batch size 1 gains here where it did not there: Anchor selection is one forward pass per request rather than per step (Appendix~\ref{sec:appendix-anchor-selection-overhead}), so its share of a longer run is smaller. Peak KV cache usage drops by up to 46.3\%, following the same batch-size-1 allocation effect as Table~\ref{tab:peak-kv-memory-grid} and closely tracking its LLaDA values, since peak usage is set by the paged memory manager both mechanisms share. In the one remaining setting, Dream on MBPP at 4-shot and batch 16, standalone dKV-Cache exceeded GPU memory while ACache completed the run, so all three tables report ACache alone there. The failure is one of activation memory in the recomputation pass rather than of cache size; ACache recomputes fewer positions and so needs less of it.

\subsection{Infix and Suffix System Efficiency}
\label{sec:appendix-infix-suffix-system-efficiency}

\begin{table*}[t]
\centering
\caption{\textbf{Recompute latency (ms) under infix reuse.} The shared affix sits between the query header and answer prompt rather than before the query.}
\label{tab:infix-recompute-grid}
\EvalTableFont
\setlength{\tabcolsep}{2.5pt}
\begin{tabular}{L{\GridColModel}L{\GridColDataset}Z{\GridColBatch}rrrrrr}
\toprule
\TableHeadMultirow{Model} & \TableHeadMultirow{Dataset} & \TableHeadMultirow{Batch} & \multicolumn{2}{c}{1-shot} & \multicolumn{2}{c}{2-shot} & \multicolumn{2}{c}{4-shot} \\
\cmidrule(lr){4-5}\cmidrule(lr){6-7}\cmidrule(lr){8-9}
 & & & \ApxTwoModelHead{Fast-dLLM} & \ApxTwoModelHead{ACache} & \ApxTwoModelHead{Fast-dLLM} & \ApxTwoModelHead{ACache} & \ApxTwoModelHead{Fast-dLLM} & \ApxTwoModelHead{ACache} \\
\midrule
\TableModelMultirow{LLaDA} & \multirow[c]{3}{*}{GSM8K} & 1 & \ApxTwoModelCell{56.6} & \ApxTwoModelCell{42.7 \RecomputePerfDrop{24.5}} & \ApxTwoModelCell{91.6} & \ApxTwoModelCell{47.3 \RecomputePerfDrop{48.4}} & \ApxTwoModelCell{123.6} & \ApxTwoModelCell{61.9 \RecomputePerfDrop{49.9}} \\
 &  & 4 & \ApxTwoModelCell{62.9} & \ApxTwoModelCell{50.5 \RecomputePerfDrop{19.7}} & \ApxTwoModelCell{101.7} & \ApxTwoModelCell{55.5 \RecomputePerfDrop{45.5}} & \ApxTwoModelCell{140.7} & \ApxTwoModelCell{69.6 \RecomputePerfDrop{50.6}} \\
 &  & 16 & \ApxTwoModelCell{90.7} & \ApxTwoModelCell{72.0 \RecomputePerfDrop{20.7}} & \ApxTwoModelCell{148.7} & \ApxTwoModelCell{80.3 \RecomputePerfDrop{46.0}} & \ApxTwoModelCell{215.0} & \ApxTwoModelCell{97.9 \RecomputePerfDrop{54.5}} \\
\cmidrule(lr){2-9}
 & \multirow[c]{3}{*}{MBPP} & 1 & \ApxTwoModelCell{61.1} & \ApxTwoModelCell{49.1 \RecomputePerfDrop{19.6}} & \ApxTwoModelCell{94.2} & \ApxTwoModelCell{55.8 \RecomputePerfDrop{40.7}} & \ApxTwoModelCell{138.0} & \ApxTwoModelCell{69.7 \RecomputePerfDrop{49.4}} \\
 &  & 4 & \ApxTwoModelCell{67.6} & \ApxTwoModelCell{56.0 \RecomputePerfDrop{17.3}} & \ApxTwoModelCell{106.3} & \ApxTwoModelCell{63.5 \RecomputePerfDrop{40.3}} & \ApxTwoModelCell{152.6} & \ApxTwoModelCell{77.1 \RecomputePerfDrop{49.5}} \\
 & & 16 & \ApxTwoModelCell{86.0} & \ApxTwoModelCell{73.3 \RecomputePerfDrop{14.7}} & \ApxTwoModelCell{133.6} & \ApxTwoModelCell{84.4 \RecomputePerfDrop{36.9}} & \ApxTwoModelCell{211.5} & \ApxTwoModelCell{104.1 \RecomputePerfDrop{50.8}} \\
\midrule
\TableModelMultirow{Dream} & \multirow[c]{3}{*}{GSM8K} & 1 & \ApxTwoModelCell{53.5} & \ApxTwoModelCell{43.5 \RecomputePerfDrop{18.7}} & \ApxTwoModelCell{74.2} & \ApxTwoModelCell{47.8 \RecomputePerfDrop{35.6}} & \ApxTwoModelCell{111.9} & \ApxTwoModelCell{57.1 \RecomputePerfDrop{48.9}} \\
 &  & 4 & \ApxTwoModelCell{60.4} & \ApxTwoModelCell{49.9 \RecomputePerfDrop{17.4}} & \ApxTwoModelCell{88.1} & \ApxTwoModelCell{56.1 \RecomputePerfDrop{36.3}} & \ApxTwoModelCell{135.0} & \ApxTwoModelCell{66.4 \RecomputePerfDrop{50.8}} \\
 &  & 16 & \ApxTwoModelCell{94.5} & \ApxTwoModelCell{76.2 \RecomputePerfDrop{19.4}} & \ApxTwoModelCell{159.3} & \ApxTwoModelCell{88.2 \RecomputePerfDrop{44.6}} & \ApxTwoModelCell{213.6} & \ApxTwoModelCell{104.6 \RecomputePerfDrop{51.0}} \\
\cmidrule(lr){2-9}
 & \multirow[c]{3}{*}{MBPP} & 1 & \ApxTwoModelCell{57.4} & \ApxTwoModelCell{48.3 \RecomputePerfDrop{15.9}} & \ApxTwoModelCell{74.4} & \ApxTwoModelCell{52.7 \RecomputePerfDrop{29.3}} & \ApxTwoModelCell{121.3} & \ApxTwoModelCell{63.5 \RecomputePerfDrop{47.7}} \\
 & & 4 & \ApxTwoModelCell{73.0} & \ApxTwoModelCell{63.1 \RecomputePerfDrop{13.5}} & \ApxTwoModelCell{103.4} & \ApxTwoModelCell{73.6 \RecomputePerfDrop{28.9}} & \ApxTwoModelCell{171.9} & \ApxTwoModelCell{84.9 \RecomputePerfDrop{50.6}} \\
 & & 16 & \ApxTwoModelCell{135.5} & \ApxTwoModelCell{110.2 \RecomputePerfDrop{18.7}} & \ApxTwoModelCell{200.7} & \ApxTwoModelCell{129.0 \RecomputePerfDrop{35.7}} & \ApxTwoModelCell{326.0} & \ApxTwoModelCell{159.1 \RecomputePerfDrop{51.2}} \\
\bottomrule
\end{tabular}
\end{table*}

\begin{table*}[t]
\centering
\caption{\textbf{End-to-end throughput under infix reuse.} Measured in generated tokens per second.}
\label{tab:infix-throughput-grid}
\EvalTableFont
\setlength{\tabcolsep}{2.5pt}
\begin{tabular}{L{\GridColModel}L{\GridColDataset}Z{\GridColBatch}rrrrrr}
\toprule
\TableHeadMultirow{Model} & \TableHeadMultirow{Dataset} & \TableHeadMultirow{Batch} & \multicolumn{2}{c}{1-shot} & \multicolumn{2}{c}{2-shot} & \multicolumn{2}{c}{4-shot} \\
\cmidrule(lr){4-5}\cmidrule(lr){6-7}\cmidrule(lr){8-9}
 & & & \SystemPerfHead{Fast-dLLM} & \SystemPerfHead{ACache} & \SystemPerfHead{Fast-dLLM} & \SystemPerfHead{ACache} & \SystemPerfHead{Fast-dLLM} & \SystemPerfHead{ACache} \\
\midrule
\TableModelMultirow{LLaDA} & \multirow[c]{3}{*}{GSM8K} & 1 & \SystemPerfCell{105} & \SystemPerfCell{98 \SystemPerfSpeedup{0.93}} & \SystemPerfCell{92} & \SystemPerfCell{95 \SystemPerfGain{1.03}} & \SystemPerfCell{82} & \SystemPerfCell{88 \SystemPerfGain{1.08}} \\
 &  & 4 & \SystemPerfCell{262} & \SystemPerfCell{250 \SystemPerfSpeedup{0.95}} & \SystemPerfCell{202} & \SystemPerfCell{234 \SystemPerfGain{1.16}} & \SystemPerfCell{162} & \SystemPerfCell{207 \SystemPerfGain{1.28}} \\
 &  & 16 & \SystemPerfCell{388} & \SystemPerfCell{392 \SystemPerfGain{1.01}} & \SystemPerfCell{273} & \SystemPerfCell{349 \SystemPerfGain{1.28}} & \SystemPerfCell{193} & \SystemPerfCell{297 \SystemPerfGain{1.54}} \\
\cmidrule(lr){2-9}
 & \multirow[c]{3}{*}{MBPP} & 1 & \SystemPerfCell{198} & \SystemPerfCell{186 \SystemPerfSpeedup{0.94}} & \SystemPerfCell{162} & \SystemPerfCell{178 \SystemPerfGain{1.10}} & \SystemPerfCell{119} & \SystemPerfCell{146 \SystemPerfGain{1.22}} \\
 & & 4 & \SystemPerfCell{368} & \SystemPerfCell{349 \SystemPerfSpeedup{0.95}} & \SystemPerfCell{265} & \SystemPerfCell{326 \SystemPerfGain{1.23}} & \SystemPerfCell{183} & \SystemPerfCell{266 \SystemPerfGain{1.45}} \\
 & & 16 & \SystemPerfCell{445} & \SystemPerfCell{445 \SystemPerfSpeedup{1.00}} & \SystemPerfCell{313} & \SystemPerfCell{408 \SystemPerfGain{1.30}} & \SystemPerfCell{205} & \SystemPerfCell{300 \SystemPerfGain{1.46}} \\
\midrule
\TableModelMultirow{Dream} & \multirow[c]{3}{*}{GSM8K} & 1 & \SystemPerfCell{108} & \SystemPerfCell{108 \SystemPerfSpeedup{0.99}} & \SystemPerfCell{88} & \SystemPerfCell{88 \SystemPerfSpeedup{0.99}} & \SystemPerfCell{58} & \SystemPerfCell{62 \SystemPerfGain{1.07}} \\
 &  & 4 & \SystemPerfCell{247} & \SystemPerfCell{243 \SystemPerfSpeedup{0.98}} & \SystemPerfCell{150} & \SystemPerfCell{160 \SystemPerfGain{1.06}} & \SystemPerfCell{81} & \SystemPerfCell{91 \SystemPerfGain{1.13}} \\
 &  & 16 & \SystemPerfCell{325} & \SystemPerfCell{332 \SystemPerfGain{1.02}} & \SystemPerfCell{174} & \SystemPerfCell{191 \SystemPerfGain{1.09}} & \SystemPerfCell{87} & \SystemPerfCell{102 \SystemPerfGain{1.17}} \\
\cmidrule(lr){2-9}
 & \multirow[c]{3}{*}{MBPP} & 1 & \SystemPerfCell{160} & \SystemPerfCell{152 \SystemPerfSpeedup{0.95}} & \SystemPerfCell{124} & \SystemPerfCell{124 \SystemPerfSpeedup{1.00}} & \SystemPerfCell{65} & \SystemPerfCell{70 \SystemPerfGain{1.08}} \\
 & & 4 & \SystemPerfCell{274} & \SystemPerfCell{265 \SystemPerfSpeedup{0.97}} & \SystemPerfCell{175} & \SystemPerfCell{185 \SystemPerfGain{1.05}} & \SystemPerfCell{77} & \SystemPerfCell{87 \SystemPerfGain{1.13}} \\
 & & 16 & \SystemPerfCell{328} & \SystemPerfCell{333 \SystemPerfGain{1.02}} & \SystemPerfCell{191} & \SystemPerfCell{206 \SystemPerfGain{1.08}} & \SystemPerfCell{81} & \SystemPerfCell{93 \SystemPerfGain{1.15}} \\
\bottomrule
\end{tabular}
\end{table*}

\begin{table*}[t]
\centering
\caption{\textbf{Peak KV cache memory usage (GB) under infix reuse.}}
\label{tab:infix-peak-kv-grid}
\EvalTableFont
\setlength{\tabcolsep}{2.5pt}
\begin{tabular}{L{\GridColModel}L{\GridColDataset}Z{\GridColBatch}rrrrrr}
\toprule
\TableHeadMultirow{Model} & \TableHeadMultirow{Dataset} & \TableHeadMultirow{Batch} & \multicolumn{2}{c}{1-shot} & \multicolumn{2}{c}{2-shot} & \multicolumn{2}{c}{4-shot} \\
\cmidrule(lr){4-5}\cmidrule(lr){6-7}\cmidrule(lr){8-9}
 & & & \ApxTwoModelHead{Fast-dLLM} & \ApxTwoModelHead{ACache} & \ApxTwoModelHead{Fast-dLLM} & \ApxTwoModelHead{ACache} & \ApxTwoModelHead{Fast-dLLM} & \ApxTwoModelHead{ACache} \\
\midrule
\TableModelMultirow{LLaDA} & \multirow[c]{3}{*}{GSM8K} & 1 & \ApxTwoModelCell{0.38} & \ApxTwoModelCell{0.62 \MemoryPerfRise{66.7}} & \ApxTwoModelCell{0.62} & \ApxTwoModelCell{1.12 \MemoryPerfRise{80.0}} & \ApxTwoModelCell{0.75} & \ApxTwoModelCell{1.38 \MemoryPerfRise{83.3}} \\
 & & 4 & \ApxTwoModelCell{1.50} & \ApxTwoModelCell{1.38 \MemoryPerfDrop{8.3}} & \ApxTwoModelCell{2.12} & \ApxTwoModelCell{2.00 \MemoryPerfDrop{5.9}} & \ApxTwoModelCell{3.00} & \ApxTwoModelCell{2.50 \MemoryPerfDrop{16.7}} \\
 & & 16 & \ApxTwoModelCell{6.00} & \ApxTwoModelCell{4.38 \MemoryPerfDrop{27.1}} & \ApxTwoModelCell{8.12} & \ApxTwoModelCell{5.50 \MemoryPerfDrop{32.3}} & \ApxTwoModelCell{12.00} & \ApxTwoModelCell{7.00 \MemoryPerfDrop{41.7}} \\
\cmidrule(lr){2-9}
 & \multirow[c]{3}{*}{MBPP} & 1 & \ApxTwoModelCell{2.12} & \ApxTwoModelCell{2.25 \MemoryPerfRise{5.9}} & \ApxTwoModelCell{2.25} & \ApxTwoModelCell{2.62 \MemoryPerfRise{16.7}} & \ApxTwoModelCell{2.50} & \ApxTwoModelCell{3.12 \MemoryPerfRise{25.0}} \\
 & & 4 & \ApxTwoModelCell{3.25} & \ApxTwoModelCell{3.25 \MemoryPerfSame{0.0}} & \ApxTwoModelCell{3.75} & \ApxTwoModelCell{3.75 \MemoryPerfSame{0.0}} & \ApxTwoModelCell{4.88} & \ApxTwoModelCell{4.25 \MemoryPerfDrop{12.8}} \\
 & & 16 & \ApxTwoModelCell{7.75} & \ApxTwoModelCell{6.75 \MemoryPerfDrop{12.9}} & \ApxTwoModelCell{9.88} & \ApxTwoModelCell{7.88 \MemoryPerfDrop{20.3}} & \ApxTwoModelCell{14.00} & \ApxTwoModelCell{8.75 \MemoryPerfDrop{37.5}} \\
\midrule
\TableModelMultirow{Dream} & \multirow[c]{3}{*}{GSM8K} & 1 & \ApxTwoModelCell{0.04} & \ApxTwoModelCell{0.07 \MemoryPerfRise{66.7}} & \ApxTwoModelCell{0.05} & \ApxTwoModelCell{0.12 \MemoryPerfRise{125.0}} & \ApxTwoModelCell{0.08} & \ApxTwoModelCell{0.15 \MemoryPerfRise{83.3}} \\
 & & 4 & \ApxTwoModelCell{0.16} & \ApxTwoModelCell{0.15 \MemoryPerfDrop{8.3}} & \ApxTwoModelCell{0.22} & \ApxTwoModelCell{0.23 \MemoryPerfRise{6.3}} & \ApxTwoModelCell{0.33} & \ApxTwoModelCell{0.27 \MemoryPerfDrop{16.7}} \\
 & & 16 & \ApxTwoModelCell{0.66} & \ApxTwoModelCell{0.48 \MemoryPerfDrop{27.1}} & \ApxTwoModelCell{0.88} & \ApxTwoModelCell{0.60 \MemoryPerfDrop{31.3}} & \ApxTwoModelCell{1.31} & \ApxTwoModelCell{0.77 \MemoryPerfDrop{41.7}} \\
\cmidrule(lr){2-9}
 & \multirow[c]{3}{*}{MBPP} & 1 & \ApxTwoModelCell{0.23} & \ApxTwoModelCell{0.25 \MemoryPerfRise{5.9}} & \ApxTwoModelCell{0.25} & \ApxTwoModelCell{0.29 \MemoryPerfRise{16.7}} & \ApxTwoModelCell{0.27} & \ApxTwoModelCell{0.34 \MemoryPerfRise{25.0}} \\
 & & 4 & \ApxTwoModelCell{0.36} & \ApxTwoModelCell{0.34 \MemoryPerfDrop{3.8}} & \ApxTwoModelCell{0.41} & \ApxTwoModelCell{0.41 \MemoryPerfSame{0.0}} & \ApxTwoModelCell{0.52} & \ApxTwoModelCell{0.46 \MemoryPerfDrop{10.5}} \\
 & & 16 & \ApxTwoModelCell{0.85} & \ApxTwoModelCell{0.72 \MemoryPerfDrop{14.5}} & \ApxTwoModelCell{1.07} & \ApxTwoModelCell{0.83 \MemoryPerfDrop{21.8}} & \ApxTwoModelCell{1.50} & \ApxTwoModelCell{0.96 \MemoryPerfDrop{36.4}} \\
\bottomrule
\end{tabular}
\end{table*}

\begin{table*}[t]
\centering
\caption{\textbf{Recompute latency (ms) under suffix reuse.} The shared affix sits after the masked generation region rather than before the query.}
\label{tab:suffix-recompute-grid}
\EvalTableFont
\setlength{\tabcolsep}{2.5pt}
\begin{tabular}{L{\GridColModel}L{\GridColDataset}Z{\GridColBatch}rrrrrr}
\toprule
\TableHeadMultirow{Model} & \TableHeadMultirow{Dataset} & \TableHeadMultirow{Batch} & \multicolumn{2}{c}{1-shot} & \multicolumn{2}{c}{2-shot} & \multicolumn{2}{c}{4-shot} \\
\cmidrule(lr){4-5}\cmidrule(lr){6-7}\cmidrule(lr){8-9}
 & & & \ApxTwoModelHead{Fast-dLLM} & \ApxTwoModelHead{ACache} & \ApxTwoModelHead{Fast-dLLM} & \ApxTwoModelHead{ACache} & \ApxTwoModelHead{Fast-dLLM} & \ApxTwoModelHead{ACache} \\
\midrule
\TableModelMultirow{LLaDA} & \multirow[c]{3}{*}{GSM8K} & 1 & \ApxTwoModelCell{56.4} & \ApxTwoModelCell{39.7 \RecomputePerfDrop{29.6}} & \ApxTwoModelCell{91.3} & \ApxTwoModelCell{46.4 \RecomputePerfDrop{49.2}} & \ApxTwoModelCell{123.6} & \ApxTwoModelCell{61.1 \RecomputePerfDrop{50.6}} \\
 &  & 4 & \ApxTwoModelCell{62.5} & \ApxTwoModelCell{46.0 \RecomputePerfDrop{26.5}} & \ApxTwoModelCell{101.7} & \ApxTwoModelCell{54.1 \RecomputePerfDrop{46.8}} & \ApxTwoModelCell{141.9} & \ApxTwoModelCell{68.9 \RecomputePerfDrop{51.5}} \\
 &  & 16 & \ApxTwoModelCell{90.6} & \ApxTwoModelCell{66.6 \RecomputePerfDrop{26.5}} & \ApxTwoModelCell{150.4} & \ApxTwoModelCell{80.4 \RecomputePerfDrop{46.6}} & \ApxTwoModelCell{220.0} & \ApxTwoModelCell{96.8 \RecomputePerfDrop{56.0}} \\
\cmidrule(lr){2-9}
 & \multirow[c]{3}{*}{MBPP} & 1 & \ApxTwoModelCell{59.3} & \ApxTwoModelCell{45.4 \RecomputePerfDrop{23.4}} & \ApxTwoModelCell{91.7} & \ApxTwoModelCell{50.8 \RecomputePerfDrop{44.6}} & \ApxTwoModelCell{134.4} & \ApxTwoModelCell{66.2 \RecomputePerfDrop{50.7}} \\
 &  & 4 & \ApxTwoModelCell{63.3} & \ApxTwoModelCell{51.6 \RecomputePerfDrop{18.5}} & \ApxTwoModelCell{97.8} & \ApxTwoModelCell{56.8 \RecomputePerfDrop{41.9}} & \ApxTwoModelCell{148.4} & \ApxTwoModelCell{73.5 \RecomputePerfDrop{50.4}} \\
 &  & 16 & \ApxTwoModelCell{81.0} & \ApxTwoModelCell{68.2 \RecomputePerfDrop{15.8}} & \ApxTwoModelCell{122.4} & \ApxTwoModelCell{72.5 \RecomputePerfDrop{40.8}} & \ApxTwoModelCell{199.8} & \ApxTwoModelCell{94.4 \RecomputePerfDrop{52.7}} \\
\midrule
\TableModelMultirow{Dream} & \multirow[c]{3}{*}{GSM8K} & 1 & \ApxTwoModelCell{52.8} & \ApxTwoModelCell{42.1 \RecomputePerfDrop{20.2}} & \ApxTwoModelCell{72.5} & \ApxTwoModelCell{46.8 \RecomputePerfDrop{35.4}} & \ApxTwoModelCell{112.5} & \ApxTwoModelCell{56.1 \RecomputePerfDrop{50.2}} \\
 &  & 4 & \ApxTwoModelCell{59.5} & \ApxTwoModelCell{47.0 \RecomputePerfDrop{21.0}} & \ApxTwoModelCell{83.5} & \ApxTwoModelCell{52.9 \RecomputePerfDrop{36.6}} & \ApxTwoModelCell{130.1} & \ApxTwoModelCell{63.9 \RecomputePerfDrop{50.9}} \\
 &  & 16 & \ApxTwoModelCell{90.0} & \ApxTwoModelCell{66.3 \RecomputePerfDrop{26.3}} & \ApxTwoModelCell{130.4} & \ApxTwoModelCell{75.9 \RecomputePerfDrop{41.8}} & \ApxTwoModelCell{209.0} & \ApxTwoModelCell{94.8 \RecomputePerfDrop{54.7}} \\
\cmidrule(lr){2-9}
 & \multirow[c]{3}{*}{MBPP} & 1 & \ApxTwoModelCell{54.5} & \ApxTwoModelCell{45.8 \RecomputePerfDrop{16.0}} & \ApxTwoModelCell{72.2} & \ApxTwoModelCell{49.6 \RecomputePerfDrop{31.3}} & \ApxTwoModelCell{118.9} & \ApxTwoModelCell{60.8 \RecomputePerfDrop{48.9}} \\
 &  & 4 & \ApxTwoModelCell{72.6} & \ApxTwoModelCell{57.8 \RecomputePerfDrop{20.5}} & \ApxTwoModelCell{95.6} & \ApxTwoModelCell{63.8 \RecomputePerfDrop{33.3}} & \ApxTwoModelCell{163.2} & \ApxTwoModelCell{79.8 \RecomputePerfDrop{51.1}} \\
 & & 16 & \ApxTwoModelCell{130.9} & \ApxTwoModelCell{101.8 \RecomputePerfDrop{22.2}} & \ApxTwoModelCell{184.1} & \ApxTwoModelCell{116.0 \RecomputePerfDrop{37.0}} & \ApxTwoModelCell{314.7} & \ApxTwoModelCell{143.5 \RecomputePerfDrop{54.4}} \\
\bottomrule
\end{tabular}
\end{table*}

\begin{table*}[t]
\centering
\caption{\textbf{End-to-end throughput under suffix reuse.} Measured in generated tokens per second.}
\label{tab:suffix-throughput-grid}
\EvalTableFont
\setlength{\tabcolsep}{2.5pt}
\begin{tabular}{L{\GridColModel}L{\GridColDataset}Z{\GridColBatch}rrrrrr}
\toprule
\TableHeadMultirow{Model} & \TableHeadMultirow{Dataset} & \TableHeadMultirow{Batch} & \multicolumn{2}{c}{1-shot} & \multicolumn{2}{c}{2-shot} & \multicolumn{2}{c}{4-shot} \\
\cmidrule(lr){4-5}\cmidrule(lr){6-7}\cmidrule(lr){8-9}
 & & & \SystemPerfHead{Fast-dLLM} & \SystemPerfHead{ACache} & \SystemPerfHead{Fast-dLLM} & \SystemPerfHead{ACache} & \SystemPerfHead{Fast-dLLM} & \SystemPerfHead{ACache} \\
\midrule
\TableModelMultirow{LLaDA} & \multirow[c]{3}{*}{GSM8K} & 1 & \SystemPerfCell{104} & \SystemPerfCell{99 \SystemPerfSpeedup{0.95}} & \SystemPerfCell{90} & \SystemPerfCell{95 \SystemPerfGain{1.06}} & \SystemPerfCell{83} & \SystemPerfCell{88 \SystemPerfGain{1.06}} \\
 & & 4 & \SystemPerfCell{263} & \SystemPerfCell{253 \SystemPerfSpeedup{0.96}} & \SystemPerfCell{200} & \SystemPerfCell{236 \SystemPerfGain{1.18}} & \SystemPerfCell{163} & \SystemPerfCell{212 \SystemPerfGain{1.30}} \\
 &  & 16 & \SystemPerfCell{393} & \SystemPerfCell{410 \SystemPerfGain{1.04}} & \SystemPerfCell{273} & \SystemPerfCell{358 \SystemPerfGain{1.31}} & \SystemPerfCell{200} & \SystemPerfCell{307 \SystemPerfGain{1.54}} \\
\cmidrule(lr){2-9}
 & \multirow[c]{3}{*}{MBPP} & 1 & \SystemPerfCell{127} & \SystemPerfCell{111 \SystemPerfSpeedup{0.87}} & \SystemPerfCell{113} & \SystemPerfCell{138 \SystemPerfGain{1.23}} & \SystemPerfCell{87} & \SystemPerfCell{100 \SystemPerfGain{1.15}} \\
 &  & 4 & \SystemPerfCell{290} & \SystemPerfCell{260 \SystemPerfSpeedup{0.90}} & \SystemPerfCell{225} & \SystemPerfCell{292 \SystemPerfGain{1.30}} & \SystemPerfCell{161} & \SystemPerfCell{223 \SystemPerfGain{1.38}} \\
 &  & 16 & \SystemPerfCell{398} & \SystemPerfCell{390 \SystemPerfSpeedup{0.98}} & \SystemPerfCell{290} & \SystemPerfCell{390 \SystemPerfGain{1.34}} & \SystemPerfCell{191} & \SystemPerfCell{297 \SystemPerfGain{1.56}} \\
\midrule
\TableModelMultirow{Dream} & \multirow[c]{3}{*}{GSM8K} & 1 & \SystemPerfCell{109} & \SystemPerfCell{98 \SystemPerfSpeedup{0.90}} & \SystemPerfCell{96} & \SystemPerfCell{98 \SystemPerfGain{1.02}} & \SystemPerfCell{90} & \SystemPerfCell{97 \SystemPerfGain{1.08}} \\
 & & 4 & \SystemPerfCell{285} & \SystemPerfCell{264 \SystemPerfSpeedup{0.93}} & \SystemPerfCell{229} & \SystemPerfCell{250 \SystemPerfGain{1.09}} & \SystemPerfCell{183} & \SystemPerfCell{237 \SystemPerfGain{1.30}} \\
 & & 16 & \SystemPerfCell{414} & \SystemPerfCell{420 \SystemPerfGain{1.01}} & \SystemPerfCell{297} & \SystemPerfCell{375 \SystemPerfGain{1.26}} & \SystemPerfCell{222} & \SystemPerfCell{338 \SystemPerfGain{1.52}} \\
\cmidrule(lr){2-9}
 & \multirow[c]{3}{*}{MBPP} & 1 & \SystemPerfCell{195} & \SystemPerfCell{178 \SystemPerfSpeedup{0.91}} & \SystemPerfCell{175} & \SystemPerfCell{180 \SystemPerfGain{1.03}} & \SystemPerfCell{137} & \SystemPerfCell{167 \SystemPerfGain{1.22}} \\
 &  & 4 & \SystemPerfCell{405} & \SystemPerfCell{383 \SystemPerfSpeedup{0.95}} & \SystemPerfCell{319} & \SystemPerfCell{363 \SystemPerfGain{1.14}} & \SystemPerfCell{218} & \SystemPerfCell{320 \SystemPerfGain{1.47}} \\
 &  & 16 & \SystemPerfCell{527} & \SystemPerfCell{533 \SystemPerfGain{1.01}} & \SystemPerfCell{386} & \SystemPerfCell{486 \SystemPerfGain{1.26}} & \SystemPerfCell{247} & \SystemPerfCell{411 \SystemPerfGain{1.66}} \\
\bottomrule
\end{tabular}
\end{table*}

\begin{table*}[t]
\centering
\caption{\textbf{Peak KV cache memory usage (GB) under suffix reuse.}}
\label{tab:suffix-peak-kv-grid}
\EvalTableFont
\setlength{\tabcolsep}{2.5pt}
\begin{tabular}{L{\GridColModel}L{\GridColDataset}Z{\GridColBatch}rrrrrr}
\toprule
\TableHeadMultirow{Model} & \TableHeadMultirow{Dataset} & \TableHeadMultirow{Batch} & \multicolumn{2}{c}{1-shot} & \multicolumn{2}{c}{2-shot} & \multicolumn{2}{c}{4-shot} \\
\cmidrule(lr){4-5}\cmidrule(lr){6-7}\cmidrule(lr){8-9}
 & & & \ApxTwoModelHead{Fast-dLLM} & \ApxTwoModelHead{ACache} & \ApxTwoModelHead{Fast-dLLM} & \ApxTwoModelHead{ACache} & \ApxTwoModelHead{Fast-dLLM} & \ApxTwoModelHead{ACache} \\
\midrule
\TableModelMultirow{LLaDA} & \multirow[c]{3}{*}{GSM8K} & 1 & \ApxTwoModelCell{0.38} & \ApxTwoModelCell{0.62 \MemoryPerfRise{66.7}} & \ApxTwoModelCell{0.50} & \ApxTwoModelCell{1.12 \MemoryPerfRise{125.0}} & \ApxTwoModelCell{0.75} & \ApxTwoModelCell{1.38 \MemoryPerfRise{83.3}} \\
 & & 4 & \ApxTwoModelCell{1.50} & \ApxTwoModelCell{1.38 \MemoryPerfDrop{8.3}} & \ApxTwoModelCell{2.00} & \ApxTwoModelCell{2.00 \MemoryPerfSame{0.0}} & \ApxTwoModelCell{3.00} & \ApxTwoModelCell{2.50 \MemoryPerfDrop{16.7}} \\
 & & 16 & \ApxTwoModelCell{6.00} & \ApxTwoModelCell{4.38 \MemoryPerfDrop{27.1}} & \ApxTwoModelCell{8.00} & \ApxTwoModelCell{5.12 \MemoryPerfDrop{35.9}} & \ApxTwoModelCell{12.00} & \ApxTwoModelCell{7.00 \MemoryPerfDrop{41.7}} \\
\cmidrule(lr){2-9}
 & \multirow[c]{3}{*}{MBPP} & 1 & \ApxTwoModelCell{2.12} & \ApxTwoModelCell{2.25 \MemoryPerfRise{5.9}} & \ApxTwoModelCell{2.25} & \ApxTwoModelCell{2.50 \MemoryPerfRise{11.1}} & \ApxTwoModelCell{2.50} & \ApxTwoModelCell{3.12 \MemoryPerfRise{25.0}} \\
 & & 4 & \ApxTwoModelCell{3.25} & \ApxTwoModelCell{3.00 \MemoryPerfDrop{7.7}} & \ApxTwoModelCell{3.75} & \ApxTwoModelCell{3.38 \MemoryPerfDrop{10.0}} & \ApxTwoModelCell{4.75} & \ApxTwoModelCell{4.25 \MemoryPerfDrop{10.5}} \\
 & & 16 & \ApxTwoModelCell{7.75} & \ApxTwoModelCell{6.38 \MemoryPerfDrop{17.7}} & \ApxTwoModelCell{9.75} & \ApxTwoModelCell{7.00 \MemoryPerfDrop{28.2}} & \ApxTwoModelCell{13.75} & \ApxTwoModelCell{8.75 \MemoryPerfDrop{36.4}} \\
\midrule
\TableModelMultirow{Dream} & \multirow[c]{3}{*}{GSM8K} & 1 & \ApxTwoModelCell{0.04} & \ApxTwoModelCell{0.07 \MemoryPerfRise{66.7}} & \ApxTwoModelCell{0.05} & \ApxTwoModelCell{0.12 \MemoryPerfRise{125.0}} & \ApxTwoModelCell{0.08} & \ApxTwoModelCell{0.15 \MemoryPerfRise{83.3}} \\
 & & 4 & \ApxTwoModelCell{0.16} & \ApxTwoModelCell{0.15 \MemoryPerfDrop{8.3}} & \ApxTwoModelCell{0.22} & \ApxTwoModelCell{0.22 \MemoryPerfSame{0.0}} & \ApxTwoModelCell{0.33} & \ApxTwoModelCell{0.27 \MemoryPerfDrop{16.7}} \\
 & & 16 & \ApxTwoModelCell{0.66} & \ApxTwoModelCell{0.48 \MemoryPerfDrop{27.1}} & \ApxTwoModelCell{0.88} & \ApxTwoModelCell{0.56 \MemoryPerfDrop{35.9}} & \ApxTwoModelCell{1.31} & \ApxTwoModelCell{0.77 \MemoryPerfDrop{41.7}} \\
\cmidrule(lr){2-9}
 & \multirow[c]{3}{*}{MBPP} & 1 & \ApxTwoModelCell{0.23} & \ApxTwoModelCell{0.25 \MemoryPerfRise{5.9}} & \ApxTwoModelCell{0.25} & \ApxTwoModelCell{0.27 \MemoryPerfRise{11.1}} & \ApxTwoModelCell{0.27} & \ApxTwoModelCell{0.34 \MemoryPerfRise{25.0}} \\
 & & 4 & \ApxTwoModelCell{0.36} & \ApxTwoModelCell{0.34 \MemoryPerfDrop{3.8}} & \ApxTwoModelCell{0.41} & \ApxTwoModelCell{0.38 \MemoryPerfDrop{6.7}} & \ApxTwoModelCell{0.52} & \ApxTwoModelCell{0.46 \MemoryPerfDrop{10.5}} \\
 & & 16 & \ApxTwoModelCell{0.85} & \ApxTwoModelCell{0.70 \MemoryPerfDrop{17.7}} & \ApxTwoModelCell{1.07} & \ApxTwoModelCell{0.74 \MemoryPerfDrop{30.8}} & \ApxTwoModelCell{1.50} & \ApxTwoModelCell{0.96 \MemoryPerfDrop{36.4}} \\
\bottomrule
\end{tabular}
\end{table*}

Tables~\ref{tab:infix-recompute-grid}--\ref{tab:suffix-peak-kv-grid} repeat the recompute latency, throughput, and peak KV cache measurements of Tables~\ref{tab:system-recompute-grid}--\ref{tab:peak-kv-memory-grid} with the shared affix placed as an infix or a suffix. Recompute latency reductions remain similar across positions: 13.5--54.5\% for infix and 15.8--56.0\% for suffix, close to the 17.0--56.7\% range for prefix. Throughput follows the same pattern as prefix, reaching up to $1.54\times$ for infix and $1.66\times$ for suffix against $1.71\times$ for prefix, but the shortest, 1-shot affix takes longer to amortize: ACache is 1--13\% slower than baseline at batch sizes 1 and 4 for both infix and suffix, and 2\% slower in one setting at batch size 16, whereas every prefix batch-4 and batch-16 setting already improves. Beyond 1-shot, only Dream on GSM8K infix at 2-shot and batch size 1 is slower, by 1\%. Peak KV cache usage drops by up to 41.7\%, vs. 45.8\% for prefix; at batch size 1 it rises by up to 0.63 GB, vs. 0.25 GB for prefix, as a relocated affix also occupies a workspace (Algorithm~\ref{alg:rotation-dispatch}).

\subsection{Non-Anchor Cache Ablation}
\label{sec:appendix-non-anchor-cache}

\begin{table*}[t]
\centering
\caption{\textbf{Keeping versus discarding the non-anchor affix cache, 1-shot.} Accuracy (\%) on Fast-dLLM. \textit{Anchors only} selects the Anchor Tokens but discards the rest of the affix cache. Bold marks the better value per column. ACache rows repeat Table~\ref{tab:anchor-selection-comparison}; as there, Avg.\ omits BABILong suffix.}
\label{tab:non-anchor-ablation}
\EvalTableFont
\setlength{\tabcolsep}{3pt}
\begin{tabular}{L{\AnchorColModel}L{\AnchorColRatio}L{\AnchorColMethod}*{9}{Z{\AnchorColData}}Z{\AnchorColData}}
\toprule
\TableHeadMultirow{Model} & \TableHeadMultirow{Ratio} & \TableHeadMultirow{Variant} & \multicolumn{3}{c}{GSM8K} & \multicolumn{3}{c}{MBPP} & \multicolumn{3}{c}{BABILong} & \TableHeadMultirow{Avg.} \\
\cmidrule(lr){4-6}\cmidrule(lr){7-9}\cmidrule(lr){10-12}
 & & & Pfx. & Inf. & Suf. & Pfx. & Inf. & Suf. & Pfx. & Inf. & Suf. & \\
\midrule
\multirow[c]{4}{*}{\raisebox{-0.35ex}{LLaDA}} & \multirow[c]{2}{*}{0.2} & ACache & \textbf{73.6} & \textbf{69.9} & \textbf{70.3} & \textbf{34.9} & \textbf{15.4} & \textbf{25.1} & 99.6 & 98.9 & \textbf{76.1} & \textbf{61.0} \\
 & & Anchors only & 71.3 & \textbf{69.9} & 69.1 & 2.9 & 6.3 & 4.9 & \textbf{99.8} & \textbf{99.9} & 15.4 & 53.0 \\
\cmidrule(lr){2-13}
 & \multirow[c]{2}{*}{0.3} & ACache & \textbf{74.1} & \textbf{71.0} & \textbf{70.8} & \textbf{34.8} & \textbf{24.7} & \textbf{26.8} & 99.7 & 98.9 & \textbf{78.6} & \textbf{62.6} \\
 & & Anchors only & 73.1 & 70.2 & \textbf{70.8} & 5.3 & 16.2 & 7.7 & \textbf{99.8} & \textbf{99.7} & 56.2 & 55.3 \\
\cmidrule(lr){1-13}
\multirow[c]{4}{*}{\raisebox{-0.35ex}{Dream}} & \multirow[c]{2}{*}{0.2} & ACache & \textbf{70.7} & 74.0 & 70.7 & \textbf{50.0} & \textbf{36.7} & 35.9 & \textbf{95.2} & 86.4 & \textbf{84.5} & 64.9 \\
 & & Anchors only & 69.1 & \textbf{76.7} & \textbf{74.8} & 45.4 & 29.6 & \textbf{49.6} & 92.7 & \textbf{90.6} & 64.0 & \textbf{66.1} \\
\cmidrule(lr){2-13}
 & \multirow[c]{2}{*}{0.3} & ACache & \textbf{72.3} & 72.1 & 70.9 & \textbf{51.3} & \textbf{37.2} & \textbf{50.7} & \textbf{94.9} & 87.1 & \textbf{91.5} & \textbf{67.1} \\
 & & Anchors only & 71.2 & \textbf{74.2} & \textbf{73.5} & 49.3 & 29.1 & 49.3 & 91.1 & \textbf{89.2} & 77.8 & 65.9 \\
\bottomrule
\end{tabular}
\end{table*}

Table~\ref{tab:non-anchor-ablation} compares ACache with a variant that selects the same Anchor Tokens but discards every non-anchor affix cache entry. This tests whether ACache reduces to token-level KV compression or eviction~\citep{zhang2023eviction, xiao2024streaming-llm, li2024snapkv, pang2024anllm, huang2026maskkv, song2026sparse-dllm}, which keeps only important tokens: if the non-anchor tokens were disposable, discarding them should cost nothing once the Anchors are recomputed. The two variants coincide at Anchor ratio 1.0, where nothing is left to discard; the table uses the two ratios of Table~\ref{tab:anchor-selection-comparison}. Discarding costs LLaDA 8.0 points of average accuracy at Anchor ratio 0.2 and 7.2 at 0.3, concentrated in MBPP. On Dream the averages differ by at most 1.2 points, with the variant ahead at 0.2 and behind at 0.3, and it leads in several individual settings, by up to 13.7 points on MBPP suffix at 0.2. The non-anchor affix cache is thus load-bearing on LLaDA but not on Dream. Keeping it is the safe default, and cheap in memory: the read-slot map stores a single copy shared by every request (Section~\ref{sec:system-design}).

\subsection{Top, Random, and Bottom Selection}
\label{sec:appendix-random-bottom-selection}

\begin{table*}[t]
\centering
\caption{\textbf{Top, random, and bottom Anchor selection, LLaDA, 1-shot.} Accuracy (\%) on Fast-dLLM; Top is ACache. Bold marks the best per column and ratio; Avg.\ omits BABILong suffix.}
\label{tab:random-bottom-selection}
\EvalTableFont
\setlength{\tabcolsep}{3pt}
\begin{tabular}{L{\AnchorColRatio}L{\AnchorColMethod}*{9}{Z{\AnchorColData}}Z{\AnchorColData}}
\toprule
\TableHeadMultirow{Ratio} & \TableHeadMultirow{Selection} & \multicolumn{3}{c}{GSM8K} & \multicolumn{3}{c}{MBPP} & \multicolumn{3}{c}{BABILong} & \TableHeadMultirow{Avg.} \\
\cmidrule(lr){3-5}\cmidrule(lr){6-8}\cmidrule(lr){9-11}
 & & Pfx. & Inf. & Suf. & Pfx. & Inf. & Suf. & Pfx. & Inf. & Suf. & \\
\midrule
\multirow[c]{3}{*}{0.2} & Top & \textbf{73.6} & \textbf{69.9} & \textbf{70.3} & \textbf{34.9} & \textbf{15.4} & \textbf{25.1} & \textbf{99.6} & \textbf{98.9} & 76.1 & \textbf{61.0} \\
 & Random & 73.2 & 69.0 & 69.0 & 34.1 & 11.1 & 18.6 & \textbf{99.6} & 80.4 & 84.5 & 56.9 \\
 & Bottom & 71.6 & 69.2 & 68.4 & 34.4 & 4.1 & 8.5 & 99.1 & 74.2 & \textbf{98.1} & 53.7 \\
\cmidrule(lr){1-12}
\multirow[c]{3}{*}{0.3} & Top & \textbf{74.1} & \textbf{71.0} & \textbf{70.8} & \textbf{34.8} & \textbf{24.7} & \textbf{26.8} & \textbf{99.7} & \textbf{98.9} & 78.6 & \textbf{62.6} \\
 & Random & 72.0 & 68.2 & 69.4 & 33.8 & 17.3 & 23.3 & 99.6 & 84.6 & 79.5 & 58.5 \\
 & Bottom & 71.4 & 68.1 & 68.0 & 34.0 & 4.9 & 8.5 & 99.2 & 74.8 & \textbf{97.9} & 53.6 \\
\bottomrule
\end{tabular}
\end{table*}

Table~\ref{tab:random-bottom-selection} keeps ACache's recomputation budget fixed and changes only which affix tokens are recomputed: those ACache ranks most important (Top, its default), a uniformly random subset of the same size drawn per request (Random), or those it ranks least important (Bottom). Accuracy follows the ranking. Outside BABILong suffix, Top matches or beats both alternatives in every setting at both ratios, and Random stays at or above Bottom in all but three, trailing it there by at most 0.3 points. On average, Top leads Random by 4.1 points at both ratios, and Random leads Bottom by 3.2 at ratio 0.2 and 4.9 at 0.3. The gaps concentrate on MBPP infix and suffix and on BABILong infix: at ratio 0.2, MBPP suffix drops from 25.1\% under Top to 18.6\% under Random and 8.5\% under Bottom. ACache's ranking thus tracks how much recomputing each affix token helps. BABILong suffix, where recomputation lowers accuracy (Section~\ref{sec:accuracy-shared-affixes}), reverses the order.

\subsection{System-Prompt Reuse}
\label{sec:appendix-system-prompt}

\begin{table*}[t]
\centering
\caption{\textbf{System-prompt reuse on GSM8K: strict accuracy (\%) on Fast-dLLM.} As in BiCache, an answer counts only if the number after ``\#\#\#\#'' matches. BiCache$^\dagger$ is the official implementation; Ratio is its mean equivalent Anchor ratio. Bold marks the best reuse method per model; underline, the second-best.}
\label{tab:system-prompt-accuracy}
\EvalTableFont
\setlength{\tabcolsep}{3pt}
\begin{tabular}{L{\AnchorColModel}L{1in}L{\AnchorColRatio}Z{\AnchorColData}}
\toprule
Model & Method & Ratio & Acc. \\
\midrule
\multirow[c]{5}{*}{\raisebox{-0.35ex}{LLaDA}} & Direct reuse & 0 & 19.8 \\
\addlinespace[1.5pt]
 & \multirow[c]{2}{*}{ACache} & 0.2 & 59.1 \\
 & & 0.3 & \underline{65.4} \\
\addlinespace[1.5pt]
 & BiCache$^\dagger$ & 0.50 & \textbf{66.9} \\
\addlinespace[1.5pt]
 & Full recomputation & 1 & 69.0 \\
\midrule
\multirow[c]{5}{*}{\raisebox{-0.35ex}{Dream}} & Direct reuse & 0 & 78.5 \\
\addlinespace[1.5pt]
 & \multirow[c]{2}{*}{ACache} & 0.2 & \textbf{79.5} \\
 & & 0.3 & \underline{78.8} \\
\addlinespace[1.5pt]
 & BiCache$^\dagger$ & 0.57 & 70.0 \\
\addlinespace[1.5pt]
 & Full recomputation & 1 & 79.4 \\
\bottomrule
\end{tabular}
\end{table*}

\begin{table*}[t]
\centering
\caption{\textbf{System-prompt reuse on GSM8K: end-to-end throughput (tokens/s).} Fast-dLLM is the full recomputation of Table~\ref{tab:system-prompt-accuracy}; ACache uses Anchor ratio 0.2; BiCache$^\dagger$ supports only batch size 1.}
\label{tab:system-prompt-throughput}
\EvalTableFont
\setlength{\tabcolsep}{2.5pt}
\begin{tabular}{L{\GridColModel}rrrrrrr}
\toprule
\TableHeadMultirow{Model} & \multicolumn{3}{c}{Batch 1} & \multicolumn{2}{c}{Batch 4} & \multicolumn{2}{c}{Batch 16} \\
\cmidrule(lr){2-4}\cmidrule(lr){5-6}\cmidrule(lr){7-8}
 & \SystemPerfHead{BiCache$^\dagger$} & \SystemPerfHead{Fast-dLLM} & \SystemPerfHead{ACache} & \SystemPerfHead{Fast-dLLM} & \SystemPerfHead{ACache} & \SystemPerfHead{Fast-dLLM} & \SystemPerfHead{ACache} \\
\midrule
LLaDA & \SystemPerfCell{62} & \SystemPerfCell{76} & \SystemPerfCell{91 \SystemPerfGain{1.19}} & \SystemPerfCell{138} & \SystemPerfCell{214 \SystemPerfGain{1.55}} & \SystemPerfCell{164} & \SystemPerfCell{290 \SystemPerfGain{1.77}} \\
Dream & \SystemPerfCell{92} & \SystemPerfCell{101} & \SystemPerfCell{126 \SystemPerfGain{1.26}} & \SystemPerfCell{167} & \SystemPerfCell{273 \SystemPerfGain{1.64}} & \SystemPerfCell{194} & \SystemPerfCell{357 \SystemPerfGain{1.84}} \\
\bottomrule
\end{tabular}
\end{table*}

Following BiCache~\citep{go2026bicache}, we reuse a system prompt as the shared span: exactly the GSM8K system prompt in its official implementation, which builds on the role and safety instructions of xAI's Grok~4 system prompt~\citep{xai2025grok-prompts} and asks for a final line ``\#\#\#\#~number''. As there, the user message places the question after ``Question:'' and ends with ``Answer:'' on a new line. The serialized system message is the shared prefix, 1,334 tokens under LLaDA's chat template and 1,270 under Dream's. We measure accuracy on the runtime itself over all 1,319 test questions, scored with lm-eval, at batch size 1 so that no other request shares a batch (Appendix~\ref{sec:appendix-experimental-setup}); on LLaDA, Anchor ratio 1 reproduces Fast-dLLM's responses on all 1,319 questions. BiCache runs as in Appendix~\ref{sec:appendix-baseline-selector-adaptations}.

Table~\ref{tab:system-prompt-accuracy} shows what direct reuse loses. On LLaDA, it scores 19.8\% against 69.0\% under full recomputation, and only 25\% of its responses contain the requested ``\#\#\#\#'' line, against 89\%: the stale cache fails to convey that instruction. Anchors restore most of the loss, reaching 59.1\% at ratio 0.2 and 65.4\% at 0.3; BiCache, at a mean equivalent ratio of 0.50, recomputes more and reaches 66.9\%, 1.6 points above ratio 0.3. On Dream, the stale cache costs little: every setting but BiCache stays within 1.0 point of full recomputation. BiCache instead loses accuracy: at ratio 0.57 it scores 70.0\%, 9.4 points below full recomputation and 8.5 below even direct reuse.

In Table~\ref{tab:system-prompt-throughput}, Fast-dLLM and ACache run on the runtime (Section~\ref{sec:system-design}). Even at batch size 1, where Anchor selection is least amortized (Section~\ref{sec:system-efficiency}), ACache runs at $1.19\times$ and $1.26\times$ Fast-dLLM's throughput on LLaDA and Dream, and at $1.46\times$ and $1.38\times$ BiCache's; at batch sizes 4 and 16 it reaches $1.55\times$ to $1.84\times$ Fast-dLLM's. At batch size 16 it exceeds Section~\ref{sec:system-efficiency}'s $1.71\times$ peak, as this shared prefix is longer than any few-shot affix there (at most 1,016 tokens). BiCache trails even Fast-dLLM: as in Table~\ref{tab:bicache-throughput-grid}, it runs in its open-source engine, so this compares whole systems.

%% file: ACache.bib
@inproceedings{zheng2024sglang,
author = {Zheng, Lianmin and Yin, Liangsheng and Xie, Zhiqiang and Sun, Chuyue and Huang, Jeff and Yu, Cody Hao and Cao, Shiyi and Kozyrakis, Christos and Stoica, Ion and Gonzalez, Joseph E. and Barrett, Clark and Sheng, Ying},
title = {{SGLang}: efficient execution of structured language model programs},
year = {2024},
isbn = {9798331314385},
publisher = {Curran Associates Inc.},
address = {Red Hook, NY, USA},
booktitle = {Proceedings of the 38th International Conference on Neural Information Processing Systems},
articleno = {2000},
numpages = {27},
location = {Vancouver, BC, Canada},
series = {NeurIPS '24}
}

@inproceedings{zhang2023eviction,
author = {Zhang, Zhenyu and Sheng, Ying and Zhou, Tianyi and Chen, Tianlong and Zheng, Lianmin and Cai, Ruisi and Song, Zhao and Tian, Yuandong and R\'{e}, Christopher and Barrett, Clark and Wang, Zhangyang and Chen, Beidi},
title = {{H2O}: heavy-hitter oracle for efficient generative inference of large language models},
year = {2023},
publisher = {Curran Associates Inc.},
address = {Red Hook, NY, USA},
booktitle = {Proceedings of the 37th International Conference on Neural Information Processing Systems},
articleno = {1506},
numpages = {50},
location = {New Orleans, LA, USA},
series = {NeurIPS '23}
}

@inproceedings{xiao2024streaming-llm,
title={Efficient Streaming Language Models with Attention Sinks},
author={Guangxuan Xiao and Yuandong Tian and Beidi Chen and Song Han and Mike Lewis},
booktitle={The Twelfth International Conference on Learning Representations},
year={2024},
url={https://openreview.net/forum?id=NG7sS51zVF}
}

@inproceedings{li2022continuous,
author = {Li, Xiang Lisa and Thickstun, John and Gulrajani, Ishaan and Liang, Percy and Hashimoto, Tatsunori B.},
title = {{Diffusion-LM} improves controllable text generation},
year = {2022},
isbn = {9781713871088},
publisher = {Curran Associates Inc.},
address = {Red Hook, NY, USA},
booktitle = {Proceedings of the 36th International Conference on Neural Information Processing Systems},
articleno = {313},
numpages = {16},
location = {New Orleans, LA, USA},
series = {NeurIPS '22}
}

@inproceedings{israel2025APD,
title={Accelerating Diffusion {LLM}s via Adaptive Parallel Decoding},
author={Daniel Mingyi Israel and Guy Van den Broeck and Aditya Grover},
booktitle={The Thirty-ninth Annual Conference on Neural Information Processing Systems},
year={2025},
url={https://openreview.net/forum?id=xwqTt26NJf}
}

@inproceedings{ben-hamu2025eb,
title={Accelerated Sampling from Masked Diffusion Models via Entropy Bounded Unmasking},
author={Heli Ben-Hamu and Itai Gat and Daniel Severo and Niklas Nolte and Brian Karrer},
booktitle={The Thirty-ninth Annual Conference on Neural Information Processing Systems},
year={2025},
url={https://openreview.net/forum?id=WBcBhT1NKO}
}

@article{ni2025data-learner,
  title={Diffusion language models are super data learners},
  author={Ni, Jinjie and Liu, Qian and Dou, Longxu and Du, Chao and Wang, Zili and Yan, Hang and Pang, Tianyu and Shieh, Michael Qizhe},
  journal={arXiv preprint arXiv:2511.03276},
  year={2025}
}

@article{he2026reasoning-with-latent,
  title={Reasoning with Latent Tokens in Diffusion Language Models},
  author={He, Andre and Welleck, Sean and Fried, Daniel},
  journal={arXiv preprint arXiv:2602.03769},
  year={2026}
}

@inproceedings{lou2024sedd,
author = {Lou, Aaron and Meng, Chenlin and Ermon, Stefano},
title = {Discrete diffusion modeling by estimating the ratios of the data distribution},
year = {2024},
publisher = {JMLR.org},
booktitle = {Proceedings of the 41st International Conference on Machine Learning},
articleno = {1333},
numpages = {30},
location = {Vienna, Austria},
series = {ICML'24}
}

@inproceedings{sahoo2024mdlm,
author = {Sahoo, Subham Sekhar and Arriola, Marianne and Schiff, Yair and Gokaslan, Aaron and Marroquin, Edgar and Chiu, Justin T and Rush, Alexander and Kuleshov, Volodymyr},
title = {Simple and effective masked diffusion language models},
year = {2024},
isbn = {9798331314385},
publisher = {Curran Associates Inc.},
address = {Red Hook, NY, USA},
booktitle = {Proceedings of the 38th International Conference on Neural Information Processing Systems},
articleno = {4135},
numpages = {49},
location = {Vancouver, BC, Canada},
series = {NeurIPS '24}
}

@article{ye2025dream,
  title={{Dream 7B}: Diffusion Large Language Models},
  author={Ye, Jiacheng and Xie, Zhihui and Zheng, Lin and Gao, Jiahui and Wu, Zirui and Jiang, Xin and Li, Zhenguo and Kong, Lingpeng},
  journal={arXiv preprint arXiv:2508.15487},
  year={2025}
}

@inproceedings{austin2021d3pm,
author = {Austin, Jacob and Johnson, Daniel D. and Ho, Jonathan and Tarlow, Daniel and van den Berg, Rianne},
title = {Structured denoising diffusion models in discrete state-spaces},
year = {2021},
isbn = {9781713845393},
publisher = {Curran Associates Inc.},
address = {Red Hook, NY, USA},
booktitle = {Proceedings of the 35th International Conference on Neural Information Processing Systems},
articleno = {1376},
numpages = {13},
series = {NeurIPS '21}
}

@inproceedings{nie2025llada,
  title={Large Language Diffusion Models},
  author={Shen Nie and Fengqi Zhu and Zebin You and Xiaolu Zhang and Jingyang Ou and Jun Hu and Jun Zhou and Yankai Lin and Ji-Rong Wen and Chongxuan Li},
  booktitle={Advances in Neural Information Processing Systems},
  year={2025},
}

@inproceedings{wu2026fast-dllm,
title={Fast-d{LLM}: Training-free Acceleration of Diffusion {LLM} by Enabling {KV} Cache and Parallel Decoding},
author={Chengyue Wu and Hao Zhang and Shuchen Xue and Zhijian Liu and Shizhe Diao and Ligeng Zhu and Ping Luo and Song Han and Enze Xie},
booktitle={The Fourteenth International Conference on Learning Representations},
year={2026},
url={https://openreview.net/forum?id=3Z3Is6hnOT}
}

@inproceedings{hu2026flash-dlm,
title={Flash{DLM}: Accelerating Diffusion Language Model Inference via Efficient {KV} Caching and Guided Diffusion},
author={Zhanqiu Hu and Jian Meng and Yash Akhauri and Mohamed S. Abdelfattah and Jae-sun Seo and Zhiru Zhang and Udit Gupta},
booktitle={The Fourteenth International Conference on Learning Representations},
year={2026},
url={https://openreview.net/forum?id=KUfKvlX3VY}
}

@inproceedings{song2026sparse-dllm,
  title={{Sparse-dLLM}: Accelerating diffusion llms with dynamic cache eviction},
  author={Song, Yuerong and Liu, Xiaoran and Li, Ruixiao and Liu, Zhigeng and Huang, Zengfeng and Guo, Qipeng and He, Ziwei and Qiu, Xipeng},
  booktitle={Proceedings of the AAAI Conference on Artificial Intelligence},
  volume={40},
  number={39},
  pages={33038--33046},
  year={2026}
}

@inproceedings{huang2026maskkv,
    title = "Mask Tokens as Prophet: Fine-Grained Cache Eviction for Efficient d{LLM} Inference",
    author = "Huang, Jianuo  and
      Zhang, Yaojie  and
      Yang, Yicun  and
      Huang, Benhao  and
      Zhang, Linfeng",
    editor = "Liakata, Maria  and
      Moreira, Viviane P.  and
      Zhang, Jiajun  and
      Jurgens, David",
    booktitle = "Findings of the {A}ssociation for {C}omputational {L}inguistics: {ACL} 2026",
    month = jul,
    year = "2026",
    address = "San Diego, California, United States",
    publisher = "Association for Computational Linguistics",
    url = "https://aclanthology.org/2026.findings-acl.170/",
    doi = "10.18653/v1/2026.findings-acl.170",
    pages = "3456--3479",
    ISBN = "979-8-89176-395-1"
}

@inproceedings{jiang2026d2cache,
title={d\${\textasciicircum}2\$Cache: Accelerating Diffusion-Based {LLM}s via Dual Adaptive Caching},
author={Yuchu Jiang and Yue Cai and Xiangzhong Luo and Jiale Fu and Jiarui Wang and Chonghan Liu and Xu Yang},
booktitle={The Fourteenth International Conference on Learning Representations},
year={2026},
url={https://openreview.net/forum?id=SjInfpK5RM}
}

@inproceedings{nguyen-tri2026attention,
title={Attention Is All You Need for {KV} Cache in Diffusion {LLM}s},
author={Quan Nguyen-Tri and Mukul Ranjan and Zhiqiang Shen},
booktitle={The Fourteenth International Conference on Learning Representations},
year={2026},
url={https://openreview.net/forum?id=zkUbhdAiFJ}
}

@inproceedings{lee2026dyllm,
title={Dy{LLM}: Efficient Diffusion {LLM} Inference via Saliency-based Token Selection and Partial Attention},
author={Younjoo Lee and Seungkyun Dan and Junghoo Lee and Jaiyoung Park and Jung Ho Ahn},
booktitle={Forty-third International Conference on Machine Learning},
year={2026},
url={https://openreview.net/forum?id=0azUrmsSyA}
}

@article{zuo2026window-diffusion,
  title={Window-Diffusion: Accelerating Diffusion Language Model Inference with Windowed Token Pruning and Caching},
  author={Zuo, Fengrui and Ke, Zhiwei and Liu, Yiming and Lou, Wenqi and Wang, Chao and Zhou, Xuehai},
  journal={arXiv preprint arXiv:2601.20332},
  year={2026}
}

@inproceedings{zhu2026es-dllm,
title={{ES}-d{LLM}: Efficient Inference for Diffusion Large Language Models by Early-Skipping},
author={Zijian Zhu and Fei Ren and Zhanhong Tan and Kaisheng Ma},
booktitle={The Fourteenth International Conference on Learning Representations},
year={2026},
url={https://openreview.net/forum?id=O2WvMkJbws}
}

@inproceedings{wu2026dynamic-dllm,
title={Dynamic-d{LLM}: Dynamic Cache-Budget and Adaptive Parallel Decoding for Training-Free Acceleration of Diffusion {LLM}},
author={Tianyi Wu and Xiaoxi Sun and Yanhua Jiao and Yulin Li and Yixin Chen and Yun-Hao Cao and Yi-Qi Hu and Zhuotao Tian},
booktitle={The Fourteenth International Conference on Learning Representations},
year={2026},
url={https://openreview.net/forum?id=SdnkB5pGbq}
}

@article{cheong2026entropy-cache,
  title={{EntropyCache}: Decoded Token Entropy Guided {KV} Caching for Diffusion Language Models},
  author={Cheong, Minsoo and Son, Donghyun and Lim, Woosang and Yoo, Sungjoo},
  journal={arXiv preprint arXiv:2603.18489},
  year={2026}
}

@inproceedings{liang2026focus,
title={{FOCUS}: {DLLM}s Know How to Tame Their Compute Bound},
author={Kaihua Liang and Xin Tan and An Zhong and Hong Xu and Marco Canini},
booktitle={Forty-third International Conference on Machine Learning},
year={2026},
url={https://openreview.net/forum?id=40fUEdwvH3}
}

@article{kong2025localleap,
  title={Accelerating diffusion llm inference via local determinism propagation},
  author={Kong, Fanheng and Zhang, Jingyuan and Liu, Yahui and Wu, Zirui and Tian, Yu and Zhou, Guorui and others},
  journal={arXiv preprint arXiv:2510.07081},
  year={2025}
}

@article{xiao2026streaming-dllm,
  title={Streaming-d{LLM}: Accelerating Diffusion LLMs via Suffix Pruning and Dynamic Decoding},
  author={Xiao, Zhongyu and Hao, Zhiwei and Guo, Jianyuan and Luo, Yong and Liu, Jia and Xu, Jie and Hu, Han},
  journal={arXiv preprint arXiv:2601.17917},
  year={2026}
}

@article{luo2026dsb,
  title={{DSB}: Dynamic Sliding Block Scheduling for Diffusion LLMs},
  author={Luo, Lizhuo and Li, Shenggui and Wen, Yonggang and Zhang, Tianwei},
  journal={arXiv preprint arXiv:2602.05992},
  year={2026}
}

@article{luo2026dawn,
  title={{DAWN}: Dependency-Aware Fast Inference for Diffusion LLMs},
  author={Luo, Lizhuo and Shi, Zhuoran and Luo, Jiajun and Wang, Zhi and Ren, Shen and Wang, Wenya and Zhang, Tianwei},
  journal={arXiv preprint arXiv:2602.06953},
  year={2026}
}

@article{gim2024prompt_cache,
  title={{Prompt Cache}: Modular attention reuse for low-latency inference},
  author={Gim, In and Chen, Guojun and Lee, Seung-seob and Sarda, Nikhil and Khandelwal, Anurag and Zhong, Lin},
  journal={Proceedings of Machine Learning and Systems},
  volume={6},
  pages={325--338},
  year={2024}
}

@inproceedings{srivatsa2025preble,
title={Preble: Efficient Distributed Prompt Scheduling for {LLM} Serving},
author={Vikranth Srivatsa and Zijian He and Reyna Abhyankar and Dongming Li and Yiying Zhang},
booktitle={The Thirteenth International Conference on Learning Representations},
year={2025},
url={https://openreview.net/forum?id=meKEKDhdnx}
}

@inproceedings{pan2025fasttree,
 author = {Pan, Zaifeng and Ding, Yitong and Guan, Yue and Wang, Zheng and Yu, Zhongkai and Tang, Xulong and Wang, Yida and Ding, Yufei},
 booktitle = {Proceedings of Machine Learning and Systems},
 editor = {M. Zaharia and G. Joshi and Y. Lin},
 pages = {},
 publisher = {MLSys},
 title = {{FastTree}: Optimizing Attention Kernel and Runtime for Tree-Structured LLM Inference},
 url = {https://proceedings.mlsys.org/paper_files/paper/2025/file/96894468eb44631a32d7ebd56f9892c7-Paper-Conference.pdf},
 volume = {7},
 year = {2025}
}

@inproceedings{yuan2026dualmap,
title={{DualMap}: Enabling Both Cache Affinity and Load Balancing for Distributed {LLM} Serving},
author={Ying Yuan and Pengfei Zuo and Bo Wang and Zhangyu Chen and Zhipeng Tan and Zhou Yu},
booktitle={The Fourteenth International Conference on Learning Representations},
year={2026},
url={https://openreview.net/forum?id=zCadrJ32Xn}
}

@inproceedings{yao2025cache-blend,
  title={{CacheBlend}: Fast large language model serving for rag with cached knowledge fusion},
  author={Yao, Jiayi and Li, Hanchen and Liu, Yuhan and Ray, Siddhant and Cheng, Yihua and Zhang, Qizheng and Du, Kuntai and Lu, Shan and Jiang, Junchen},
  booktitle={Proceedings of the twentieth European conference on computer systems},
  pages={94--109},
  year={2025}
}

@article{zhang2026sparsex,
  title={{SparseX}: Efficient Segment-Level KV Cache Sharing for Interleaved LLM Serving},
  author={Zhang, Quqing and Chen, Kai and Liao, Ning and Lin, Zehao and Tang, Bo and Xiong, Feiyu and Li, Zhiyu and Wang, Xiaoxing},
  journal={arXiv preprint arXiv:2606.01751},
  year={2026}
}

@article{liu2026redknot,
  title={{RedKnot}: Efficient Long-Context LLM Serving with Head-Aware KV Reuse and SegPagedAttention},
  author={Liu, Yang and Luo, Zhaokai and Jin, Huayi and Wang, Zhiyong and He, Ruozhou and Wang, Boyu and Chen, Guanjie and Xie, Tao and Hu, Junhao},
  journal={arXiv preprint arXiv:2606.06256},
  year={2026}
}

@article{li2026cachetune,
  title={Adaptive KV Cache Reuse for Fast Long-Context LLM Serving},
  author={Liu, Song and Liu, Yan and Cui, Jinhua and Nie, Shiqiang and Wang, Jinyu and Wu, Weiguo and others},
  journal={arXiv preprint arXiv:2605.24022},
  year={2026}
}

@article{xu2026pill,
  title={Predict, Don't Iterate: Efficient Adaptive-Length Infilling for Diffusion Language Models},
  author={Xu, Haobo and Chen, Sirui and Bei, Yuanchen and Chen, Lingjie and Yan, Yuchen and Fu, Dongqi and He, Jingrui and Tong, Hanghang},
  journal={arXiv preprint arXiv:2609.02108},
  year={2026}
}

@inproceedings{wang2026prophetkv,
title={Prophet{KV}: User-Query-Driven Selective Recomputation for Efficient {KV} Cache Reuse in Retrieval-Augmented Generation},
author={Shihao Wang and Jiahao Chen and Yanqi Pan and Hao Huang and Yichen Hao and Xiangyu Zou and Wen Xia and Chongyang Qiu and Wentao Zhang and PengFei Wang},
booktitle={Forty-third International Conference on Machine Learning},
year={2026},
url={https://openreview.net/forum?id=SwGwmKIgUb}
}

@inproceedings{ma2025dkv-cache,
  title={{dKV-Cache}: The Cache for Diffusion Language Models},
  author={Xinyin Ma and Runpeng Yu and Gongfan Fang and Xinchao Wang},
  booktitle={Advances in Neural Information Processing Systems},
  year={2025}
}

@inproceedings{liu2026dllm-cache,
title={d{LLM}-Cache: Accelerating Diffusion Large Language Models with Adaptive Caching},
author={Zhiyuan Liu and Yicun Yang and Yaojie Zhang and Junjie Chen and Chang Zou and Qingyan Wei and Shaobo Wang and Yichen Zhu and Linfeng Zhang},
booktitle={Forty-third International Conference on Machine Learning},
year={2026},
url={https://openreview.net/forum?id=guLrgDb2jB}
}

@misc{geeeekexplorer2025NanoVLLM,
  author={{Nano-vLLM}},
  title={{Nano-vLLM}: A lightweight implementation of {vLLM}},
  year={2025},
  howpublished={\url{https://github.com/GeeeekExplorer/Nano-vLLM}},
  note={GitHub repository}
}

@misc{eval-harness,
  author       = {Gao, Leo and Tow, Jonathan and Abbasi, Baber and Biderman, Stella and Black, Sid and DiPofi, Anthony and Foster, Charles and Golding, Laurence and Hsu, Jeffrey and Le Noac'h, Alain and Li, Haonan and McDonell, Kyle and Muennighoff, Niklas and Ociepa, Chris and Phang, Jason and Reynolds, Laria and Schoelkopf, Hailey and Skowron, Aviya and Sutawika, Lintang and Tang, Eric and Thite, Anish and Wang, Ben and Wang, Kevin and Zou, Andy},
  title        = {The Language Model Evaluation Harness},
  month        = 07,
  year         = 2024,
  publisher    = {Zenodo},
  version      = {v0.4.3},
  doi          = {10.5281/zenodo.12608602},
  url          = {https://zenodo.org/records/12608602}
}

@article{ma2025dinfer,
  title={{dInfer}: An efficient inference framework for diffusion language models},
  author={Ma, Yuxin and Du, Lun and Wei, Lanning and Chen, Kun and Xu, Qian and Wang, Kangyu and Feng, Guofeng and Lu, Guoshan and Liu, Lin and Qi, Xiaojing and others},
  journal={arXiv preprint arXiv:2510.08666},
  year={2025}
}

@inproceedings{long2026focus-dllm,
    title = "Focus-d{LLM}: Accelerating Long-Context Diffusion {LLM} Inference via Confidence-Guided Context Focusing",
    author = "Long, Lingkun  and
      Huang, Yushi  and
      Bai, Shihao  and
      Gong, Ruihao  and
      Zhang, Jun  and
      Zhou, Ao  and
      Yang, Jianlei",
    editor = "Liakata, Maria  and
      Moreira, Viviane P.  and
      Zhang, Jiajun  and
      Jurgens, David",
    booktitle = "Proceedings of the 64th Annual Meeting of the {A}ssociation for {C}omputational {L}inguistics (Volume 1: Long Papers)",
    month = jul,
    year = "2026",
    address = "San Diego, California, United States",
    publisher = "Association for Computational Linguistics",
    url = "https://aclanthology.org/2026.acl-long.556/",
    doi = "10.18653/v1/2026.acl-long.556",
    pages = "12146--12160",
    ISBN = "979-8-89176-390-6"
}

@article{frumkin2026dare,
  title={DARE: Diffusion Language Model Activation Reuse for Efficient Inference},
  author={Frumkin, Natalia and Wang, Bokun and Chiang, Hung-Yueh and Chang, Chi-Chih and Abdelfattah, Mohamed S and Marculescu, Diana},
  journal={arXiv preprint arXiv:2605.08134},
  year={2026}
}

@article{xiong2026prefilling-dllm,
  title={{Prefilling-dLLM}: Predictive Prefilling for Long-Context Inference in Diffusion Language Models},
  author={Xiong, Jing and Han, Qi and Gong, Shansan and Hsieh, Yunta and Zheng, Boyuan and Wu, Chengyue and Tao, Chaofan and Zhao, Chenyang and Wong, Ngai},
  journal={arXiv preprint arXiv:2606.10537},
  year={2026}
}

@inproceedings{li2026positional-bias,
  title={Where to Place the Query? Unveiling and Mitigating Positional Bias in In-Context Learning for Diffusion LLMs via Decoding Dynamics},
  author={Li, Zhengheng and Li, Panrui and Liu, Xuyang and Xia, Puzhi},
  booktitle={International Conference on Intelligent Computing},
  pages={602--612},
  year={2026},
  organization={Springer}
}

@article{yang2026wavefilter,
  title={WaveFilter: Enhancing the Long-Context Capability of Diffusion LLMs via Wavelet-Guided KV Cache Filtering},
  author={Yang, Jinnan and Wang, Yan and Bi, Zhen and Wu, Kehao and Li, Xiaojie and Lou, Jungang and Li, Zechao and Liu, Jing},
  journal={arXiv preprint arXiv:2606.00724},
  year={2026}
}

@article{wu2026elastic-dllm,
  title={Elastic-dLLM: Position Preserving Context Compression and Augmentation of Diffusion LLMs},
  author={Wu, Junyi and Zhao, Tianchen and Zhang, Shaoqiu and Zhang, Linfeng and Dai, Guohao and Wang, Yu},
  journal={arXiv preprint arXiv:2605.18165},
  year={2026}
}

@article{go2026bicache,
  title={Enabling KV Caching of Shared Prefix for Diffusion Language Models},
  author={Go, Younghun and Han, Jaehoon and Shin, Changyong and Yoo, Chuck and Yang, Gyeongsik},
  journal={arXiv preprint arXiv:2606.07571},
  year={2026}
}

@inproceedings{pope2023efficiently,
 author = {Pope, Reiner and Douglas, Sholto and Chowdhery, Aakanksha and Devlin, Jacob and Bradbury, James and Heek, Jonathan and Xiao, Kefan and Agrawal, Shivani and Dean, Jeff},
 booktitle = {Proceedings of Machine Learning and Systems},
 editor = {D. Song and M. Carbin and T. Chen},
 pages = {606--624},
 publisher = {Curan},
 title = {Efficiently Scaling Transformer Inference},
 url = {https://proceedings.mlsys.org/paper_files/paper/2023/file/c4be71ab8d24cdfb45e3d06dbfca2780-Paper-mlsys2023.pdf},
 volume = {5},
 year = {2023}
}

@inproceedings{kwon2023vllm,
  title={Efficient memory management for large language model serving with {PagedAttention}},
  author={Kwon, Woosuk and Li, Zhuohan and Zhuang, Siyuan and Sheng, Ying and Zheng, Lianmin and Yu, Cody Hao and Gonzalez, Joseph and Zhang, Hao and Stoica, Ion},
  booktitle={Proceedings of the 29th symposium on operating systems principles},
  pages={611--626},
  year={2023}
}

@inproceedings {yu2022orca,
author = {Gyeong-In Yu and Joo Seong Jeong and Geon-Woo Kim and Soojeong Kim and Byung-Gon Chun},
title = {Orca: A Distributed Serving System for {Transformer-Based} Generative Models},
booktitle = {16th USENIX Symposium on Operating Systems Design and Implementation (OSDI 22)},
year = {2022},
isbn = {978-1-939133-28-1},
address = {Carlsbad, CA},
pages = {521--538},
url = {https://www.usenix.org/conference/osdi22/presentation/yu},
publisher = {USENIX Association},
month = jul
}

@article{cobbe2021gsm8k,
  title={Training verifiers to solve math word problems},
  author={Cobbe, Karl and Kosaraju, Vineet and Bavarian, Mohammad and Chen, Mark and Jun, Heewoo and Kaiser, Lukasz and Plappert, Matthias and Tworek, Jerry and Hilton, Jacob and Nakano, Reiichiro and others},
  journal={arXiv preprint arXiv:2110.14168},
  year={2021}
}

@article{austin2021mbpp,
  title={Program synthesis with large language models},
  author={Austin, Jacob and Odena, Augustus and Nye, Maxwell and Bosma, Maarten and Michalewski, Henryk and Dohan, David and Jiang, Ellen and Cai, Carrie and Terry, Michael and Le, Quoc and others},
  journal={arXiv preprint arXiv:2108.07732},
  year={2021}
}

@article{kuratov2024babilong,
  title={{BABILong}: Testing the limits of llms with long context reasoning-in-a-haystack},
  author={Kuratov, Yuri and Bulatov, Aydar and Anokhin, Petr and Rodkin, Ivan and Sorokin, Dmitry and Sorokin, Artyom and Burtsev, Mikhail},
  journal={Advances in Neural Information Processing Systems},
  volume={37},
  pages={106519--106554},
  year={2024}
}

@inproceedings{lu2026adablock-dllm,
title={{AdaBlock}-d{LLM}: Semantic-Aware Diffusion {LLM} Inference via Adaptive Block Size},
author={Guanxi Lu and Hao Mark Chen and Yuto Karashima and Zhican Wang and Daichi Fujiki and Hongxiang Fan},
booktitle={The Fourteenth International Conference on Learning Representations},
year={2026},
url={https://openreview.net/forum?id=0Cv9PwL7cI}
}

@article{ouyang2026unicache,
author = {Ouyang, Bei and Qiao, Yifan and Xing, Jiarong},
title = {UniCache: Unifying Prefix Cache Eviction for Heterogeneous LLM Serving Workloads},
year = {2026},
issue_date = {June 2026},
publisher = {Association for Computing Machinery},
address = {New York, NY, USA},
volume = {10},
number = {2},
url = {https://doi.org/10.1145/3805652},
doi = {10.1145/3805652},
journal = {Proc. ACM Meas. Anal. Comput. Syst.},
month = may,
articleno = {54},
numpages = {27}
}

@inproceedings {qin2025mooncake,
author = {Ruoyu Qin and Zheming Li and Weiran He and Jialei Cui and Feng Ren and Mingxing Zhang and Yongwei Wu and Weimin Zheng and Xinran Xu},
title = {{Mooncake}: Trading More Storage for Less Computation {\textemdash} A {KVCache-centric} Architecture for Serving {LLM} Chatbot},
booktitle = {23rd USENIX Conference on File and Storage Technologies (FAST 25)},
year = {2025},
isbn = {978-1-939133-45-8},
address = {Santa Clara, CA},
pages = {155--170},
url = {https://www.usenix.org/conference/fast25/presentation/qin},
publisher = {USENIX Association},
month = feb
}

@inproceedings{ye2024chunkattention,
    title = "{C}hunk{A}ttention: Efficient Self-Attention with Prefix-Aware {KV} Cache and Two-Phase Partition",
    author = "Ye, Lu  and
      Tao, Ze  and
      Huang, Yong  and
      Li, Yang",
    editor = "Ku, Lun-Wei  and
      Martins, Andre  and
      Srikumar, Vivek",
    booktitle = "Proceedings of the 62nd Annual Meeting of the Association for Computational Linguistics (Volume 1: Long Papers)",
    month = aug,
    year = "2024",
    address = "Bangkok, Thailand",
    publisher = "Association for Computational Linguistics",
    url = "https://aclanthology.org/2024.acl-long.623/",
    doi = "10.18653/v1/2024.acl-long.623",
    pages = "11608--11620"
}

@misc{xai2025grok-prompts,
  author={{xAI}},
  title={grok-prompts: Prompts for the {Grok} chat assistant and the @grok bot on {X}},
  year={2025},
  howpublished={\url{https://github.com/xai-org/grok-prompts}},
  note={GitHub repository}
}

@misc{vllm-omni2026,
  author={{vLLM-Omni}},
  title={Continuous Batching for {Step-Wise} Diffusion},
  year={2026},
  howpublished={\url{https://docs.vllm.ai/projects/vllm-omni/en/v0.20.0/design/feature/diffusion_continuous_batching/}},
  note={vLLM-Omni documentation}
}

@inproceedings{hu2025epic,
title={{EPIC}: Efficient Position-Independent Caching for Serving Large Language Models},
author={Junhao Hu and Wenrui Huang and Weidong Wang and Haoyi Wang and tiancheng hu and zhang qin and Hao Feng and Xusheng Chen and Yizhou Shan and Tao Xie},
booktitle={Forty-second International Conference on Machine Learning},
year={2025},
url={https://openreview.net/forum?id=qjd3ZUiHRT}
}

@inproceedings{sohl-dickstein2015diffusion,
author = {Sohl-Dickstein, Jascha and Weiss, Eric A. and Maheswaranathan, Niru and Ganguli, Surya},
title = {Deep unsupervised learning using nonequilibrium thermodynamics},
year = {2015},
publisher = {JMLR.org},
booktitle = {Proceedings of the 32nd International Conference on Machine Learning - Volume 37},
pages = {2256–2265},
numpages = {10},
location = {Lille, France},
series = {ICML'15}
}

@inproceedings{ho2020ddpm,
author = {Ho, Jonathan and Jain, Ajay and Abbeel, Pieter},
title = {Denoising diffusion probabilistic models},
year = {2020},
isbn = {9781713829546},
publisher = {Curran Associates Inc.},
address = {Red Hook, NY, USA},
booktitle = {Proceedings of the 34th International Conference on Neural Information Processing Systems},
articleno = {574},
numpages = {12},
location = {Vancouver, BC, Canada},
series = {NIPS '20}
}

@inproceedings{hoogeboom2021multinomial,
author = {Hoogeboom, Emiel and Nielsen, Didrik and Jaini, Priyank and Forr{\'e}, Patrick and Welling, Max},
title = {Argmax flows and multinomial diffusion: learning categorical distributions},
year = {2021},
isbn = {9781713845393},
publisher = {Curran Associates Inc.},
address = {Red Hook, NY, USA},
booktitle = {Proceedings of the 35th International Conference on Neural Information Processing Systems},
articleno = {953},
numpages = {12},
series = {NIPS '21}
}

@inproceedings{ou2025radd,
title={Your Absorbing Discrete Diffusion Secretly Models the Conditional Distributions of Clean Data},
author={Jingyang Ou and Shen Nie and Kaiwen Xue and Fengqi Zhu and Jiacheng Sun and Zhenguo Li and Chongxuan Li},
booktitle={The Thirteenth International Conference on Learning Representations},
year={2025},
url={https://openreview.net/forum?id=sMyXP8Tanm}
}

@article{inception2025mercury,
  title={Mercury: Ultra-fast language models based on diffusion},
  author={{Inception Labs} and Khanna, Samar and Kharbanda, Siddhant and Li, Shufan and Varma, Harshit and Wang, Eric and Birnbaum, Sawyer and Luo, Ziyang and Miraoui, Yanis and Palrecha, Akash and others},
  journal={arXiv preprint arXiv:2506.17298},
  year={2025}
}

@article{song2025seed-diffusion,
  title={{Seed Diffusion}: A large-scale diffusion language model with high-speed inference},
  author={Song, Yuxuan and Zhang, Zheng and Luo, Cheng and Gao, Pengyang and Xia, Fan and Luo, Hao and Li, Zheng and Yang, Yuehang and Yu, Hongli and Qu, Xingwei and others},
  journal={arXiv preprint arXiv:2508.02193},
  year={2025}
}

@inproceedings{tae2025tess2,
    title = "{TESS} 2: A Large-Scale Generalist Diffusion Language Model",
    author = "Tae, Jaesung  and
      Ivison, Hamish  and
      Kumar, Sachin  and
      Cohan, Arman",
    editor = "Che, Wanxiang  and
      Nabende, Joyce  and
      Shutova, Ekaterina  and
      Pilehvar, Mohammad Taher",
    booktitle = "Proceedings of the 63rd Annual Meeting of the Association for Computational Linguistics (Volume 1: Long Papers)",
    month = jul,
    year = "2025",
    address = "Vienna, Austria",
    publisher = "Association for Computational Linguistics",
    url = "https://aclanthology.org/2025.acl-long.1029/",
    doi = "10.18653/v1/2025.acl-long.1029",
    pages = "21171--21188",
    ISBN = "979-8-89176-251-0"
}

@article{nie2026illada,
  title={Improved Large Language Diffusion Models},
  author={Nie, Shen and Min, Qiyang and Xu, Shaoxuan and Huang, Zihao and Song, Yuxuan and Shan, Yong and Lin, Yankai and Zhao, Wayne Xin and Li, Chongxuan and Wen, Ji-Rong},
  journal={arXiv preprint arXiv:2606.25331},
  year={2026}
}

@inproceedings{zhu2026llada-1.5,
    title = "{LL}a{DA} 1.5: Variance-Reduced Preference Optimization for Large Language Diffusion Models",
    author = "Zhu, Fengqi  and
      Wang, Rongzhen  and
      Nie, Shen  and
      Zhang, Xiaolu  and
      Wu, Chunwei  and
      Zhou, Jun  and
      Lin, Yankai  and
      Wen, Ji-Rong  and
      Li, Chongxuan",
    editor = "Liakata, Maria  and
      Moreira, Viviane P.  and
      Zhang, Jiajun  and
      Jurgens, David",
    booktitle = "Proceedings of the 64th Annual Meeting of the {A}ssociation for {C}omputational {L}inguistics (Volume 1: Long Papers)",
    month = jul,
    year = "2026",
    address = "San Diego, California, United States",
    publisher = "Association for Computational Linguistics",
    url = "https://aclanthology.org/2026.acl-long.524/",
    doi = "10.18653/v1/2026.acl-long.524",
    pages = "11425--11460",
    ISBN = "979-8-89176-390-6"
}

@article{zhu2025llada-moe,
  title={LLaDA-MoE: A sparse MoE diffusion language model},
  author={Zhu, Fengqi and You, Zebin and Xing, Yipeng and Huang, Zenan and Liu, Lin and Zhuang, Yihong and Lu, Guoshan and Wang, Kangyu and Wang, Xudong and Wei, Lanning and others},
  journal={arXiv preprint arXiv:2509.24389},
  year={2025}
}

@inproceedings{fu2026efficient-dlm,
title={Efficient-{DLM}: From Autoregressive to Diffusion Language Models, and Beyond in Speed},
author={Yonggan Fu and Lexington Whalen and Zhifan Ye and Xin Dong and Shizhe Diao and Jingyu Liu and Chengyue Wu and Hao Zhang and Enze Xie and Song Han and Maksim Khadkevich and Jan Kautz and Yingyan Celine Lin and Pavlo Molchanov},
booktitle={Forty-third International Conference on Machine Learning},
year={2026},
url={https://openreview.net/forum?id=ynRnOqI2W8}
}

@inproceedings{wu2026fast-dllm-v2,
title={Fast-d{LLM} v2: Efficient Block-Diffusion {LLM}},
author={Chengyue Wu and Hao Zhang and Shuchen Xue and Shizhe Diao and Yonggan Fu and Zhijian Liu and Pavlo Molchanov and Ping Luo and Song Han and Enze Xie},
booktitle={The Fourteenth International Conference on Learning Representations},
year={2026},
url={https://openreview.net/forum?id=1NZ3DHF9nT}
}

@article{diffusiongemma2026,
  title={DiffusionGemma Technical Report},
  author={{DiffusionGemma Team} and Ta{\"\i}ga, Adrien Ali and Assiene, James and Calandriello, Daniele and Chaabouni, Rahma and Gante, Jo{\~a}o and von Glehn, Tamara and Keating, Nate and Knutsen, Chris and Kukla, Martin and others},
  journal={arXiv preprint arXiv:2608.00146},
  year={2026}
}

@inproceedings{han2023ssd-lm,
    title = "{SSD}-{LM}: Semi-autoregressive Simplex-based Diffusion Language Model for Text Generation and Modular Control",
    author = "Han, Xiaochuang  and
      Kumar, Sachin  and
      Tsvetkov, Yulia",
    editor = "Rogers, Anna  and
      Boyd-Graber, Jordan  and
      Okazaki, Naoaki",
    booktitle = "Proceedings of the 61st Annual Meeting of the Association for Computational Linguistics (Volume 1: Long Papers)",
    month = jul,
    year = "2023",
    address = "Toronto, Canada",
    publisher = "Association for Computational Linguistics",
    url = "https://aclanthology.org/2023.acl-long.647/",
    doi = "10.18653/v1/2023.acl-long.647",
    pages = "11575--11596"
}

@inproceedings{arriola2025block-diffusion,
title={Block Diffusion: Interpolating Between Autoregressive and Diffusion Language Models},
author={Marianne Arriola and Subham Sekhar Sahoo and Aaron Gokaslan and Zhihan Yang and Zhixuan Qi and Jiaqi Han and Justin T Chiu and Volodymyr Kuleshov},
booktitle={The Thirteenth International Conference on Learning Representations},
year={2025},
url={https://openreview.net/forum?id=tyEyYT267x}
}

@inproceedings{gong2025diffullama,
title={Scaling Diffusion Language Models via Adaptation from Autoregressive Models},
author={Shansan Gong and Shivam Agarwal and Yizhe Zhang and Jiacheng Ye and Lin Zheng and Mukai Li and Chenxin An and Peilin Zhao and Wei Bi and Jiawei Han and Hao Peng and Lingpeng Kong},
booktitle={The Thirteenth International Conference on Learning Representations},
year={2025},
url={https://openreview.net/forum?id=j1tSLYKwg8}
}

@inproceedings{nie2025smdm,
title={Scaling up Masked Diffusion Models on Text},
author={Shen Nie and Fengqi Zhu and Chao Du and Tianyu Pang and Qian Liu and Guangtao Zeng and Min Lin and Chongxuan Li},
booktitle={The Thirteenth International Conference on Learning Representations},
year={2025},
url={https://openreview.net/forum?id=WNvvwK0tut}
}

@inproceedings{shi2024md4,
title={Simplified and Generalized Masked Diffusion for Discrete Data},
author={Jiaxin Shi and Kehang Han and Zhe Wang and Arnaud Doucet and Michalis Titsias},
booktitle={The Thirty-eighth Annual Conference on Neural Information Processing Systems},
year={2024},
url={https://openreview.net/forum?id=xcqSOfHt4g}
}

@inproceedings{cheng2026sdar,
    title = "{SDAR}: A Synergistic Diffusion-{A}uto{R}egression Paradigm for Scalable Sequence Generation",
    author = "Cheng, Shuang  and
      Bian, Yihan  and
      Liu, Dawei  and
      Jiang, Yuhua  and
      Liu, Yihao  and
      Zhang, Linfeng  and
      Yao, Qian  and
      Tian, Zhongbo  and
      Wang, Wenhai  and
      Guo, Qipeng  and
      Chen, Kai  and
      Qi, Biqing  and
      Zhou, Bowen",
    editor = "Liakata, Maria  and
      Moreira, Viviane P.  and
      Zhang, Jiajun  and
      Jurgens, David",
    booktitle = "Findings of the {A}ssociation for {C}omputational {L}inguistics: {ACL} 2026",
    month = jul,
    year = "2026",
    address = "San Diego, California, United States",
    publisher = "Association for Computational Linguistics",
    url = "https://aclanthology.org/2026.findings-acl.1110/",
    doi = "10.18653/v1/2026.findings-acl.1110",
    pages = "22058--22075",
    ISBN = "979-8-89176-395-1"
}

@article{bie2025llada2,
  title={{LLaDA2.0}: Scaling Up Diffusion Language Models to 100{B}},
  author={Bie, Tiwei and Cao, Maosong and Chen, Kun and Du, Lun and Gong, Mingliang and Gong, Zhuochen and Gu, Yanmei and Hu, Jiaqi and Huang, Zenan and Lan, Zhenzhong and others},
  journal={arXiv preprint arXiv:2512.15745},
  year={2025}
}

@inproceedings{pang2024anllm,
    title = "Anchor-based Large Language Models",
    author = "Pang, Jianhui  and
      Ye, Fanghua  and
      Wong, Derek  and
      He, Xin  and
      Chen, Wanshun  and
      Wang, Longyue",
    editor = "Ku, Lun-Wei  and
      Martins, Andre  and
      Srikumar, Vivek",
    booktitle = "Findings of the Association for Computational Linguistics: ACL 2024",
    month = aug,
    year = "2024",
    address = "Bangkok, Thailand",
    publisher = "Association for Computational Linguistics",
    url = "https://aclanthology.org/2024.findings-acl.295/",
    doi = "10.18653/v1/2024.findings-acl.295",
    pages = "4958--4976"
}

@article{yao2026follow,
  title={Follow the Latent Roadmap: Navigating Revocable Decoding for Diffusion LLMs with Anchor Tokens},
  author={Yao, Yizhen and Zhu, Qinglin and Zhao, Runcong and Dai, Xiangxiang and Xiang, Yanzheng and He, Yulan and Gui, Lin},
  journal={arXiv preprint arXiv:2606.16847},
  year={2026}
}

@article{kedia2026sangam,
  title={Sangam: Efficiently Serving Diffusion LLMs with the AR Stack},
  author={Kedia, Nitin and Agarwal, Saurabh and Lee, Myungjin and Akella, Aditya},
  journal={arXiv preprint arXiv:2607.04206},
  year={2026}
}

@article{fan2025dllm-serve,
  title={Taming the memory footprint crisis: System design for production diffusion llm serving},
  author={Fan, Jiakun and Zhang, Yanglin and Li, Xiangchen and Nikolopoulos, Dimitrios S},
  journal={arXiv preprint arXiv:2512.17077},
  year={2025}
}

@inproceedings{dao2022flashattention,
author = {Dao, Tri and Fu, Daniel Y. and Ermon, Stefano and Rudra, Atri and R{\'e}, Christopher},
title = {FLASHATTENTION: fast and memory-efficient exact attention with IO-awareness},
year = {2022},
isbn = {9781713871088},
publisher = {Curran Associates Inc.},
address = {Red Hook, NY, USA},
booktitle = {Proceedings of the 36th International Conference on Neural Information Processing Systems},
articleno = {1189},
numpages = {16},
location = {New Orleans, LA, USA},
series = {NIPS '22}
}

@inproceedings{liu2023scissorhands,
 author = {Liu, Zichang and Desai, Aditya and Liao, Fangshuo and Wang, Weitao and Xie, Victor and Xu, Zhaozhuo and Kyrillidis, Anastasios and Shrivastava, Anshumali},
 booktitle = {Advances in Neural Information Processing Systems},
 doi = {10.52202/075280-2279},
 editor = {A. Oh and T. Naumann and A. Globerson and K. Saenko and M. Hardt and S. Levine},
 pages = {52342--52364},
 publisher = {Curran Associates, Inc.},
 title = {Scissorhands: Exploiting the Persistence of Importance Hypothesis for LLM KV Cache Compression at Test Time},
 url = {https://proceedings.neurips.cc/paper_files/paper/2023/file/a452a7c6c463e4ae8fbdc614c6e983e6-Paper-Conference.pdf},
 volume = {36},
 year = {2023}
}

@inproceedings{li2024snapkv,
 author = {Li, Yuhong and Huang, Yingbing and Yang, Bowen and Venkitesh, Bharat and Locatelli, Acyr and Ye, Hanchen and Cai, Tianle and Lewis, Patrick and Chen, Deming},
 booktitle = {Advances in Neural Information Processing Systems},
 doi = {10.52202/079017-0722},
 editor = {A. Globerson and L. Mackey and D. Belgrave and A. Fan and U. Paquet and J. Tomczak and C. Zhang},
 pages = {22947--22970},
 publisher = {Curran Associates, Inc.},
 title = {SnapKV: LLM Knows What You are Looking for Before Generation},
 url = {https://proceedings.neurips.cc/paper_files/paper/2024/file/28ab418242603e0f7323e54185d19bde-Paper-Conference.pdf},
 volume = {37},
 year = {2024}
}

@article{su2024roformer,
title = {RoFormer: Enhanced transformer with Rotary Position Embedding},
journal = {Neurocomputing},
volume = {568},
pages = {127063},
year = {2024},
issn = {0925-2312},
doi = {https://doi.org/10.1016/j.neucom.2023.127063},
url = {https://www.sciencedirect.com/science/article/pii/S0925231223011864},
author = {Jianlin Su and Murtadha Ahmed and Yu Lu and Shengfeng Pan and Wen Bo and Yunfeng Liu}
}

@inproceedings{tillet2019triton,
author = {Tillet, Philippe and Kung, H. T. and Cox, David},
title = {Triton: an intermediate language and compiler for tiled neural network computations},
year = {2019},
isbn = {9781450367196},
publisher = {Association for Computing Machinery},
address = {New York, NY, USA},
url = {https://doi.org/10.1145/3315508.3329973},
doi = {10.1145/3315508.3329973},
booktitle = {Proceedings of the 3rd ACM SIGPLAN International Workshop on Machine Learning and Programming Languages},
pages = {10–19},
numpages = {10},
location = {Phoenix, AZ, USA},
series = {MAPL 2019}
}

@article{kasa2026fast-dllm-pp,
  title={Fast-dLLM++: Fr{\'e}chet Profile Decoding for Faster Diffusion LLM Inference},
  author={Kasa, Siva Rajesh and Dai, Yasong and Negi, Sumit and Li, Hongdong},
  journal={arXiv preprint arXiv:2606.02955},
  year={2026}
}

@article{lee2026polestar,
  title={Polestar: Drift-Aware Cache Calibration and Token Commitment for Efficient Inference of Diffusion LLMs},
  author={Lee, Mingyu and Ramachandran, Akshat and Kundu, Souvik and Krishna, Tushar},
  journal={arXiv preprint arXiv:2607.14107},
  year={2026}
}

@article{he2026archer,
  title={Archer: Adaptive Reuse of Cached Hidden States for Efficient Rollback in Diffusion Language Models},
  author={He, Xuning and Sheng, Zinan and Tao, Yongding and Liu, Huanyu and Li, Ge and Jiang, Xue and Dong, Yihong},
  journal={arXiv preprint arXiv:2608.08086},
  year={2026}
}

@article{kwon2026herald,
  title={{HERALD}: High-Throughput Block Diffusion LLM Serving via CPU-GPU Cooperative KV Cache Retrieval},
  author={Kwon, Omin and Kim, Doyeon and Park, Jongseok and Lee, Seung Yul and Stoica, Ion and Lee, Jae W},
  journal={arXiv preprint arXiv:2606.21633},
  year={2026}
}

@article{wang2026groupkv,
  title={{GroupKV}: Hierarchical KV Cache Management for Long-Context Diffusion LLM Inference},
  author={Wang, Jinhao and Hu, Zhexin and Zhou, Kangjie and Zhou, Xin and Liu, Fangfang},
  journal={arXiv preprint arXiv:2609.17573},
  year={2026}
}

@inproceedings{chen2026dpad,
title={{DP}ad: Efficient Diffusion Language Models with Suffix Dropout},
author={Xinhua Chen and Sitao Huang and Cong Guo and Chiyue Wei and Yintao He and Jianyi Zhang and Hai Helen Li and Yiran Chen},
booktitle={The Fourteenth International Conference on Learning Representations},
year={2026},
url={https://openreview.net/forum?id=0yOsSMU1eY}
}

@inproceedings{sun2026spa-cache,
title={Singular Proxies for Adaptive Caching in Diffusion Language Models},
author={Wenhao Sun and Rong-Cheng Tu and Yifu Ding and Zhao Jin and Jingyi Liao and Yongcheng Jing and Dacheng Tao},
booktitle={Forty-third International Conference on Machine Learning},
year={2026},
url={https://openreview.net/forum?id=4Lucu401dx}
}

@inproceedings{ye2025flashinfer,
title={{FlashInfer}: Efficient and Customizable Attention Engine for {LLM} Inference Serving},
author={Zihao Ye and Lequn Chen and Ruihang Lai and Wuwei Lin and Yineng Zhang and Stephanie Wang and Tianqi Chen and Baris Kasikci and Vinod Grover and Arvind Krishnamurthy and Luis Ceze},
booktitle={Eighth Conference on Machine Learning and Systems},
year={2025},
url={https://openreview.net/forum?id=RXPofAsL8F}
}

@inproceedings{juravsky2024hydragen,
title={Hydragen: High-Throughput {LLM} Inference with Shared Prefixes},
author={Jordan Juravsky and Bradley Brown and Ryan Saul Ehrlich and Daniel Y Fu and Christopher Re and Azalia Mirhoseini},
booktitle={Workshop on Efficient Systems for Foundation Models II @ ICML2024},
year={2024},
url={https://openreview.net/forum?id=Ye3ia34Fpp}
}
